\documentclass[10pt,twocolumn,letterpaper]{article}

\usepackage[pagenumbers]{cvpr} 

\usepackage{graphicx}
\usepackage{amsmath}
\usepackage{amssymb}
\usepackage{booktabs}
\usepackage{algorithm}
\usepackage{algpseudocode}

\usepackage[pagebackref,breaklinks,colorlinks]{hyperref}

\usepackage[capitalize]{cleveref}
\crefname{section}{Sec.}{Secs.}
\Crefname{section}{Section}{Sections}
\Crefname{table}{Table}{Tables}
\crefname{table}{Tab.}{Tabs.}

\def\cvprPaperID{8697} 
\def\confName{CVPR}
\def\confYear{2023}

\begin{document}
\setlength{\parskip}{0pt}
\definecolor{myOrange}{rgb}{0.8,0.3,0.}
\newcommand{\xifeng}[1]{\textcolor{myOrange}{#1}}
\newcommand{\julian}[1]{\textcolor{blue}{#1}}

\title{Consistent Latent Diffusion for Mesh Texturing}

\author{Julian Knodt\\
Lightspeed Studios\\
Bellevue, Washington\\
{\tt\small julianknodt@global.tencent.com}
\and
Xifeng Gao\\
Lightspeed Studios\\
Bellevue, Washington \\
{\tt\small xifgao@global.tencent.com}
}
\maketitle

\begin{abstract}
Given a 3D mesh with a UV parameterization, we introduce a novel approach to generating textures from text prompts. While prior work uses optimization from Text-to-Image Diffusion models to generate textures and geometry, this is slow and requires significant compute resources. Alternatively, there are projection based approaches that use the same Text-to-Image models that paint images onto a mesh, but lack consistency at different viewing angles, we propose a method that uses a single Depth-to-Image diffusion network, and generates a single consistent texture when rendered on the 3D surface by first unifying multiple 2D image's diffusion paths, and hoisting that to 3D with MultiDiffusion~\cite{multidiffusion}. We demonstrate our approach on a dataset containing 30 meshes, taking approximately 5 minutes per mesh. To evaluate the quality of our approach, we use CLIP-score~\cite{clipscore} and Frechet Inception Distance (FID)~\cite{frechet} to evaluate the quality of the rendering, and show our improvement over prior work.

\end{abstract}

\section{Introduction}

\begin{figure}
\centering
\includegraphics[width=0.9\linewidth]{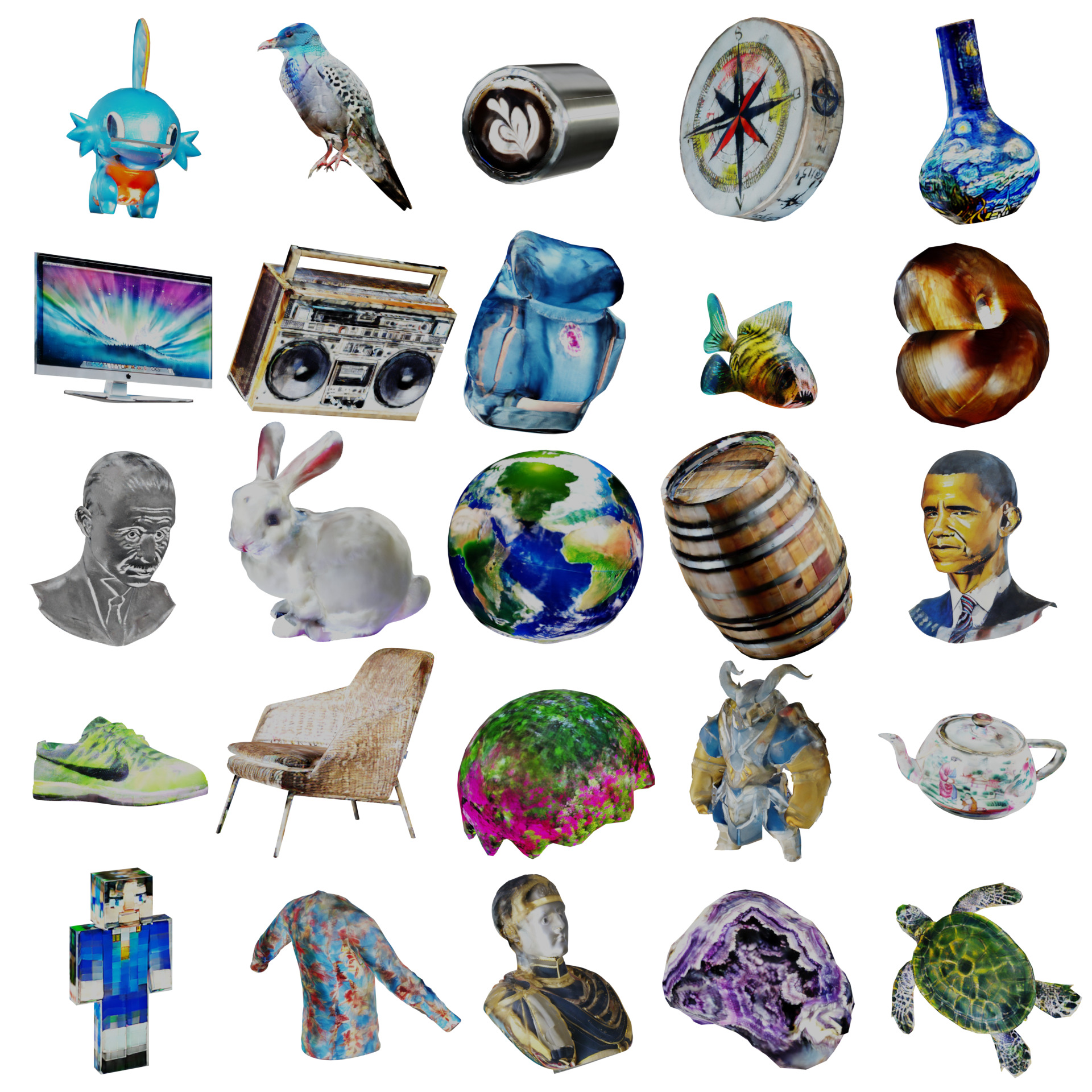}
\caption{A collection of meshes textured with our approach. We visualize rotated meshes to demonstrate that our approach fits more smoothly around surfaces, as compared to prior work such as TEXTure~\cite{TEXTure} that overfits to axis-aligned views. 3D model artist attribution provided on Github.}
\end{figure}

\label{sec:intro}

Creation of 3D models is a difficult task often requiring a trained artist and custom tooling~\cite{blender,ZBrush, UE5}, but they are common in games, shopping apps, and other applications. To reduce the burden of creating these models, recent work seeks to leverage 2D image generation to generate 3D geometry and textures. These works are often costly to run when optimizing both geometry and texture, requiring multiple GPUs and hours of training. We note that for many uses, there are already many meshes that can be used for generative texturing, without creating new geometry. This can be used for procedural asset generation in games, such as for objects like furniture, terrain, or non-playable characters, which lessens the burden for artists to create repetitive static content. With generative texturing, we can increase the diversity of content without requiring significant computational resources.

The current state of the art for mesh texturing from text~\cite{TEXTure, text2tex, texfusion, hdconsistencymodel} utilizes multiple diffusion models and a number of heuristics to stitch together multiple different views of the same mesh, varying from prior work which often was not general to all mesh surfaces and operated directly using convolutions on their surface~\cite{texturify, fine_texturing, textured3dgan, tmnet, texturefields}. In practice though these textures are often poor quality for multiple reasons. First, they may exhibit artifacts along inpainting edges due to the random nature of diffusion. There may also be clear shading differences between different views, and texture stretching due to projection along surfaces which are not flat with respect to the camera. We find these issues in both TEXTure~\cite{TEXTure} and Text2Tex~\cite{text2tex}, as they both iteratively backproject and stitch generated images onto the surface, and have little control over the diffusion process.

In this work, we unify the diffusion process for multiple views, to jointly denoise them to generate a consistent texture on the surface of a mesh. Inspired by MultiDiffusion~\cite{multidiffusion} for panorama generation, we aggregate multiple diffusion steps into a single image, and then back-project from each upsampled view to get a single consistent output. While MultiDiffusion~\cite{multidiffusion} demonstrates their approach on a single large image for panorama generation, we instead use a single \textit{spherical harmonic latent texture map}, to render the mesh in latent space. By backprojecting each view in latent space, multiple views can be aggregated together from a single diffusion pass. We first apply this approach in 2D, to demonstrate consistent diffusion, and then hoist this to 3D for mesh texturing.

MultiDiffusion~\cite{multidiffusion} on a single image produces high-quality consistent output by mimicking a single diffusion path from the utilized diffusion model. Unlike panorama generation, we must also consider warping introduced by texture stretch and camera angle. We utilize multiple techniques to mitigate these effects, such as weighing the importance of pixels by their orientation towards the camera, and by varying the latent texture size per model based on the texel usage of the UV parameterization.

In summary, our contributions are as follows:
\begin{enumerate}
    \item A diffusion approach that allows for pixel-wise similarity in a masked region.
    \vspace{-0.2cm}\item A generalization of MultiDiffusion~\cite{multidiffusion} to texturing 3D surfaces.
    \vspace{-0.2cm}\item A comparison of this work to TEXTure~\cite{TEXTure} and Text2Tex~\cite{text2tex}.
\end{enumerate}
\section{Related Work}

\paragraph{Mesh Texturing}
Many approaches exist to texture the surface of a mesh, such as PTEX~\cite{ptex}, HTEX~\cite{htex}, tri-planar mappings~\cite{texturify, tensorf}, linearly interpolating between per-vertex colors~\cite{colorneus}, or most commonly UV mapping~\cite{multichart_geometry}. We use UV mapping, which cuts a mesh into multiple surfaces homeomorphic to a plane, and flattens each of these surfaces into a shared texture space, upon which an image is painted. The texture can be created by an artist using Digital Content Creation tools~\cite{blender, ZBrush, UE5} or through an automatic process. During rendering, this image is resampled onto the surface of the mesh, creating the desired appearance.
UV mapping runs in real time, and is suitable for arbitrary mesh topologies, so it is widely used in rendering and games. It is also suitable for backprojecting textures, such as in~\cite{nvdiffmod, nvdiffrast, seamless}, which takes rendered images and project pixels back onto the original mesh. Our work also performs better with UV projections that have minimal distortion, and a plethora of work has gone into minimizing distortions~\cite{slim, bijective, lscm, isometric_distortion, signal_special_linear, mips, Floater2005}. xatlas~\cite{xatlas} to produce a UV mapping for each model, unless it comes with a sufficient mapping.

\paragraph{Text to Image}
There have been large leaps in text to image generation, such as Stable Diffusion~\cite{stable_diffusion}, Imagen~\cite{imagen},  and commercial software such as Midjourney AI, amongst others~\cite{kang2023gigagan, diffusion_design, ddim, clip_image_gen, cascaded_diffusion}. Most work leverages ``diffusion'', which takes a noisy image $I + \mathcal{N}(0, V)$, and outputs a new image $I + \mathcal{N}(0, V')$, such that $V' < V$, where $\mathcal{N}(0, 1)$ is the normal distribution with mean 0 and variance 1. By training a network on millions of images, conditioned on a text description of the image, a function is learned that inverts added noise, and produces highly-detailed, realistic images. These tools can match the quality of an artist, and their implications for society are still being explored.

\paragraph{Text To 3D}
Given the explosion of Text-to-Image, there has also been interest in leveraging these tools to generate textures for 3D models~\cite{TEXTure, texturify, text2tex, textguided, pointdiffusion, tmnet, texturefields, textured3dgan, fine_texturing}, and entire 3D models themselves~\cite{poole2022dreamfusion, prolificdreamer, fantasia3D, textmesh, gaudi, graf, pigan, hologan, giraffe, shape_induction, magic3d, triplane_diffusion}. The current state of the art in mesh texturing, TEXTure~\cite{TEXTure}, uses Text-to-Image, Inpainting, and Depth-to-Image models to render a mesh from multiple views and heuristics to stitch these images together to generate a single texture. For example they inpaint in a checkerboard pattern to increase consistency of their results. TEXTure requires 5 minutes to run, as it is not an optimization process, in constrast to generative optimization approaches such as DreamFusion~\cite{poole2022dreamfusion} which may take hours, and requires a cluster of GPUs, making it impractical for artistic use. As an aside, we note that some of these works may not be peer-reviewed or verified, and there are a number of commercial tools which do not document their process. 

\begin{figure*}
    \centering
    \includegraphics[width=0.8\linewidth]{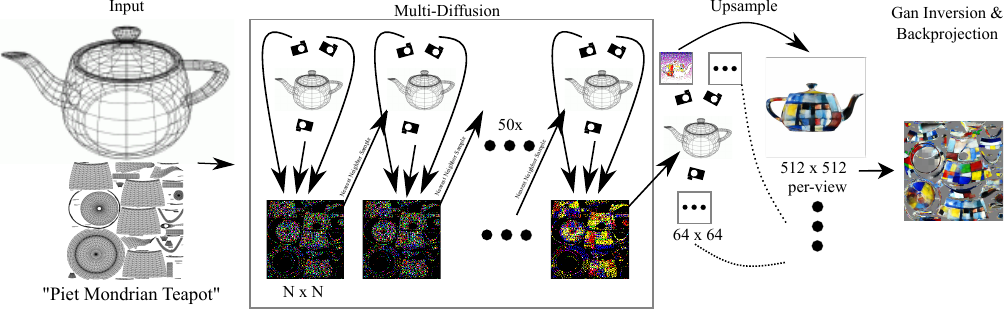}
    \put(-160, -5){\includegraphics[width=0.35\linewidth]{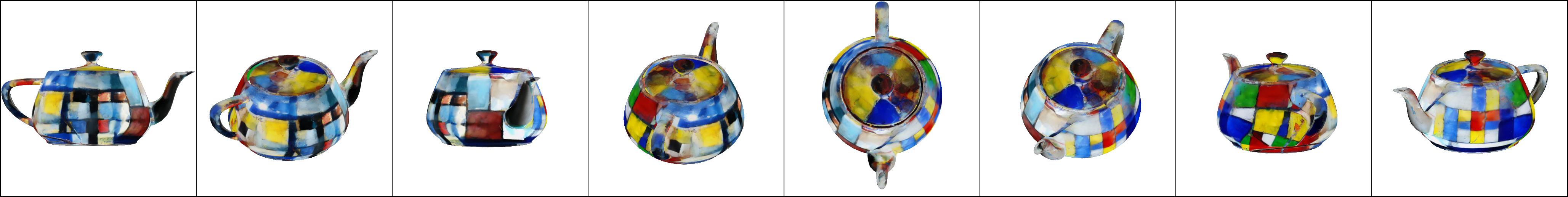}}
    \caption{Multi-Diffusion Mesh Texturing. Our input is a mesh with UV, and a text prompt. We then perform multi-diffusion with a latent texture from multiple camera views. We upsample each latent image, perform GAN inversion to stitch the images together in latent-space, and finally backproject into a single texture in image space.}
    \label{fig:pipeline}
\end{figure*}

\section{Consistent Diffusion across Batches}

\begin{figure}[ht]
    \centering
    \begin{tabular}{c c c c}
         \hspace{1.0em} Jungle \hspace{1.5em} & Angel \hspace{0.3em} & \hspace{0.3em} Gears  \hspace{-0.1em} & Piet Mondrian  \\
         \multicolumn{4}{c}{\includegraphics[width=0.9\linewidth]{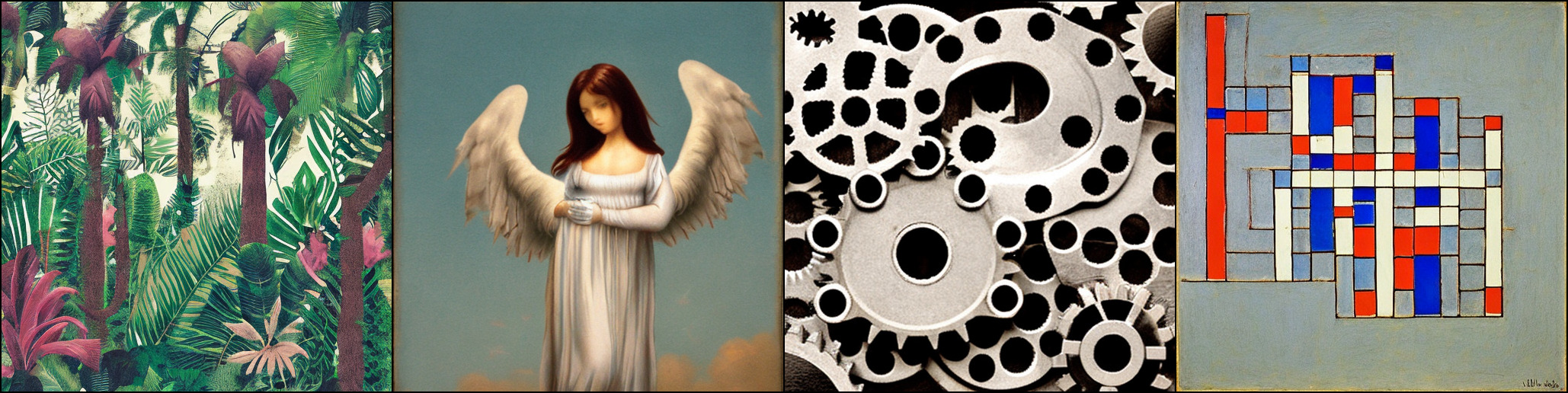}}\put(-\linewidth+8,16){\rotatebox{90}{$\alpha=1$}} \\
         \multicolumn{4}{c}{\includegraphics[width=0.9\linewidth]{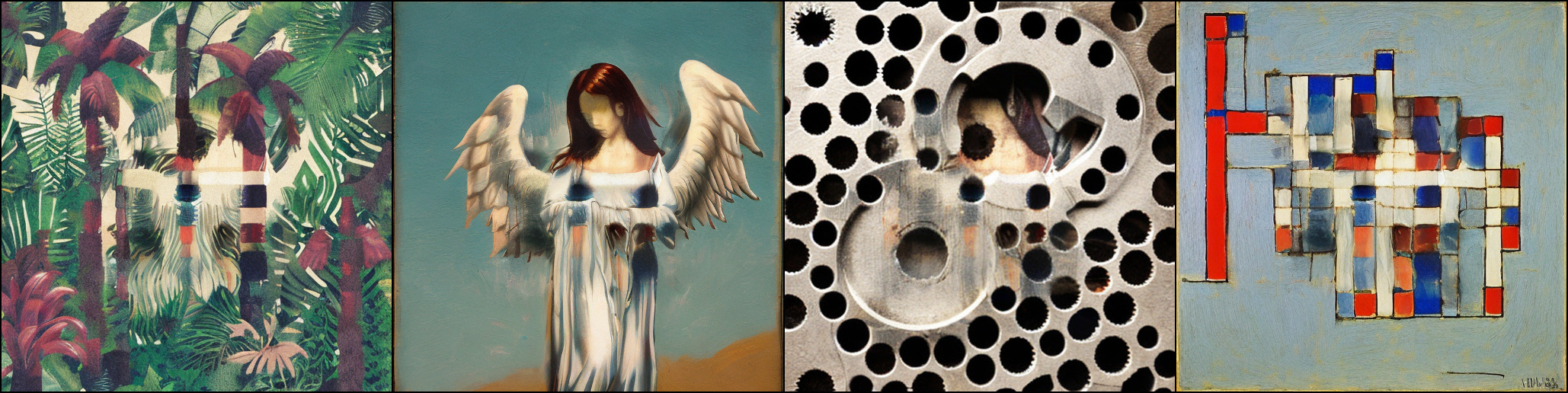}}\put(-\linewidth+8,10){\rotatebox{90}{$\alpha=0.97$}} \\
         \multicolumn{4}{c}{\includegraphics[width=0.9\linewidth]{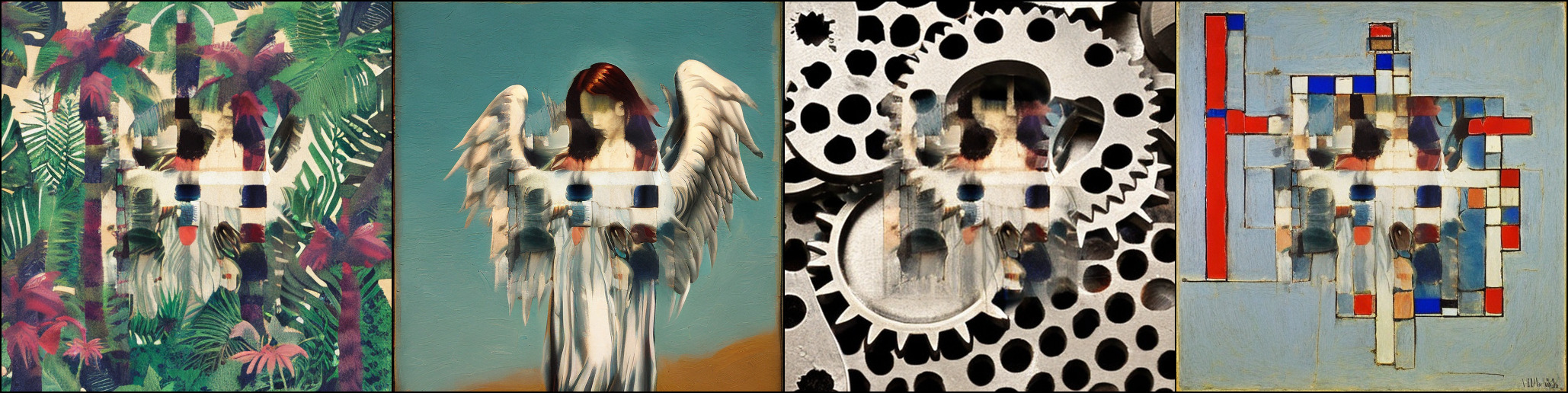}}\put(-\linewidth+8,10){\rotatebox{90}{$\alpha=0.94$}} \\
         \multicolumn{4}{c}{\includegraphics[width=0.9\linewidth]{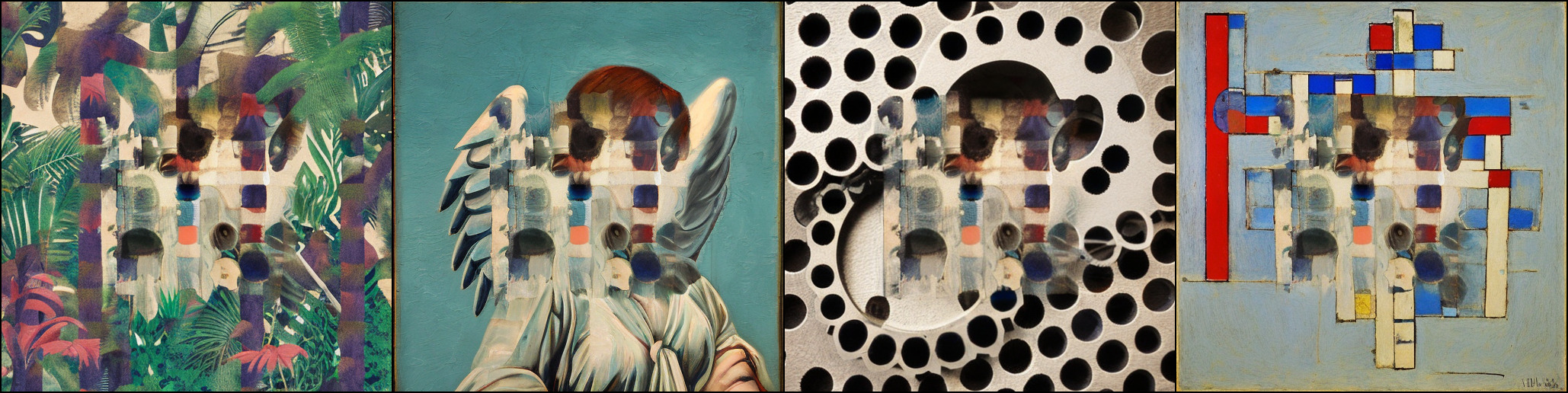}}\put(-\linewidth+8,10){\rotatebox{90}{$\alpha=0.85$}} \\
    \end{tabular}
    \caption{Consistent Latent Diffusion. For multiple prompts, diffusion paths are kept consistent in the center crop of each image while keeping the quality of the original. As $\alpha$ goes from 1 to 0, consistency increases, but similarity with $\alpha = 1$ degrades. For this example, we use the DDIM sampler with 100 steps.}
    \label{fig:consistent_diffusion}
\end{figure}

Before we generate pixel-wise consistent views on 3D meshes, we first consider consistent diffusion across multiple images with different prompts. We modify the diffusion process, first by adding the same shared noise to all images in latent space, and ensure that they remain consistent through a joint update step, which denoises based on the average of all update steps for all images. By uniformly updating all latent-space pixels, we ensure that they converge to approximately similar pixel-wise images. Pseudocode is outlined in Alg.~\ref{alg:consistent_latent_diffusion}, and example output is shown in Fig.~\ref{fig:consistent_diffusion}.

\begin{algorithm}
\caption{Consistent Latent Diffusion}
\small
\label{alg:consistent_latent_diffusion}
\begin{algorithmic}[1]
\Statex \textbf{Input:} $N$ prompts, mask $m$, Diffusion $D$, $\alpha\in[0,1]$
\Statex \textbf{Output:} $N$ images $I$ s.t. $\forall i,j : I_i[m] \approx I_j[m]$

\Statex $I_0 =$ I.I.D. Gaussian Noise $\in \mathbb{R}^{N\times512\times512}$
\Statex $S_\text{ident} =$ I.I.D. Gaussian Noise $\in \mathbb{R}^{1\times64\times64}$
\Statex $S_\text{indep} =$ I.I.D. Gaussian Noise $\in \mathbb{R}^{N\times64\times64}$ 
\Statex{$\rhd$ Share noise in masked region:}
\State $U_0 =\text{encode($I_0$)} + \text{where}(m, S_\text{ident}, S_\text{indep})$
\For{\texttt{$i \in [0, \text{steps}]$}}\Comment{Diffusion}
\State $U'_{i+1}\in\mathbb{R}^{N\times64\times64} = D(U_i)$
\State $\overline{U}_{i+1} \in\mathbb{R}^{1\times64\times64}= \frac{1}{N}\sum U'_{i+1}$
\Statex{$\rhd$ Within mask, lerp avg. and per image update:}
\State $U_{i+1} = \text{where}(m, \alpha U'_{i+1} + (1-\alpha) \overline{U}_{i+1}, U'_{i+1})$
\EndFor
\State \Return $\text{decode($U_\text{steps}$)}$ \Comment{Decode final image}
\end{algorithmic}
\end{algorithm}

It is critical that all images share the same noise. This is because each latent pixel is represented as $\mu + \delta$, $\mu\in\mathbb{R}, \delta\sim\mathcal{N}(0,\sigma)$. Averaging two latent-space pixels, $\frac{1}{2}(\mu_0 + \mu_1 + \delta_0 + \delta_1)$ breaks the assumption that $\mu + \delta$ is drawn from a distribution with variance $\sigma$. In the case that $\delta_0 = \delta_1$, the average will be $\frac{1}{2}(\mu_0 + \mu_1) + \delta$, which can be considered a sample from $\sim\mathcal{N}(\frac{1}{2}(\mu_0 + \mu_1), \sigma)$, thus preserving the variance assumed by the diffusion model.

We find that forcing all diffusion paths to exactly match leads to low-quality outputs, since it overly constrains the diffusion process. Instead, we provide some freedom to each diffusion path by introducing a parameter $\alpha\in[0,1]$, allowing more coherent outputs at the cost of exact equality. We test this diffusion process using Stable Diffusion 2.1, and observe that it is able to produce coherent and consistent output across multiple prompts as can be seen in Fig.~\ref{fig:consistent_diffusion}.


\section{Consistent Mesh Texturing}
To hoist our consistent diffusion process to 3D, we start with an untextured triangle mesh $M \triangleq (V, F), V \subset\mathbb{R}^3, F\subset V^{3}$ with a UV parameterization $\psi$ that maps each face to the 2D plane, a text prompt describing what the textured object should look like, a set of cameras, and a pretrained diffusion model. We use Stable Diffusion 2.1 with depth, and not Control Net~\cite{controlnet} as Text2Tex~\cite{text2tex} uses, but note that our approach could use either.
The final output is a texture map, such that the textured model from a fixed view should correspond to the diffusion model's single-image output.

To generate a consistent texture, we define an intermediate multi-diffusion step that optimizes a latent-space texture map given a set of views. Building on MultiDiffusion~\cite{multidiffusion},  we reuse an existing Diffusion Model, $D(I, T)\to I$, where $I$ is an image in $\mathbb{R}^{H\times W\times C}$, and $T$ is a text prompt. The diffusion process iteratively optimizes an image $I_0 \cdots I_n$, where each pixel in $I_0$ is assumed to be I.I.D. and sampled from the Gaussian distribution. Analogous to MultiDiffusion, we define another diffusion process $D'(U, T)\to U$, where $U$ is a texture map for a UV unwrapped mesh. $D'$ is meant to follow the original diffusion process $D$, and intends to minimize the following loss:
\begin{equation}
    \mathcal{L}_{Render} = \sum_{v\in V} \lVert W_v \otimes (R(v, U, M) - D(I|T, v)) \rVert_2 
\end{equation}
Where $R(v, U, M)$ is rasterization using nearest-neighbor sampling from the view $v$, given the latent texture map $U$, mesh $M$, and per-pixel weights $W_v$. By optimizing the rendered mesh with the same texture map across all views, we are merging the diffusion of all views completely. Later, we introduce spherical harmonics to control a level of independence from other views.
By minimizing this loss, we produce a texture that will be consistent with the original diffusion model from view $v$. Note that the texture map cannot be denoised directly, as the UV parameterization likely has discontinuities and is warped compared to the rendered image. To convert our final latent texture to actual rendered images, we convert the latent space of each camera view to image space, and then update the texture using differentiable rendering or other approaches~\cite{nvdiffmod, seamless}.

\paragraph{Spherical Harmonic Latent Texture Map}
In contrast to 2D Consistent Diffusion, Mesh Diffusion must use a single latent texture map. Since there is only a single view per pixel, each view is fully correlated with all other views, akin to setting $\alpha = 0$ in Consistent 2D Diffusion. As seen earlier, correlating all views reduces the quality of the output. To provide each view with some degrees of freedom, instead of storing a single latent value, we store spherical harmonic coefficients, $\text{SH}$, such that $\text{SH}(\theta, \phi) = \text{I} + \mathrm{N}(0, V)$, where $\theta$ and $\phi$ are view directions from a camera. The equation for spherical harmonics is
\begin{align}
\label{eq:phi}
\text{SH}_\text{u,v}(\theta,\phi)=\sum^\text{$N$}_{\ell=0} \sum^\ell_{m=-\ell}\text{SH}[\text{u,v}]Y^m_\ell(\theta, \phi),
\end{align}
where $\text{u,v}$ is an index into texture $\text{SH}$ containing coefficients and $Y^m_\ell$ is the real Legendre polynomial of order $\ell$.
This allows each view to be independent from other views. Spherical harmonics separates each view's latent values, allowing for higher quality per view images. Analogous to consistent latent diffusion we use a parameter $\alpha\in[0,1]$ to modulate view-independence and correlation.

To compute spherical harmonic coefficients for each denoising step, we directly solve the least-squares solution for the coefficients that minimizes the $\ell_2$ difference with each view's denoised result, incurring no noticeable cost compared to MultiDiffusion's~\cite{multidiffusion} approach. To initialize each view as random gaussian noise, and find the least square solution.
Conceptually, extending constant values to spherical harmonics is a generalization of MultiDiffusion analogous to switching from a constant BSDF to a view-dependent BSDF. For all of our experiments, we either have Spherical Harmonics of order 0 which is constant, or of order 1 which varies linearly with view direction, and fix $\alpha = 0.9$.

\paragraph{GAN Inversion for Consistency}
Even though each view uses the same latent texture map, after decoding they may not have consistent RGB pixel values. For Stable Diffusion, this is because the VAE decoder is not pixelwise-independent, and incorporates global information during decoding. On top of that, from different views texels may change their local neighborhood increasing inconsistency in RGB. This decoding error cannot be ignored, and leads to blurring if each view is mixed during backprojection in RGB. Prior work such as TEXTure~\cite{TEXTure} avoids blurring through a ``one-hot'' approach, as each texel is painted from one view, but this leads to inconsistency and seams along views. To mitigate  inconsistent VAE decoding, we perform some blending in \textit{latent space}. Akin to Blended Latent Diffusion~\cite{blendedlatent}, we mimic GAN inversion in the latent space of the diffusion model. We separate each view's latent image, and minimize the RGB difference when backprojected to all other views. Our GAN inversion is outlined in Alg.~\ref{alg:gan_inversion}, where our stopping criteria is a fixed number of steps.

\begin{algorithm}
\small
\caption{GAN Inversion Consistency}
\label{alg:gan_inversion}
\begin{algorithmic}[1]
\Statex \textbf{Input:} Per View Latents $L$, UV, Mask $M$, Weight $W$
\Statex \textbf{Output:} Optimized Per View Latents $L'$
\For{\texttt{$i \in [0, \text{steps}]$}}
    \Statex $\rhd \text{ Compute weighted average texture of all current views.}$
    \State $\overline{L} = \frac{1}{N}\sum^N_{i=0} W_i\text{backproject}(\text{decode}(L_i))$
    \For{$l \in L$}
        \Statex $\rhd \text{ For each view, backprop $\ell_1$ difference with avg.}$
        \State $\text{backprop}(\ell_1(\text{backproject}(\text{decode}(l)), \overline{L}))$
    \EndFor
    \State $L = L + \eta\nabla L$ \Comment{Optim. step}
\EndFor
\State \Return $L$
\end{algorithmic}
\end{algorithm}

While the objective is the same as backprojection in image space, it has a different optimization trajectory, because it is performed on the latent manifold. Performing the same optimization in RGB space blends semantically-meaningless RGB values, leading to blurring. We find GAN inversion to be better at mitigating small tone differences, texture shifts, and other differences caused by decoding, and by traversing latent space to fix RGB inconsistencies, there are fewer artifacts. We provide an ablation in the Appendix.

With Spherical Harmonic Latent Texture Maps, GAN inversion, and multi-view multidiffusion, our complete pipeline is given in Alg.~\ref{alg:mesh_tex_multi_diffusion}.

\begin{algorithm}
\small
\caption{Mesh Texture Multi-Diffusion}
\label{alg:mesh_tex_multi_diffusion}
\begin{algorithmic}[1]
\Statex \textbf{Input:} Mesh $M$ with UV, views $V$, Diffusion $D$
\Statex \textbf{Output:} Texture Map $\text{U}_\text{out}$
\Statex $\rhd \text{ Compute initial 0th Order SH texture map}$
\Statex $U_0 =$ i.i.d Gaussian Noise $\in \mathbb{R}^{N\times N}$
\For{\texttt{$i \in [0, \text{steps}]$}}\Comment{Multi-View Multi-Diffusion}
\For{\texttt{$v \in V$}}
\State $\text{I}' = D(Render(v, \text{U}_i, M))$ \Comment{Denoise}
\State $T_{i+1,j} = \text{backproject}(\text{I}', v, M)$
\EndFor
\Statex $\rhd \text{ Compute SH w/ Weighted Least Squares}$
\State $w = \text{V.weight}$ \Comment{Per pixel weight in each view}
\State $\text{U}_{i+1} = (1-\alpha) \text{Lstsq}(w\text{T}_{i+1,j}, w\text{V}) + \alpha\text{Lstsq}_\text{order 0}(\cdots)$
\EndFor
\State $\text{U}_\text{opt} = \text{GAN-Inv}(\text{U}_\text{last}, \text{M}.\text{uv}, \text{V}.\text{mask}, \text{V}.\text{weight})$
\State $\text{I}_\text{RGB} = \text{Decode}(\text{Render}(V, \text{U}_\text{opt}, M))$\Comment{Upsample}
\State $\text{U}_\text{out} = \text{DiffRender}(V, \text{I}_\text{RGB}, M)$\Comment{Backproject}
\State \Return $\text{U}_\text{out}$
\end{algorithmic}
\end{algorithm}

\paragraph{Mitigating Warping due to Projection by Weighing Normals}
Due to camera projections there is significant texture map warping during rasterization. Texels may change their neighbor depending on the rendering angle, which violates assumptions made during the diffusion process. In latent-space denoising~\cite{stable_diffusion}, this may lead to a number of artifacts, as the decoding step does not guarantee independence between pixels, thus optimizing a single pixel may lead to a completely different upscaled region when the mesh is rotated, leading to poor joint diffusion. While we add GAN inversion to mitigate this, we also mitigate this by weighing the importance of each pixel by the cosine similarity of the projected face's normal and the camera's viewing direction. This ensures that the surface which is flat with respect to the camera will be prioritized. This weighing is used during multi-diffusion, GAN-Inversion and backprojection. It's also necessary during back-projection, as some views may have warping artifacts, and it helps keep sharper features.

\paragraph{Reducing Inconsistency through Increased Guidance}
We find that some prompts cannot sufficiently express a desired visual image. For example, the prompt "Earth",  has artistic interpretations and photographic visuals for ``Earth''. This ambiguity may lead to a significant degradation during multi-diffusion, as multiple interpretations may not be easily stitched, leading to inconsistencies, blurring, and gray outputs. While this may be mitigated with prompt tuning or textual inversion, we find that increasing guidance scale during diffusion can lead to consistent output, at the cost of saturating colors. We ablate the choice of guidance scale for some meshes and prompts in Sec.~\ref{sec:ablate_guidance_scale}. We also find that including prompt modifiers, such as ``back'', ``front'', ``side'' based on the camera angle, akin to DreamFusion~\cite{poole2022dreamfusion}, produces better output.

\paragraph{Selecting Latent Texture Sizes}
With an arbitrary UV mapping, a specific set of views may not use enough texels to accurately recover a texture. When insufficient texels are used, the latent texture does not have enough freedom to represent a smooth texture on the surface of the mesh. On the other hand, with too many pixels every view will be independent from all other views. Thus, selecting an appropriate texture size is important to maintaining consistency with good quality. We find that the sizes $128\times128$ and $196\times196$ are good defaults, and ablate this choice in Sec.~\ref{sec:ablate_texture_size}.

\paragraph{Selecting Camera Parameters}
We sample cameras uniformly on the sphere using fibonacci sampling~\cite{sphere_points, spherical_fibonacci}. For meshes which are not viewed from below, we sample the upper hemisphere, which is done by using the absolute value of the y-coordinate of each original sample. We find 8 views provides high-quality output with sufficient consistency. We ablate this choice in Sec.~\ref{sec:ablate_camera_views}, and find that if including too many cameras it leads to poor results. In addition, we also ablate using cameras fixed to the XZ plane, which mitigates projection warping of an elevated camera, as can be seen in the Appendix.

Another design choice we make is to use orthographic cameras. While it is common to use perspective cameras that look plausible to the human eye, they introduce distortion by stretching distant objects. By using an orthographic camera, flat surfaces remain unstretched regardless of distance.

\section{Experiments}

\begin{figure}
    \centering
    \includegraphics[width=0.8\linewidth]{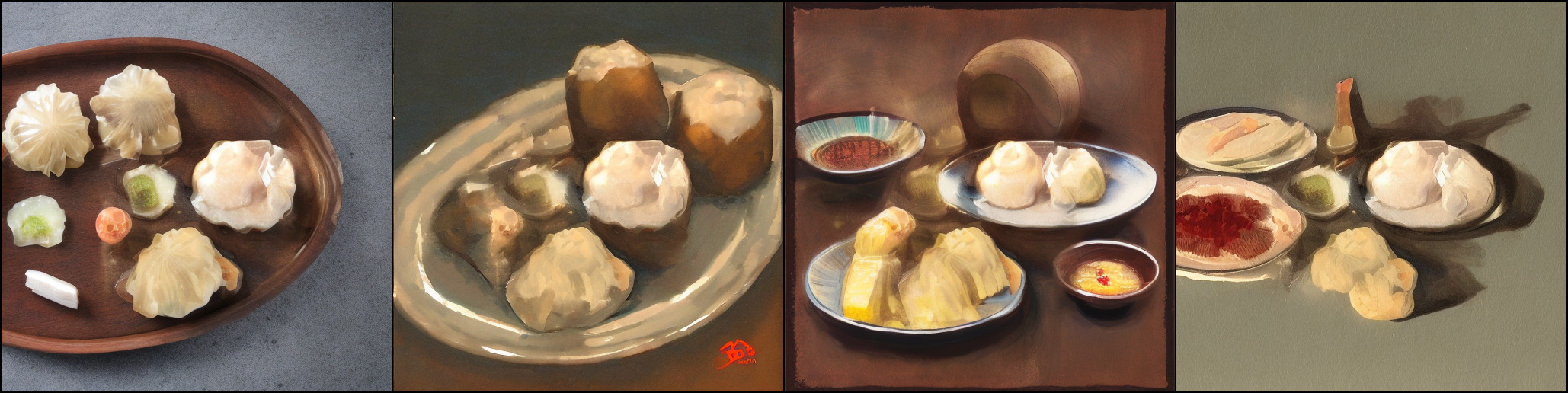}
    \includegraphics[width=0.8\linewidth]{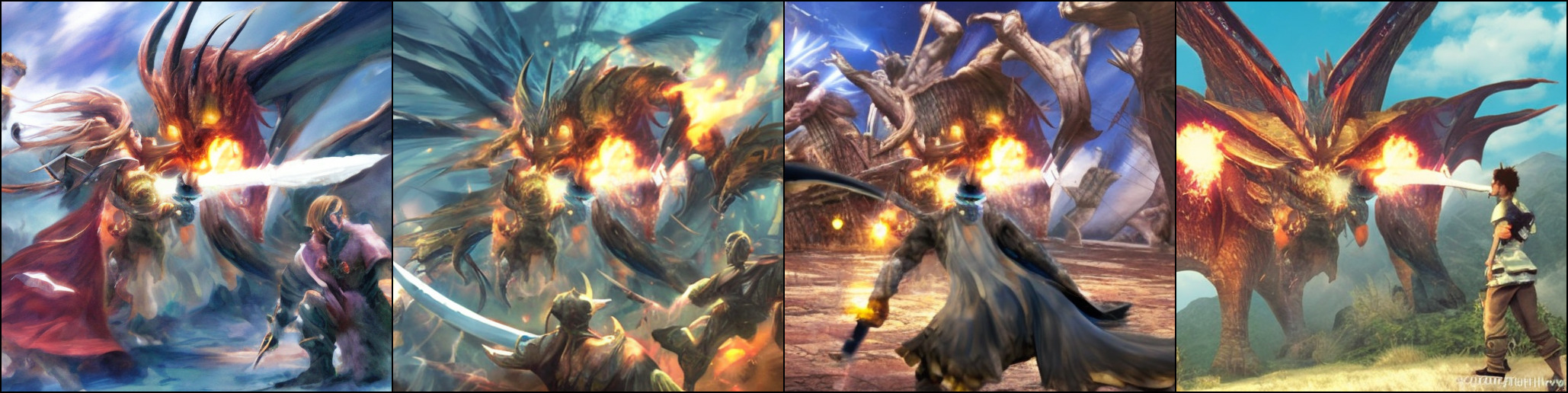}
    \includegraphics[width=0.8\linewidth]{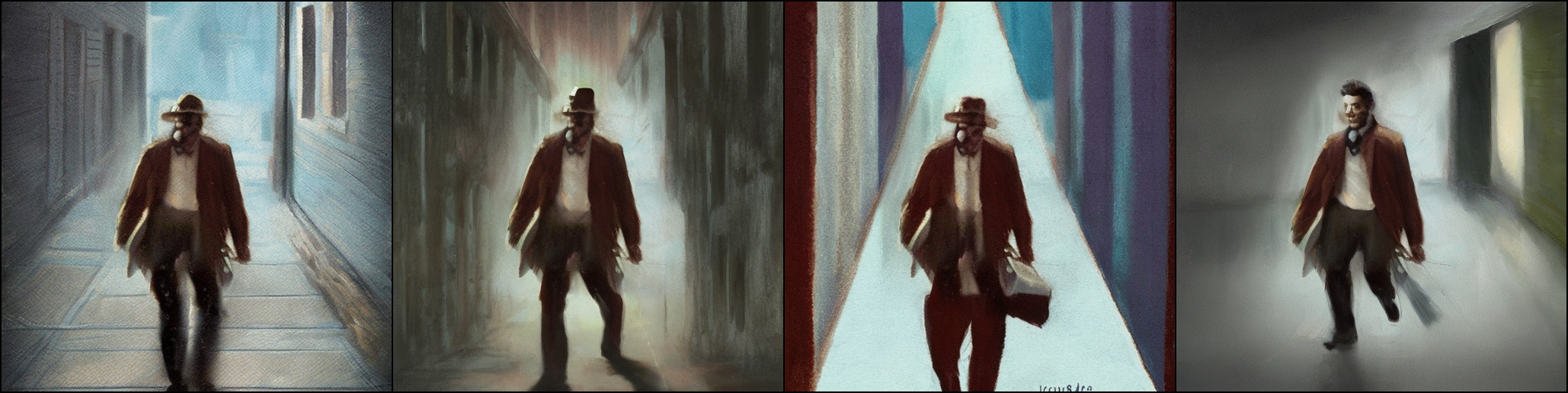}
    \includegraphics[width=0.8\linewidth]{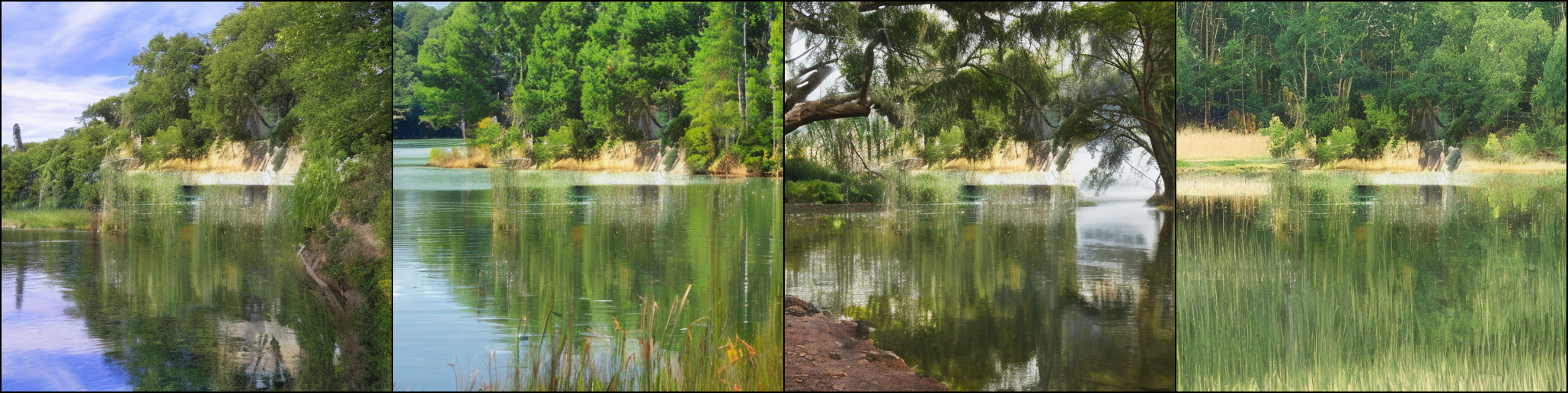}
    \caption{Consistent 2D diffusion on the same text prompts, $\alpha = 0.97$, 50 steps with the DDIM~\cite{ddim} sampler. Within the center region of the image, identical motifs are maintained, while the rest of the image is varied. The prompts from top to bottom are ``Dim sum still life'', ``Final Fantasy fighting a dragon'', ``A detective from an Edward Hopper painting running into a dark alley'', and ``Ghibli-style bamboo by a lake''.}
    \label{fig:consistent2D-diffusion-examples}
\end{figure}

\paragraph{Consistent 2D Diffusion}
We show some example results of Consistent Image Diffusion on the same text prompt, with the center 128 to 384 pixels unified. We fix $\alpha = 0.97$, and use 50 steps with the DDIM~\cite{ddim} sampler. Our approach is able to reproduce motifs across images. For example, in the produced ``Dim Sum Still Life'' images, the center crop contains similar pork buns and dumplings, but the rest of the image is different. Example results for Consistent 2D Diffusion are shown in Fig.~\ref{fig:consistent2D-diffusion-examples}.

\section{Consistent Mesh Diffusion}

We demonstrate the quality of our approach on multiple meshes with a variety of prompts, qualitatively comparing them to TEXTure~\cite{TEXTure}. To run our experiments, we use a single NVIDIA GeForce RTX 3090, with a 32 core AMD processor. Each model takes about 5 minutes to process. For the diffusion model, we use Stable-Diffusion 2's Depth2Image Pipeline from Huggingface~\cite{stable_diffusion}. We use a variety of prompts that are related to the original input mesh's shape, and include additional examples in the Appendix to show the variability of our approach.

Our dataset consists mostly of manually collected meshes from Sketchfab~\cite{sketchfab}\footnote{We were careful to select meshes where artists did not forbid use in generative AI models at the time of download.} and 1-4 prompts related to each input mesh. For example, for a crow mesh, we use the prompts ``parrot'', ``pigeon'', and ``crow''. In total, there are 34 unique meshes, with 76 total prompts. We also increase the weight of the forward facing view during the diffusion process for some meshes, as this is the most salient view.

\subsection{Quantitative Results}

We perform quantitative comparisons of our approach against TEXTure~\cite{TEXTure}, and Text2Tex~\cite{text2tex}. 
We use Frechet Inception Distance~\cite{frechet, pytorch_fid} to evaluate fidelity and similarity to the original diffusion model and CLIP-Score~\cite{clipscore} to evaluate similarity to prompt.

\begin{figure}
    \centering
    \includegraphics[width=0.49\linewidth]{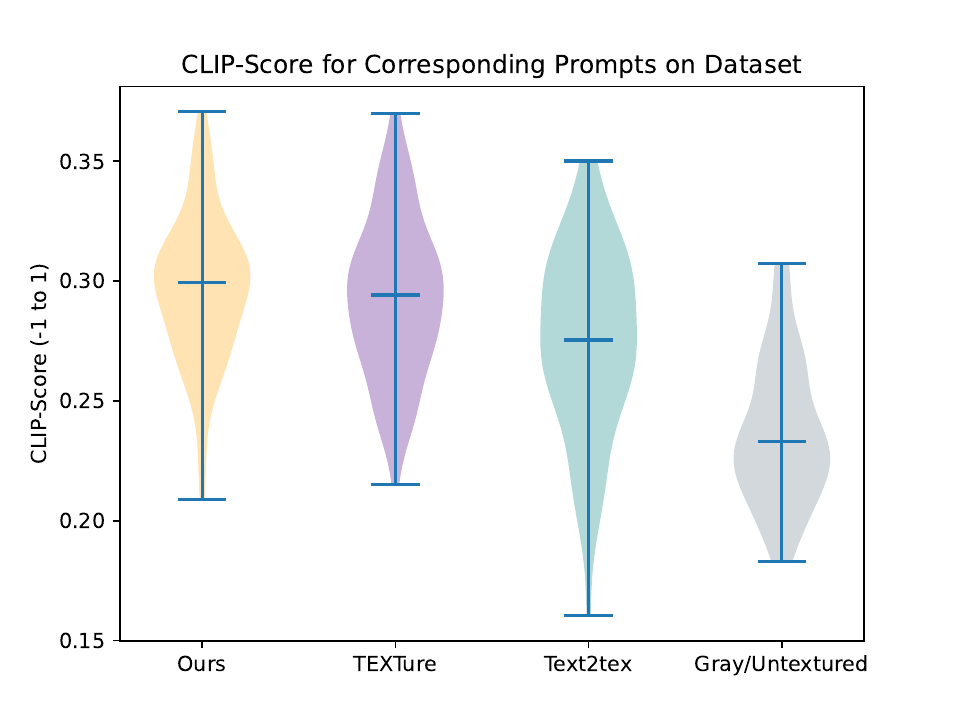}
    \includegraphics[width=0.49\linewidth]{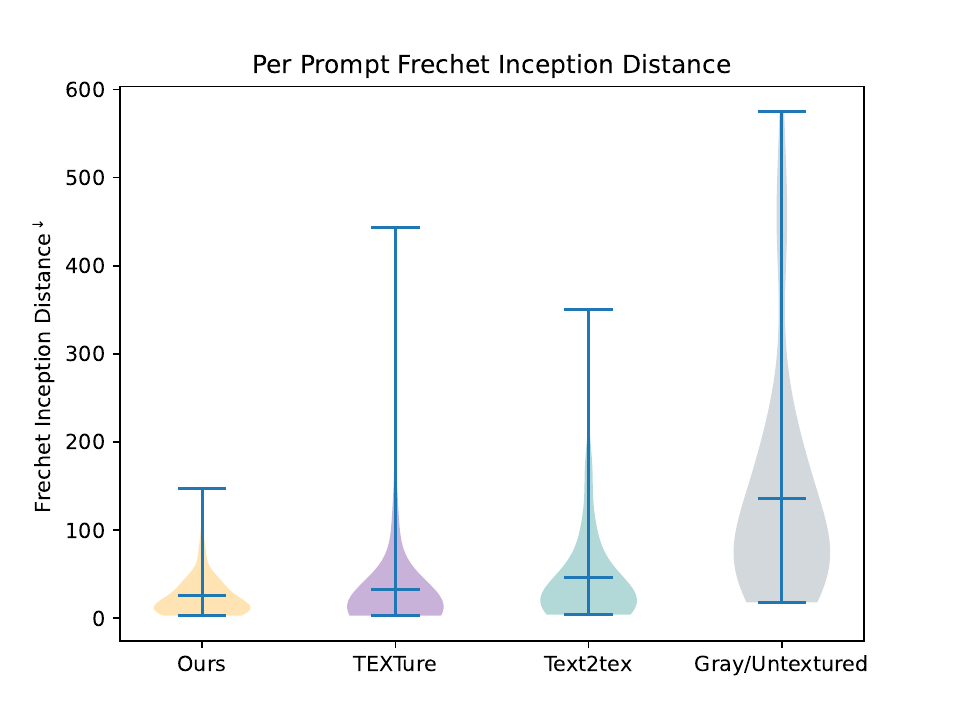}
    \begin{tabular}{|c|c|c|c|c|}
    \hline
    Median     & Ours & TEXTure & Text2tex & Gray  \\
    \cline{2-5}
    CLIP-Score$^\uparrow$ & 0.299 & 0.294 & 0.275 & 0.233  \\
    \hline
    Mean FID$^\downarrow$ & 25.54 & 32.95 & 46.38 & 136.23 \\ 
    \hline
    \end{tabular}
    \caption{CLIP-Score comparisons of our approach against other approaches. We evaluate the CLIP-Score~\cite{clipscore} on a number of views of the textured mesh against the prompt used to generate the input. CLIP-Scores range from $-1$ to $1$, where $1$ is most similar and $-1$ is least. Our approach is comparable to TEXTure~\cite{TEXTure} in CLIP-Score, and better in Frechet Inception Distance~\cite{frechet}, which we use to measure the distance from Stable Diffusion applied independently to each view.}
    \label{fig:clip_score_comparison}
\end{figure}

As shown by a gray baseline, the mesh itself provides cues that make it similar to the prompt, but adding a texture can improve correlation with the prompt. On CLIP-Score, our approach is comparable to TEXTure~\cite{TEXTure}, whereas Text2tex~\cite{text2tex} does not perform as well consistently, as shown in Fig.~\ref{fig:clip_score_comparison}.
As the distributions between TEXTure~\cite{TEXTure} and our approach are similar, it may indicate that certain meshes and prompts may be more challenging than others.

In our evaluation of fidelity, we use Stable Diffusion on 8 views independently, and then compute the frechet inception distance with 24 renderings of each retextured model. We find that our approach has a much tighter distribution than TEXTure and Text2Tex, and a lower mean, showing that on average our approach outperforms prior work.

\begin{figure}
    \centering
    \includegraphics[width=0.45\linewidth]{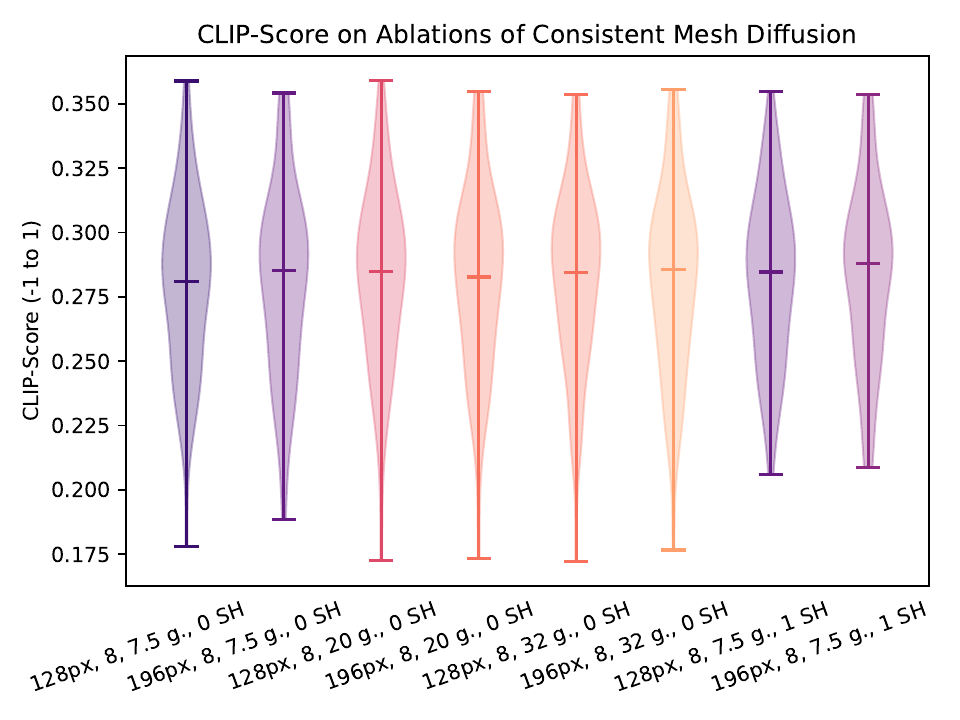}
    \includegraphics[width=0.45\linewidth]{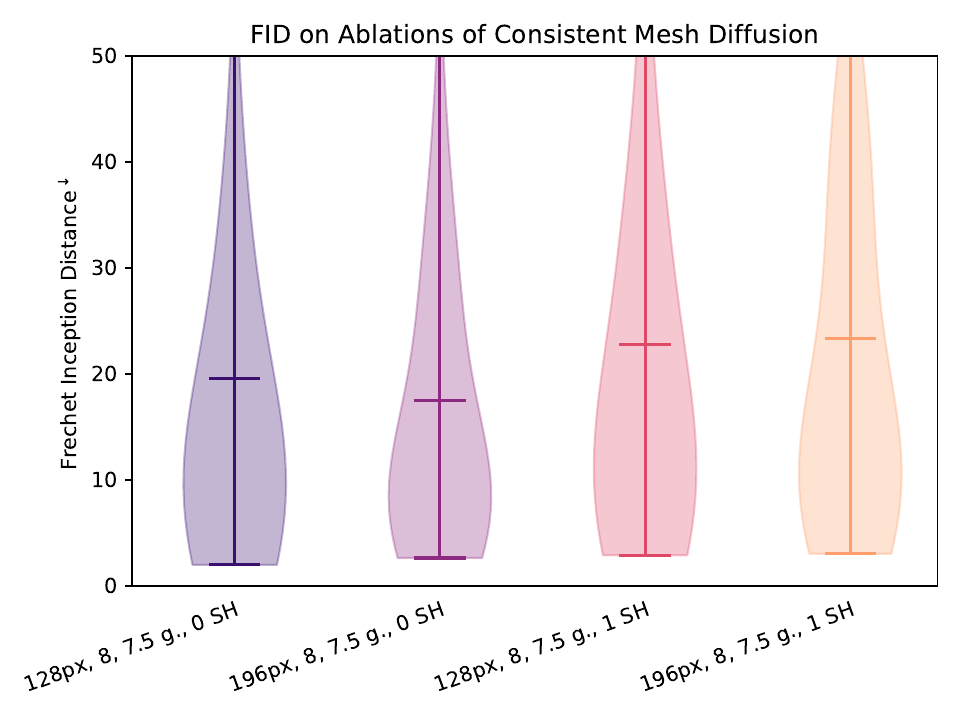}
    \begin{tabular}{|c|c|c|}
        \hline
       FID$^\downarrow$ & 196px & 128px  \\ \hline
       SH 0 & 17.47 & 19.53  \\ \hline
       SH 1 & 23.31 & 22.78  \\ \hline
    \end{tabular}
    \caption{Ablation of our approach, using CLIP-Score as a metric. We evaluate two latent texture sizes, 128 and 196 pixels, with 8 camera views, and 3 different guidance scales 7.5, 20, and 32. We find that using Spherical Harmonics of order 1 with our approach increases the consistency of results, but other parameters are best specified per mesh. We also compute the Frechet Inception Distance~\cite{frechet} for different spherical harmonic orders and different texture sizes. We find that on average, SH 0 has lower FID than SH1, but texel size does not have a clear trend across different spherical harmonic orders. TODO discuss fid ablations}
    \label{fig:clip_ablations}
\end{figure}

We also show an ablation of our approach with different hyper-parameters in Fig.~\ref{fig:clip_ablations}. Specifically we evaluate the choice of latent texture-size, number of cameras, and guidance scale. There isn't a consistent pattern for which hyper-parameters are better or worse, and is best to be evaluated per mesh. For the comparison of the datasets above, we took the max over the results with 8 cameras, 7.5 guidance scale, but varying texel size, as it is important to select that per mesh. We note that over all our datasets, taking the max shows an even larger improvement of $0.305$.

\subsection{Qualitative Comparisons}

\begin{figure*}[ht]
    \centering
    \begin{tabular}{c c c c c c}
        \multicolumn{2}{c}{Ours} & \multicolumn{2}{c}{TEXTure~\cite{TEXTure}} & \multicolumn{2}{c}{Text2Tex~\cite{text2tex}} \\
        Front & Back & Front & Back & Front & Back \\
        \vspace{-0.2cm}
        \includegraphics[width=0.15\linewidth]{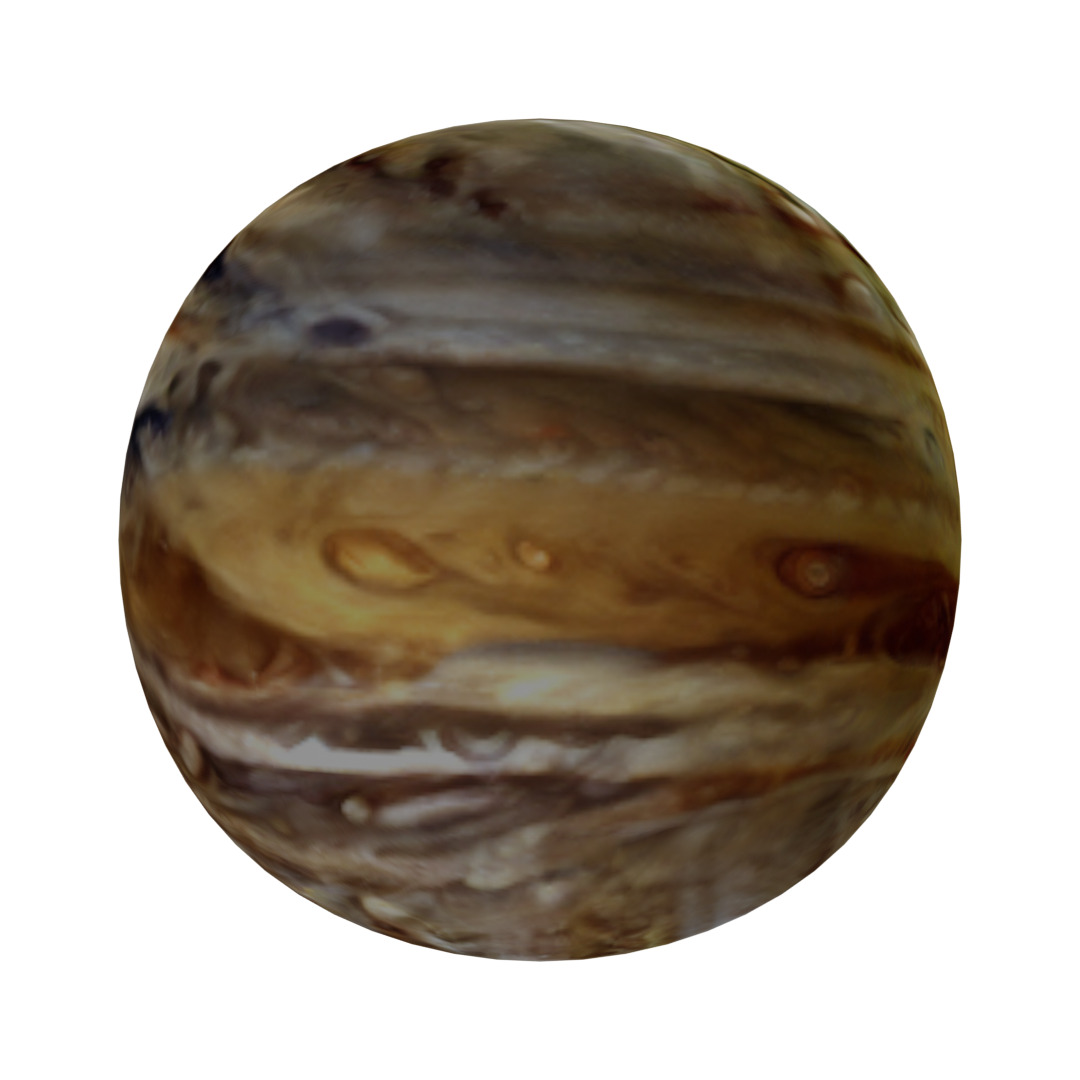} &
        \includegraphics[width=0.15\linewidth]{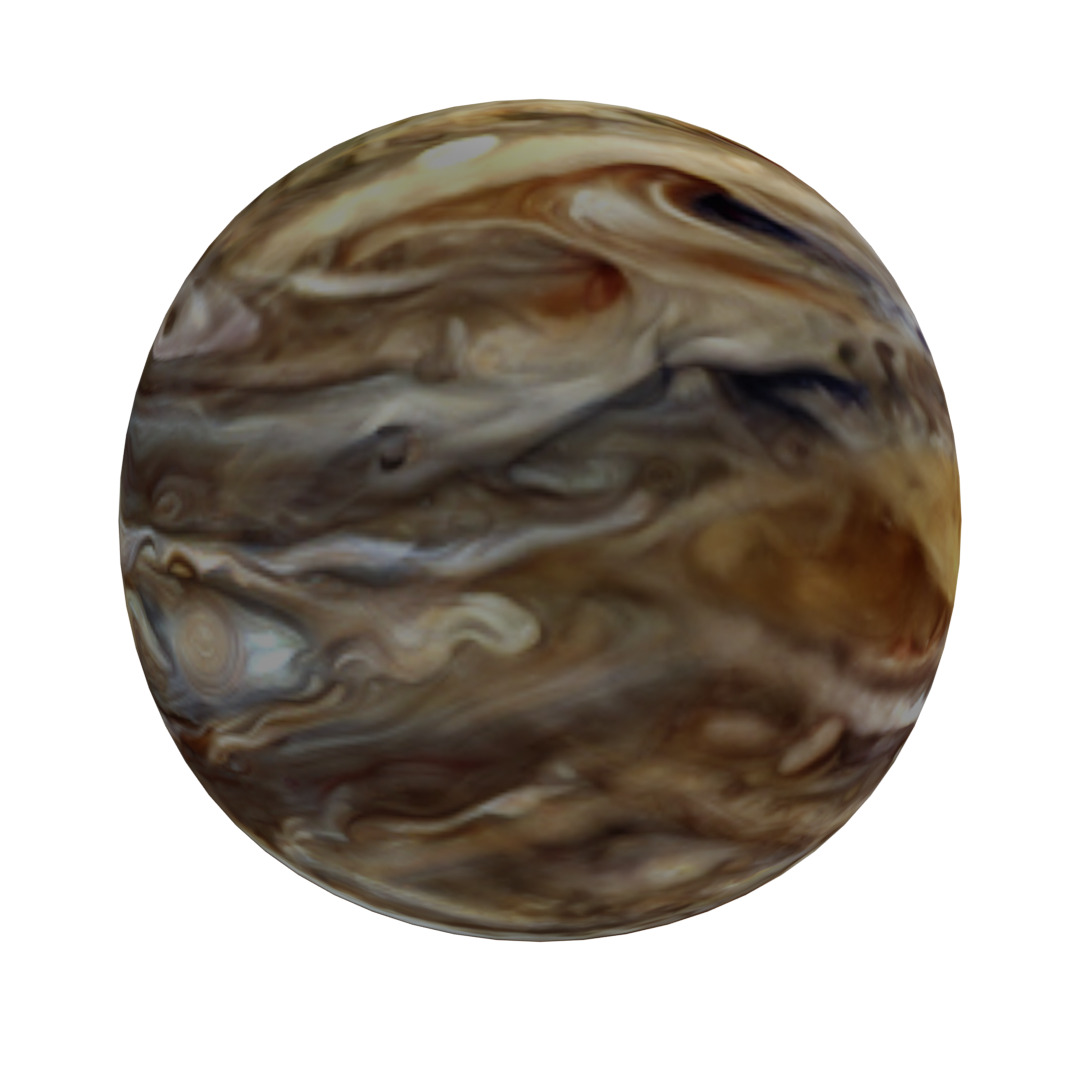} &
        \includegraphics[width=0.15\linewidth]{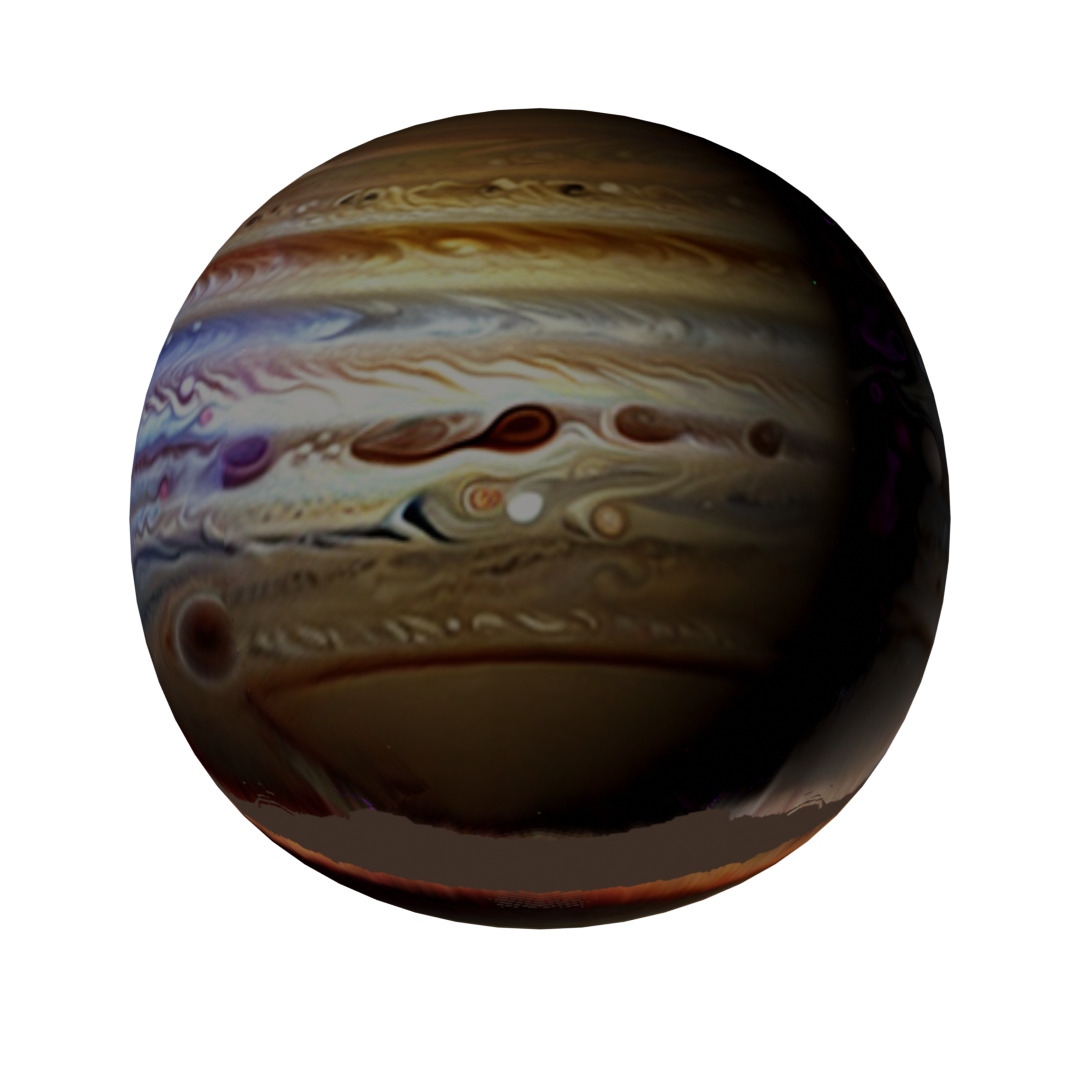} &
        \includegraphics[width=0.15\linewidth]{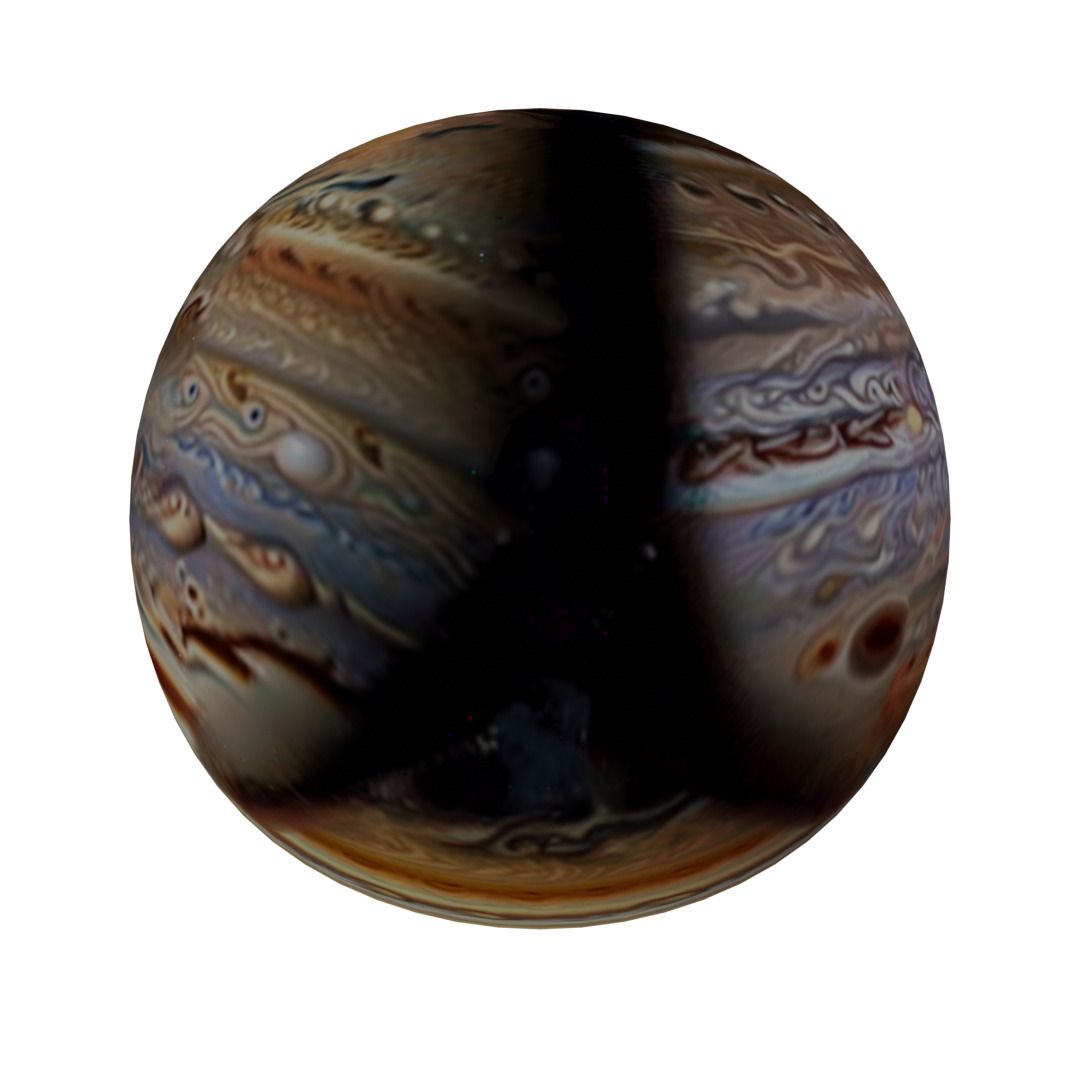} &
        \includegraphics[width=0.15\linewidth]{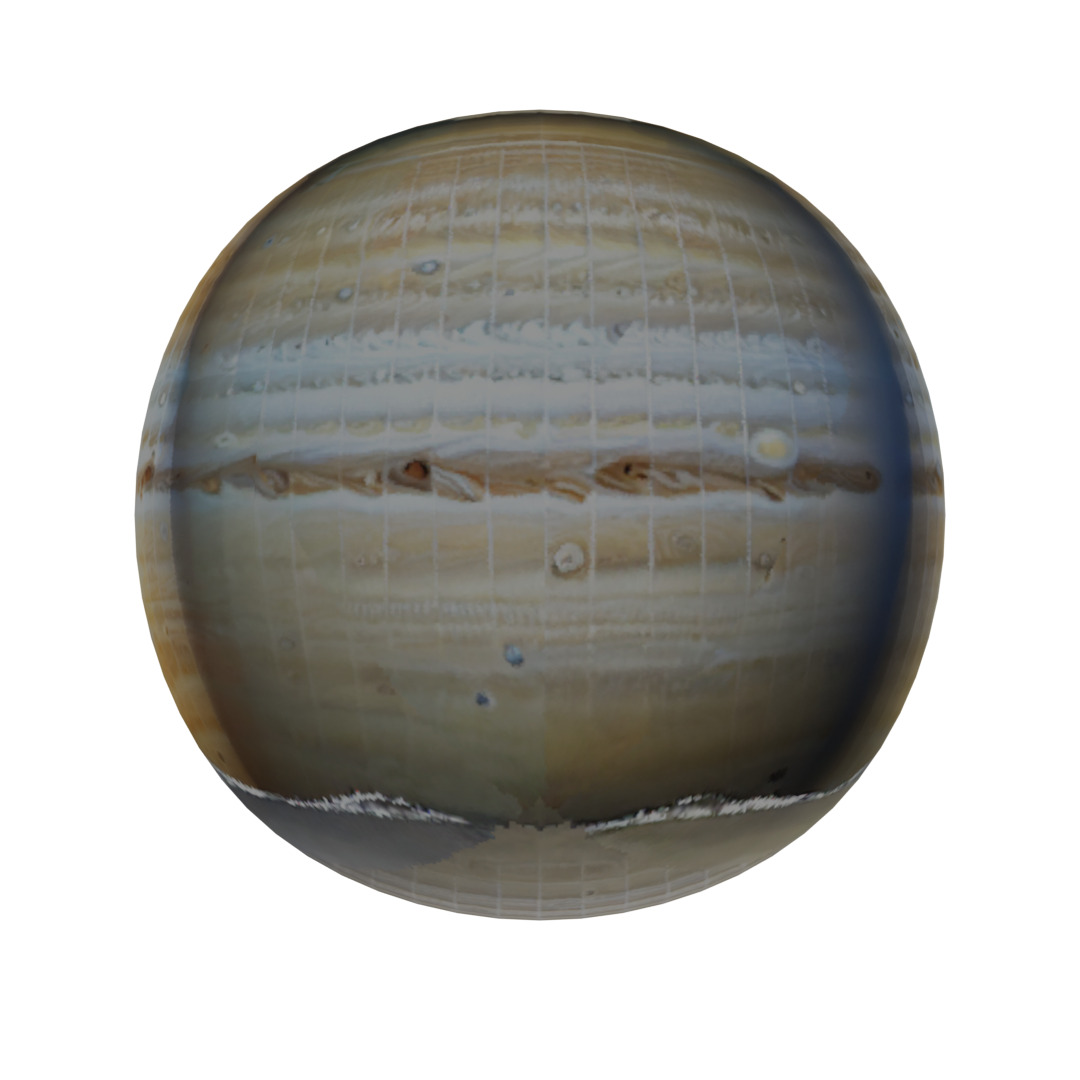} &
        \includegraphics[width=0.15\linewidth]{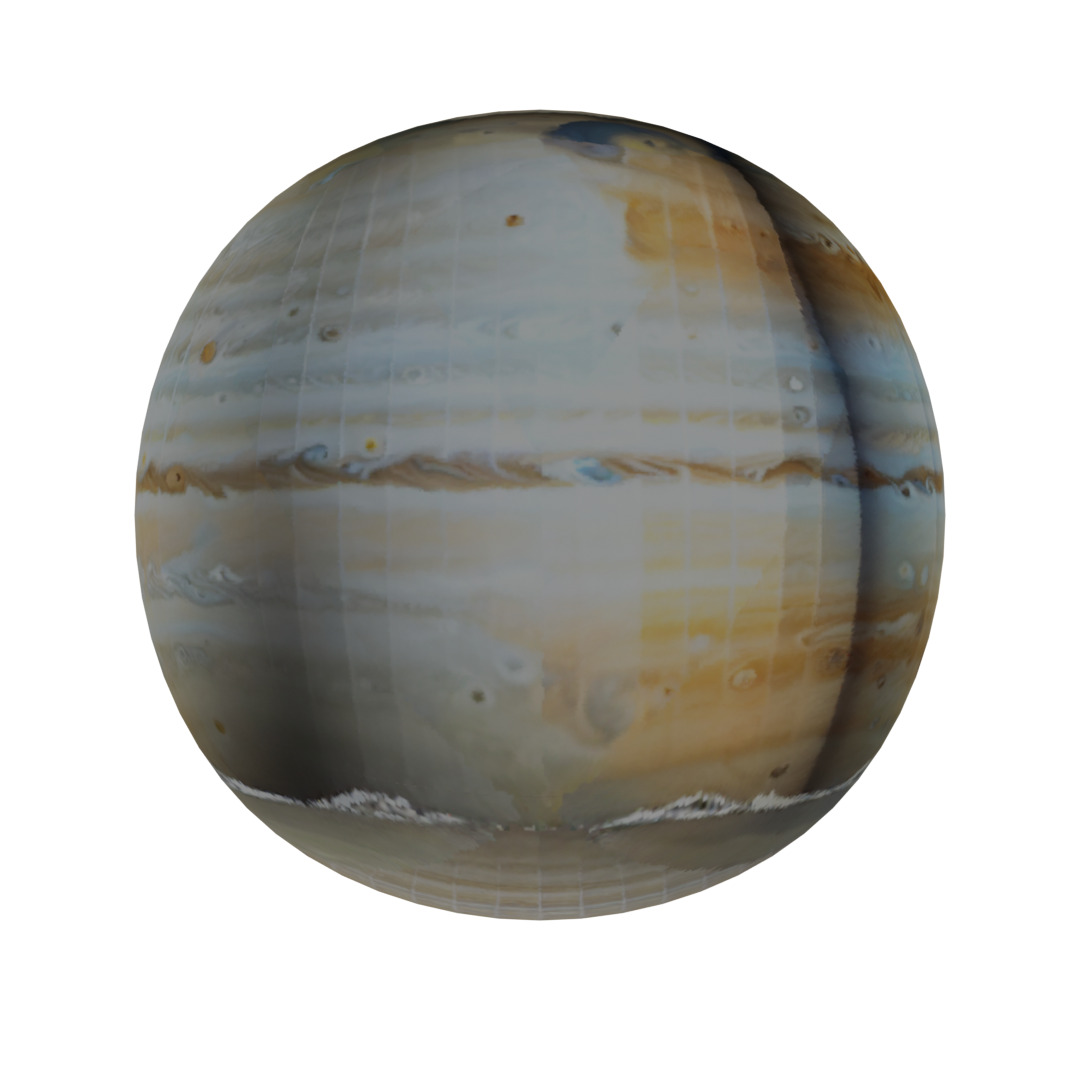} \\
        \vspace{-0.2cm}
        \includegraphics[width=0.15\linewidth]{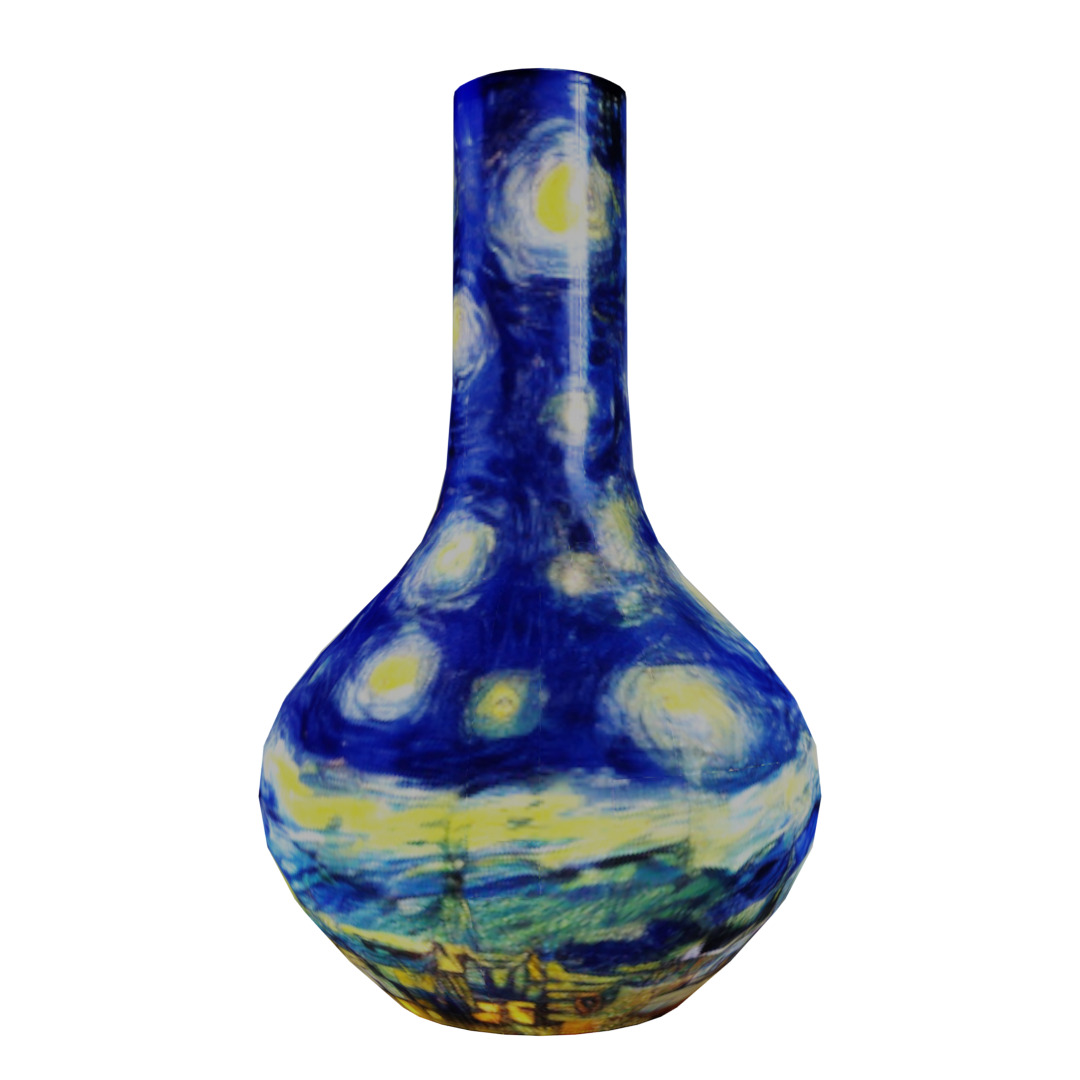} &
        \includegraphics[width=0.15\linewidth]{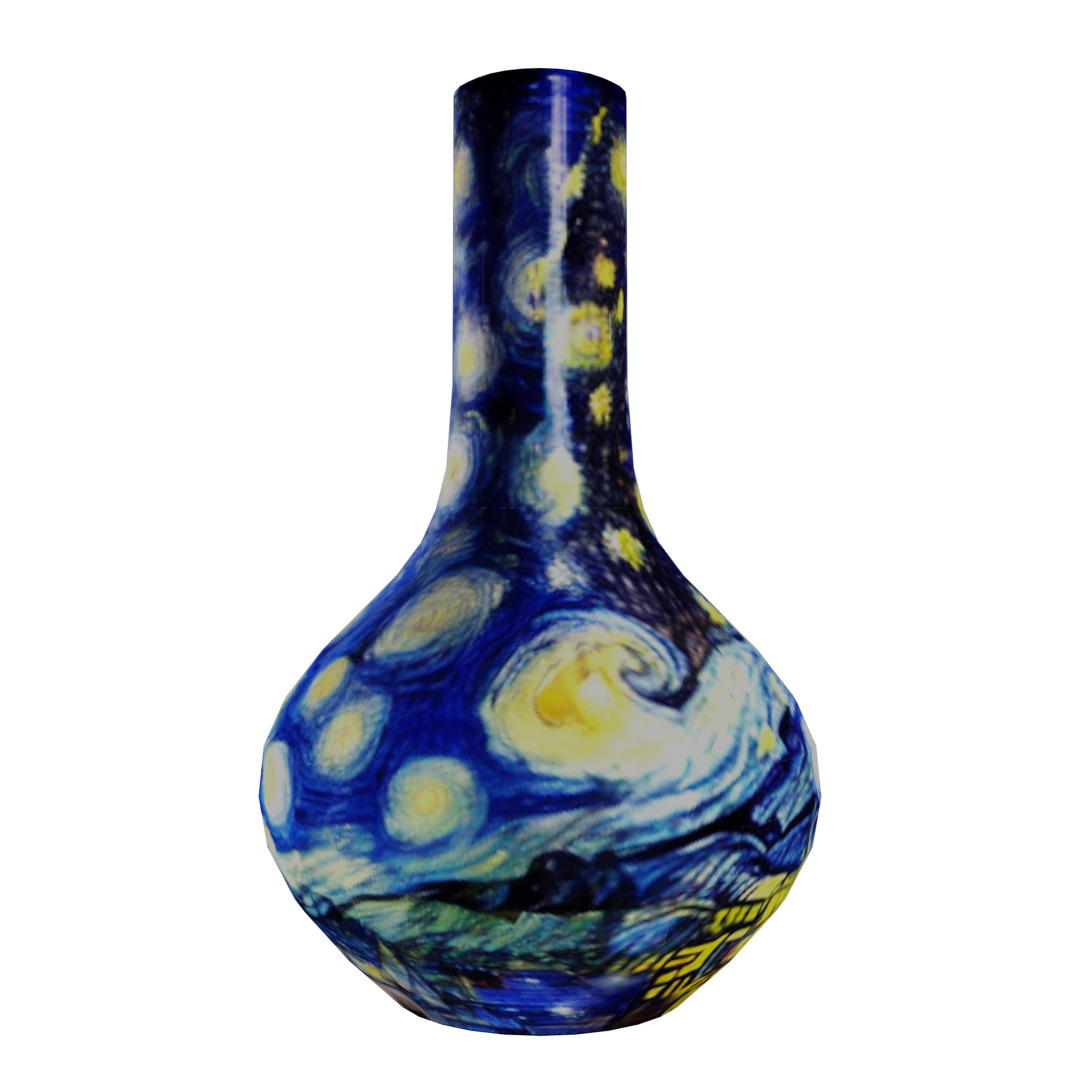} &
        \includegraphics[width=0.15\linewidth]{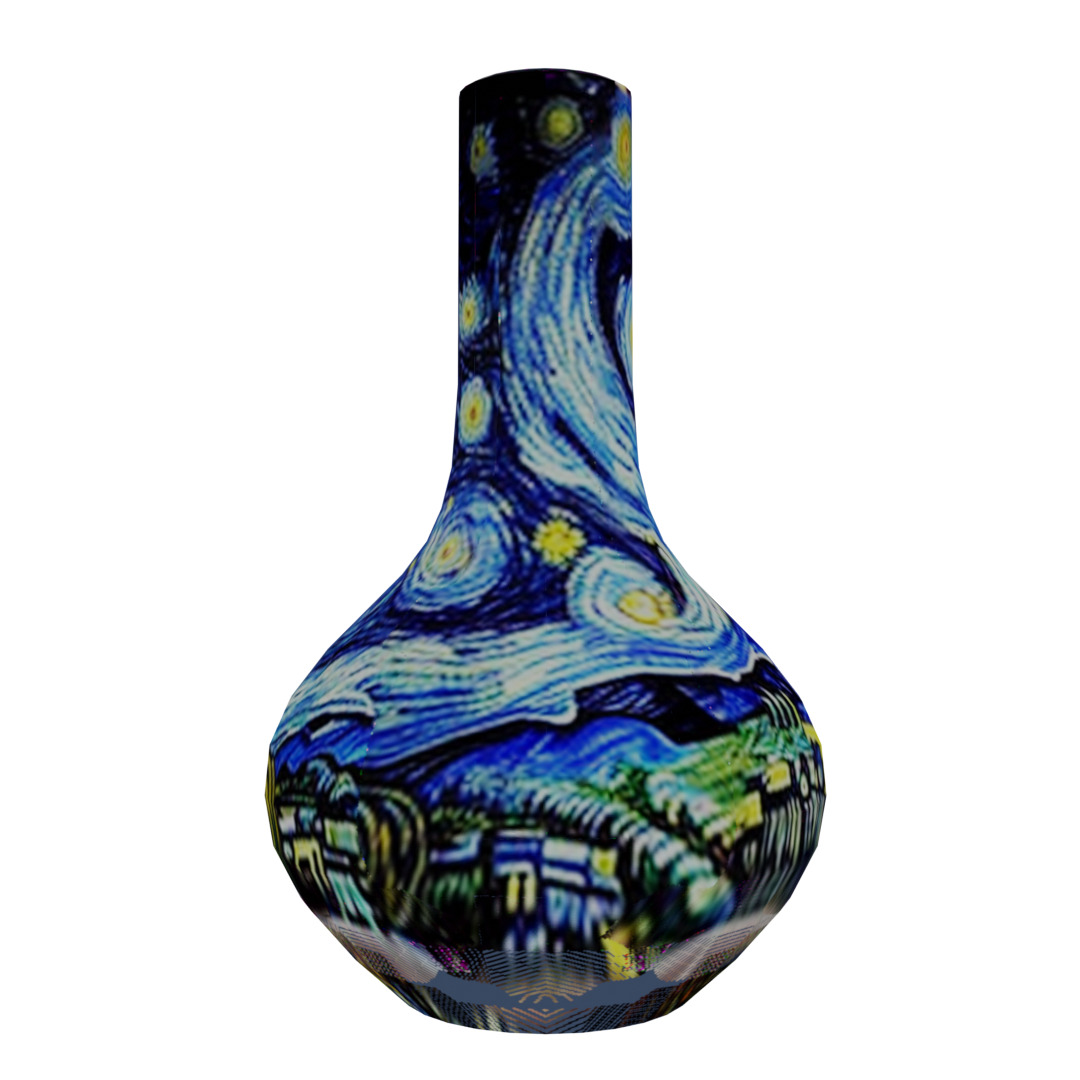} &
        \includegraphics[width=0.15\linewidth]{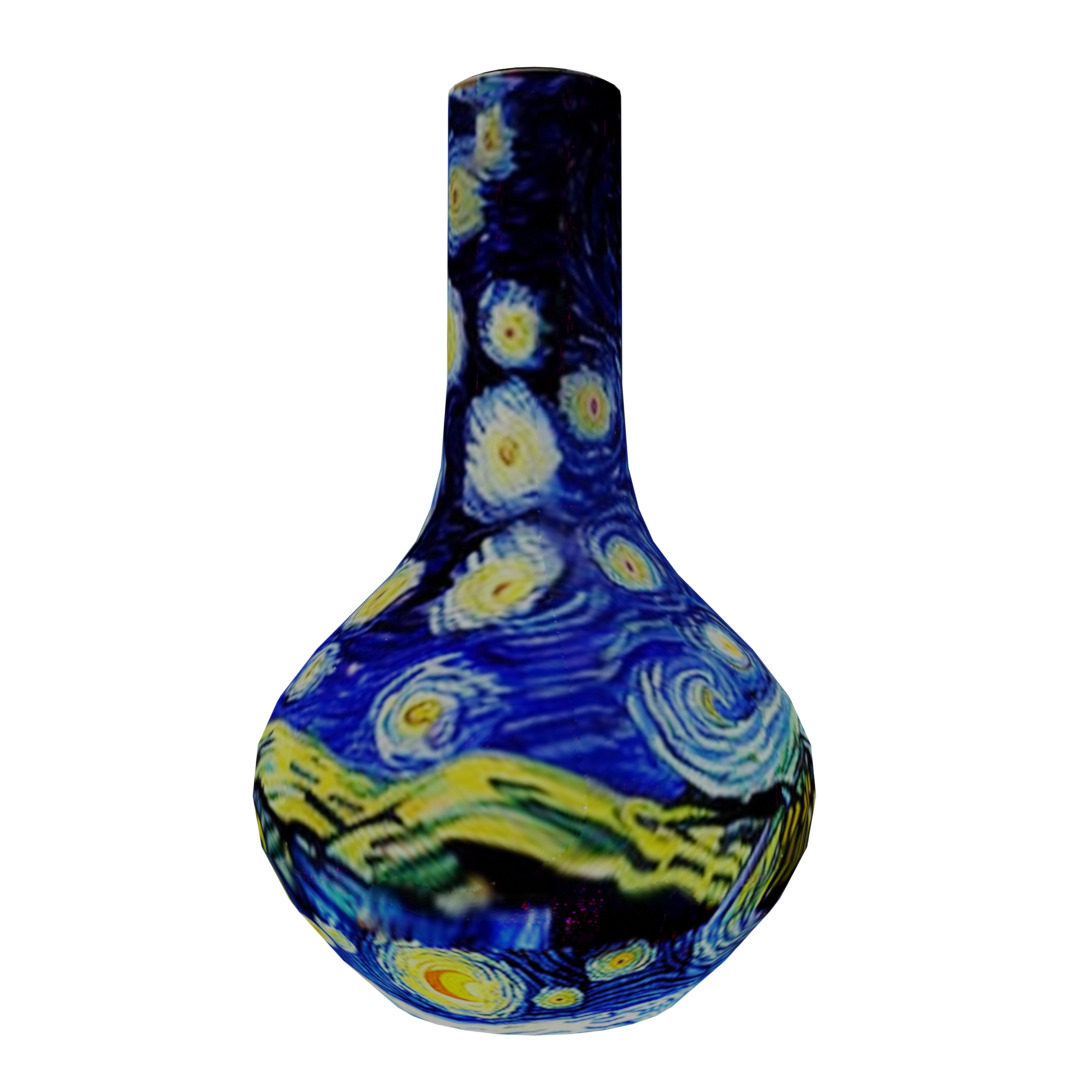} &
        \includegraphics[width=0.15\linewidth]{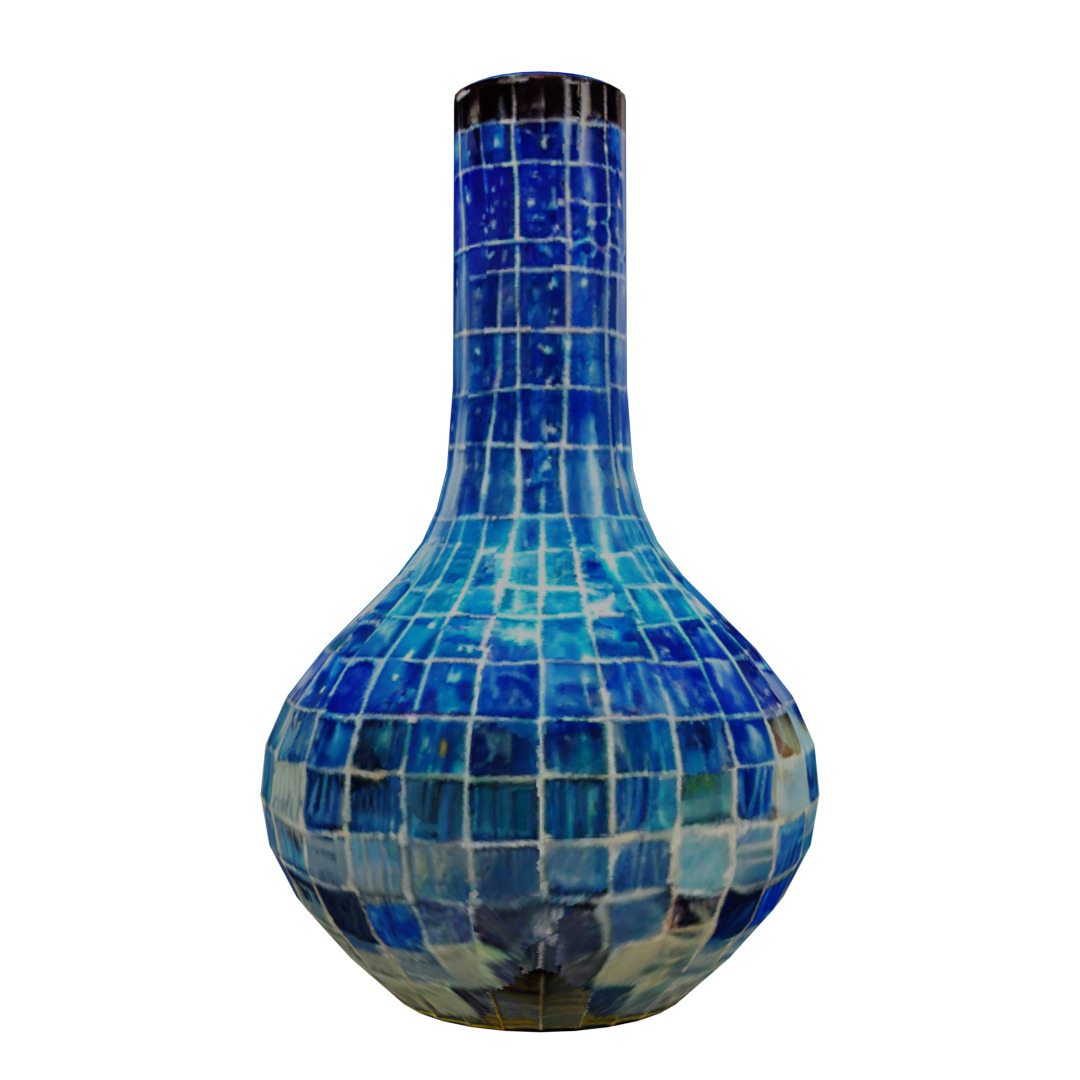} &
        \includegraphics[width=0.15\linewidth]{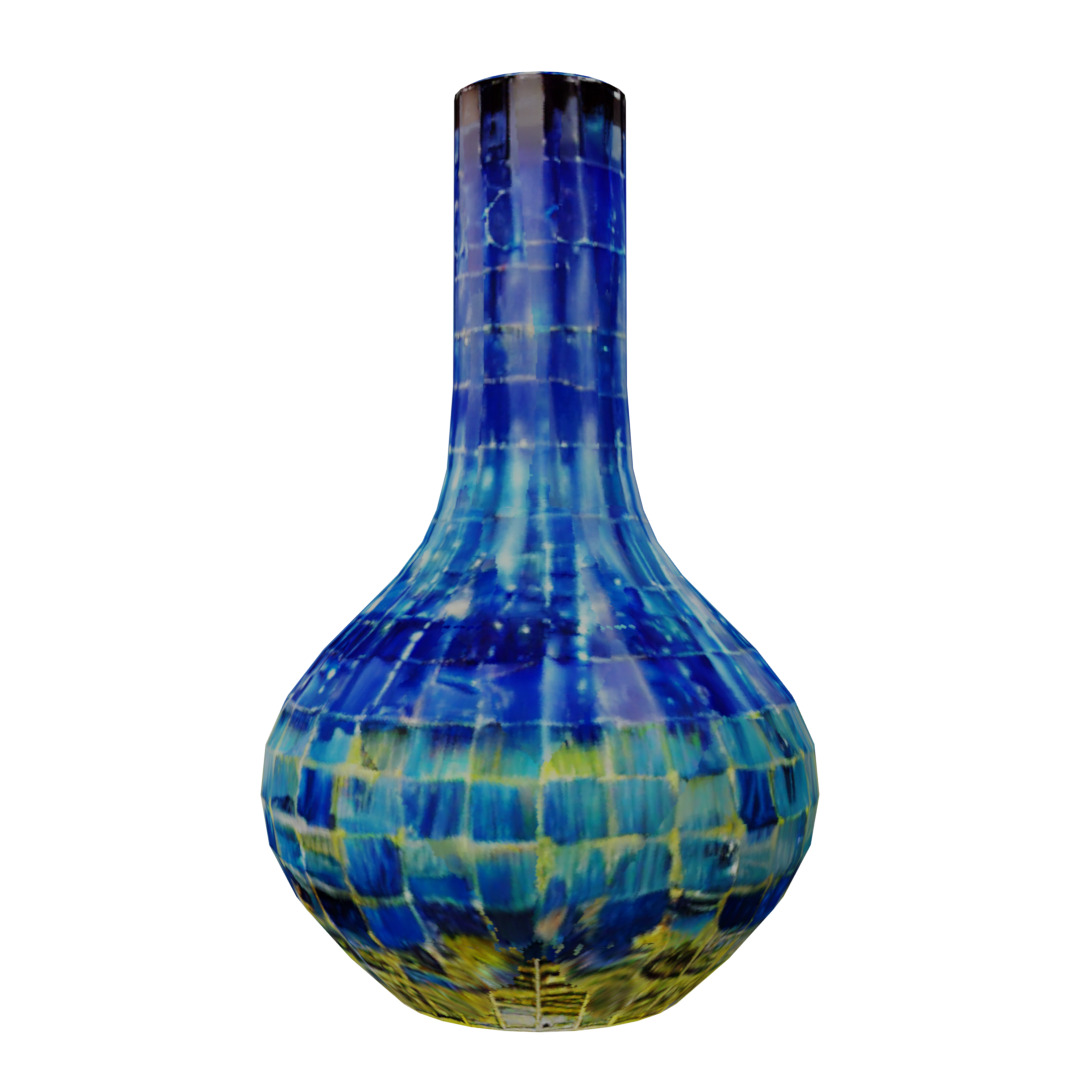} \\
        \includegraphics[width=0.15\linewidth]{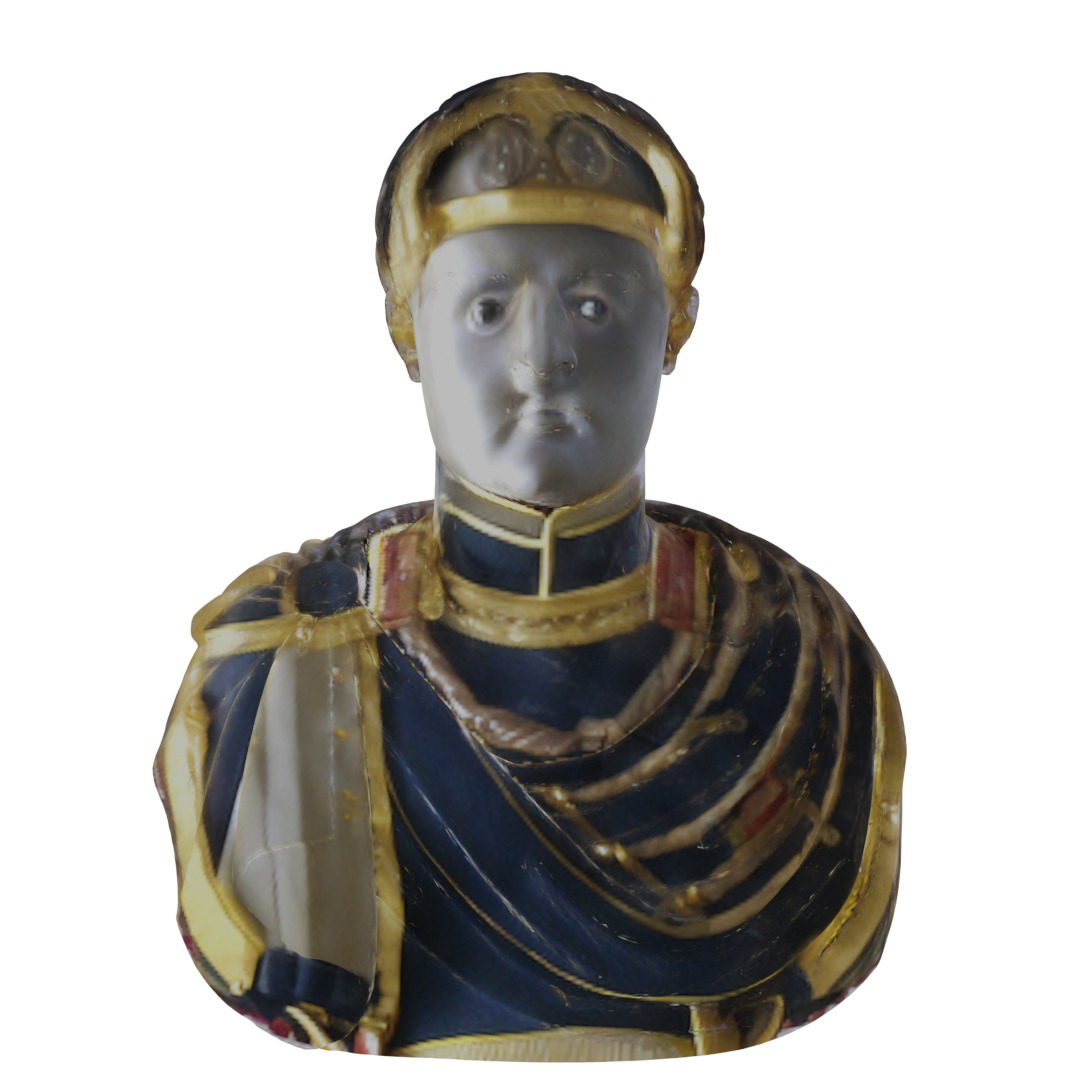} &
        \includegraphics[width=0.15\linewidth]{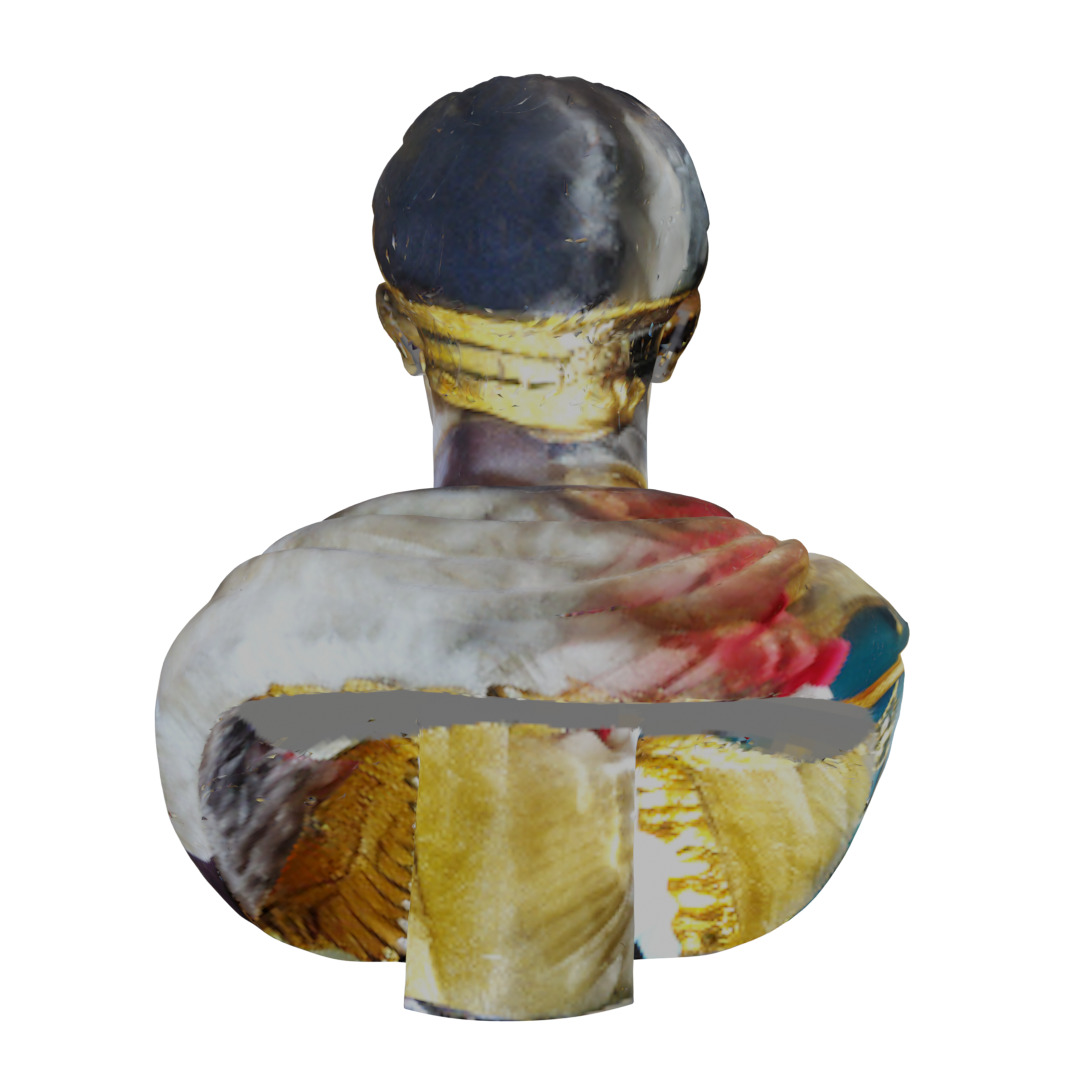} &
        \includegraphics[width=0.15\linewidth]{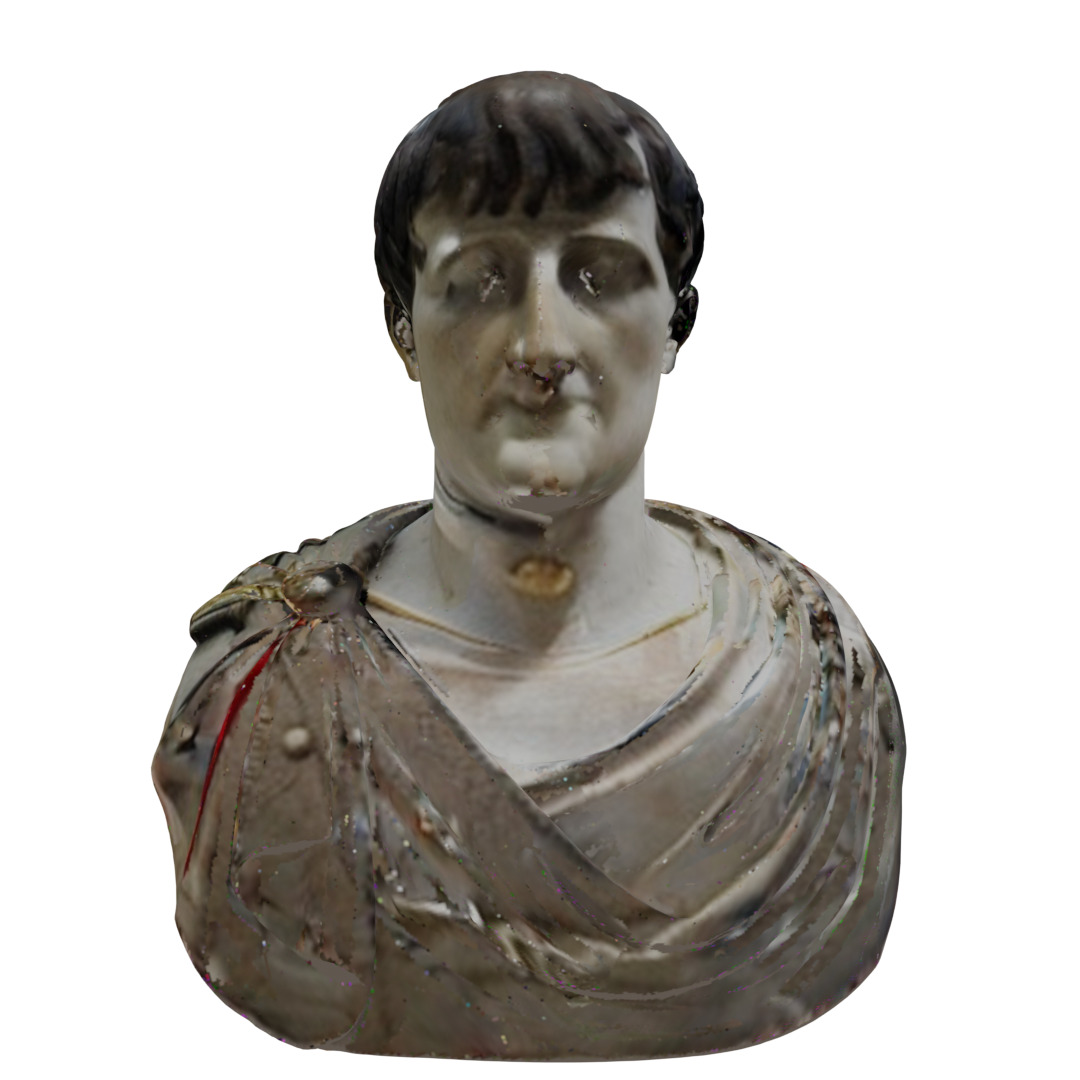} &
        \includegraphics[width=0.15\linewidth]{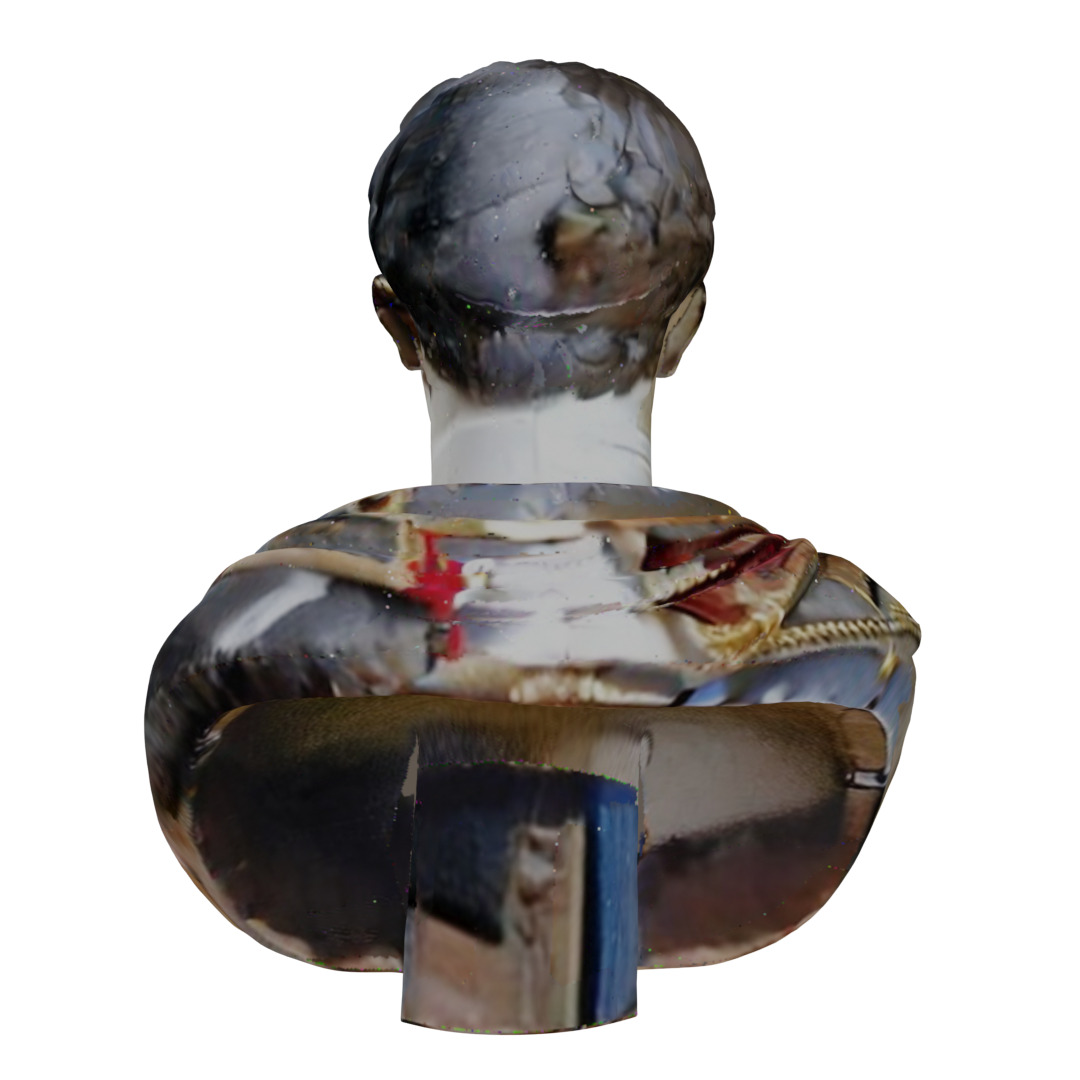} &
        \includegraphics[width=0.15\linewidth]{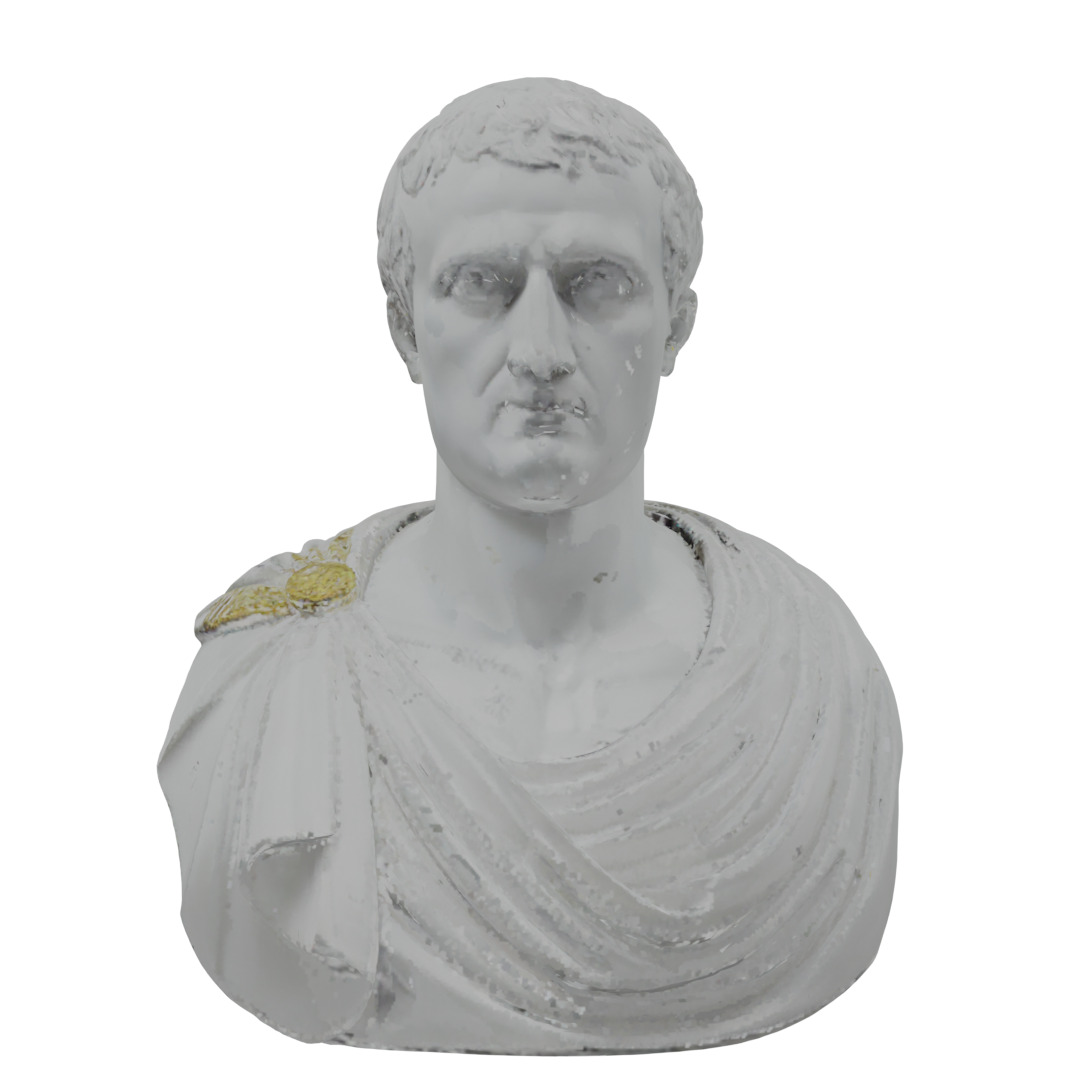} &
        \includegraphics[width=0.15\linewidth]{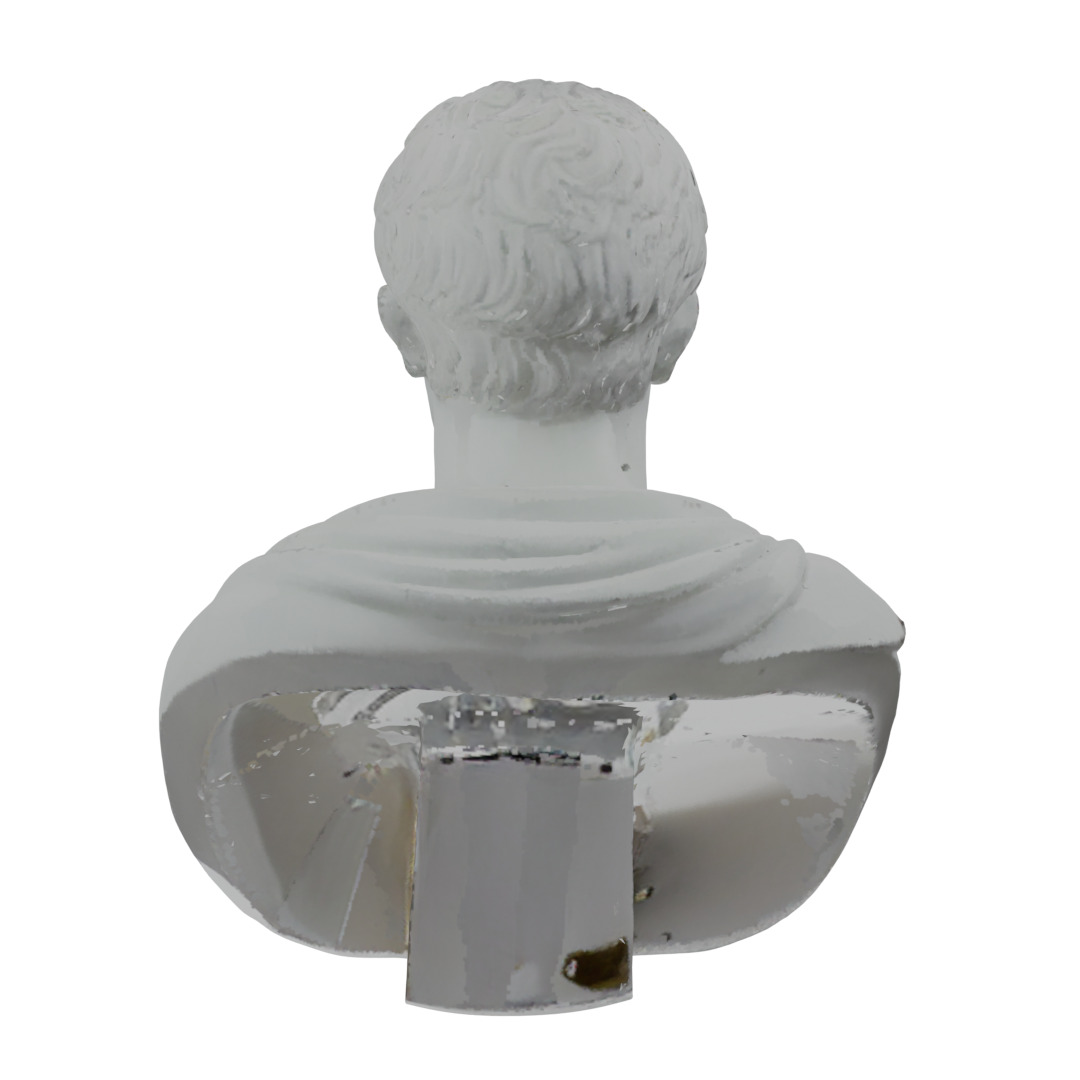} \\
    \end{tabular}
    
    \caption{Comparison of our work to TEXTure~\cite{TEXTure} and Text2Tex~\cite{text2tex} using non-cherry picked examples on prompts ``Jupiter'', ``Starry Night Van Gogh Vase'', and ``Napoleon''. Our approach reduces the number of noticeable seams between different views that were used when generating. Our approach also has more consistent lighting since it is a single diffusion process, and reduces stretching on the produced texture. In each image we show a front and back view of the mesh. We note that the quality of each may vary significantly depending on the random seed. For all experiments, we fix the seed for all approaches.}
    \label{fig:qual_compare}
\end{figure*}

We compare our approach to TEXTure~\cite{TEXTure} on a number of meshes in Fig.~\ref{fig:qual_compare}. We use the official TEXTure~\cite{TEXTure} and Text2Tex~\cite{text2tex} codebases to perform our comparisons. We note that TEXTure's implementations suffers from salt-and-pepper noise due to their backprojection approach, and does not completely fill visible regions with texture. For example, on the sphere textured with ``jupiter'', there is a patch that is untextured directly visible from the front view. Text2Tex also produces noticeable seams between different textured regions. Our approach blends all views, so it more smoothly transitions between views. We note that it is possible to create geometry that will have untextured regions for all works, but find that for TEXTure even a simple input such as a sphere has untextured regions. We also note that while our work and TEXTure use Stable Diffusion 2.1 with Depth, Text2Tex utilizes Control Net~\cite{controlnet} with Depth, which may partially explain the difference in results.

For the ``Starry Night Van Gogh Vase'', our result and TEXTure's results are good, but note that there is significant warping at the bottom of the vase in TEXTure's front view, whereas ours more naturally curves around the bottom. Text2Tex is sensitive to sharp normal changes, and thus produces a number of artifacts on the vase, such as edges between mesh faces, and does not match the prompt closely.

For the Napoleon model, the front of TEXTure does not look good, as it is all a single muted color. The back of both our approach and TEXTure both exhibit some artifacts, but ours has a consistent color scheme, and maintains the headband from the front to the back.

The run-time for TEXTure and our approach is about 5 minutes, and Text2Tex with 20 update steps takes about 20 minutes. All these approaches stand in contrast to DreamFusion~\cite{poole2022dreamfusion} or Fantasia 3D~\cite{fantasia3D}, which may take hours and require multiple GPUs and hours of optimization. The primary costs of our approach is GAN inversion, which takes about 4 minutes, and the diffusion process, which takes about 40 seconds, with backprojection taking 20 seconds.

\subsection{Ablations}

We ablate multiple hyperparameters of our method. Depending on UV parameterization, and the specific mesh, we find that tuning these parameters can produce much higher quality textures. A quantitative comparison is shown in Fig.~\ref{fig:clip_ablations}, and we discuss each parameter below.

\paragraph{Guidance Scale}
\label{sec:ablate_guidance_scale}
One issue is that the diffusion model sometimes produces results that are too varied. Like Score Distillation Sampling~\cite{poole2022dreamfusion}, we try increasing the guidance scale. This somewhat mitigates inconsistent output from the diffusion model, and reduces blurring in the final result. We test our approach with the guidance scale of 7.5, 20 and 32, and find that if 7.5 is blurry, 20 and 32 will lead to higher consistency at the cost of over-saturated colors, which is a known issue with diffusion models. We visualize an example in the Appendix.

\label{sec:ablate_texture_size}
\paragraph{Texture Size}
Since the UV parameterization is not guaranteed to effectively use the texture space uniformly or efficiently, the latent texture's size may change the quality of the final output. To demonstrate the importance of selecting a good latent texture size, we perform texture sampling on a single cube model which only uses two-thirds of the texture space, and each face uses one-ninth of the texture space, shown in the Appendix.
When observing a single face, it may have significantly fewer pixels than the 64x64 images Stable Diffusion requires, leading to poor results. We demonstrate that this is only present when texture size is too low, and increasing it looks normal. We also demonstrate that if each texture has too many texels, each view will no longer correspond with any other view.

\label{sec:ablate_camera_views}
\paragraph{Camera Views} We also ablate using multiple different camera views. For some examples it is not clear if 8 camera views is reasonable, so we increase the number of cameras during the MultiDiffusion and backprojection step. We visualize one model with 8, 16, and 32 camera views in the Appendix 
When increasing the number of cameras, it reduces high-frequency detail but removes seams between views.

\label{sec:ablate_sh_parameters}
\paragraph{Spherical Harmonic Selection} Spherical harmonics can also improve the quality for some models. We visualize the difference in quality for some models in the Appendix. 
We find that for some models it can improve the performance, but for others it may not, such as when the texture size already makes the model have per-view independence. We find that increasing the order of spherical harmonics can preserve more high-frequency detail for some meshes

\label{sec:ablate_flat_camera_views}
\paragraph{Flat Camera Sampling} Finally, we also test using cameras sampled entirely on the XZ plane. We find that for some models, it produces more coherent output, as it reduces stretching due to camera elevation. On the other hand, using flat cameras leads to more areas being untextured. For some models, this is not problematic, but varies on the specific model being textured. We demonstrate the effect of using flat cameras in the Appendix. 

\section{Discussion}

\paragraph{Limitations} Our approach, like TEXTure~\cite{TEXTure} still may suffer from the multi-Janus problem, which is when multiple faces are generated from different views. We consider it outside the scope of this work, and can be better handled by works such as ~\cite{consistency3Dgen}. Our approach also suffers from the same issue as TEXTure~\cite{TEXTure} where sometimes the diffusion model may entirely ignore the given depth, leading to poor texturing results. This often can be mitigated simply by choosing a different seed. Finally, we note that if visual detail is provided by the geometry itself, then our approach may not follow those cues.

\paragraph{Text Prompt Imprecision}
\label{sec:precise_text_prompts}
We find that a key issues with texturing a mesh from a text prompt is that the problem is ill-posed. A text prompt cannot precisely specify many details, and because of that ambiguity it is difficult to produce consistent multi-view images. One example from TEXTure~\cite{TEXTure}'s repository is ``next-gen Nascar'', for texturing a car, but this prompt is meaningless, as ``Nascar'' doesn't refer to a car, but refers to the race and brand, and it is not clear what ``next-gen'' adds. Since the prompt itself is nonsensical, there is not a clear output. Instead, works like Zero-1 to 3~\cite{zero1to3} or SyncDreamer~\cite{syncdreamer} that use an image to produce 3D views of a single object specify a more exact input, and should suffer from less ambiguity.

\section{Conclusion}

We extend MultiDiffusion~\cite{multidiffusion} to mesh texturing, retaining expressiveness from 2D diffusion models. Our approach is the same speed as TEXTure~\cite{TEXTure}, and has higher consistency. Our approach is fairly robust to a variety of prompts for a fixed mesh, and is able to handle arbitrary camera positions so can cover the entire mesh surface. We hope that this will enable games to generate a variety of assets cheaply for their game.

{\small
\bibliographystyle{ieee_fullname}
\bibliography{egbib}
}

\appendix
\section{Additional Results}

In this supplementary document, we provide ablations, a summary of differences between concurrent work and ours, and additional results from our work.

\begin{figure}[ht]
    \begin{tabular}{l c c c}
        & \multicolumn{3}{c}{Latent Texture Size} \\
        UV of \put(-10,-10){\includegraphics[width=0.2\linewidth]{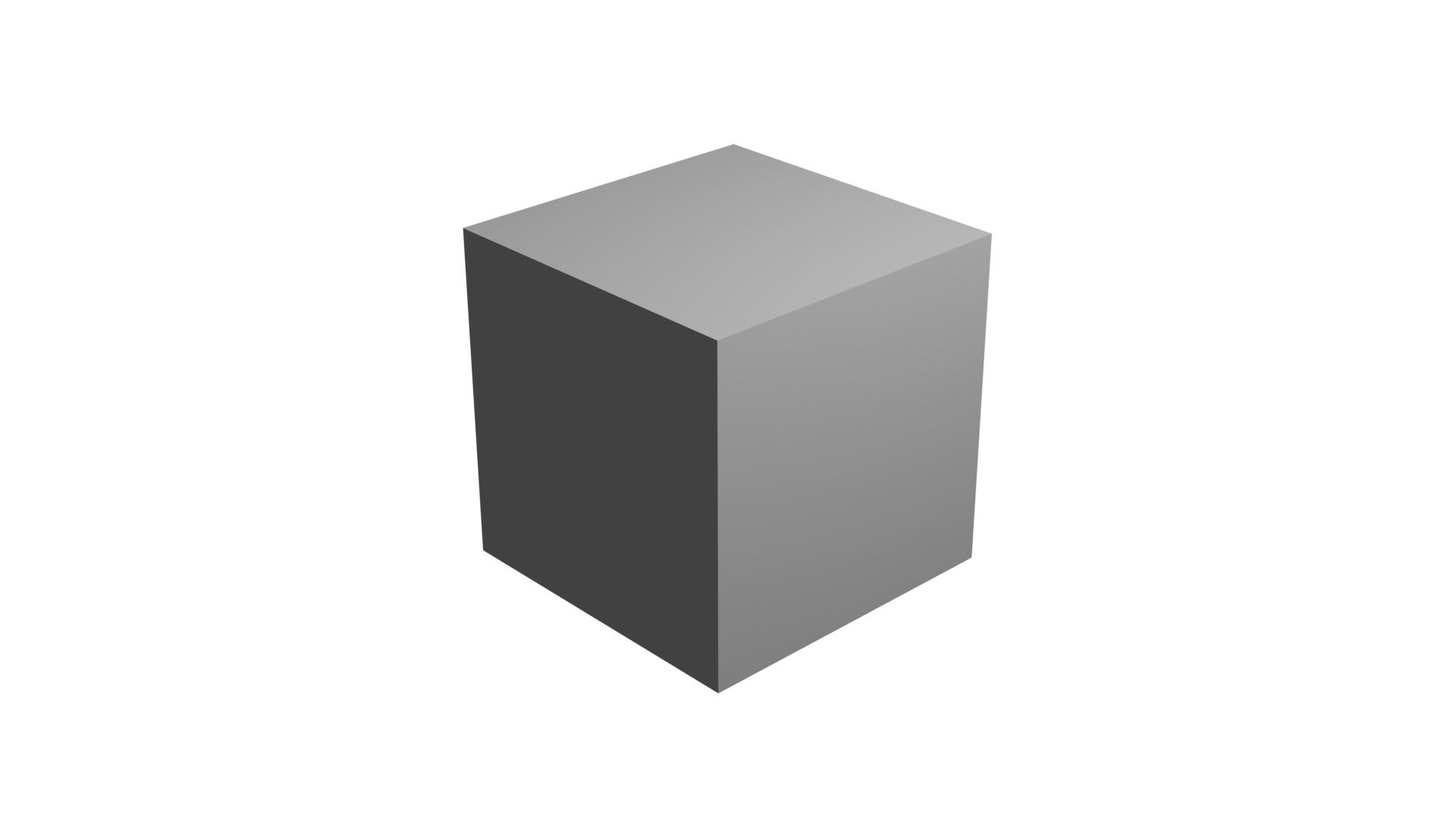}} & $128\times128$ & $196\times196$ & $256\times256$  \\
        \includegraphics[width=0.2\linewidth]{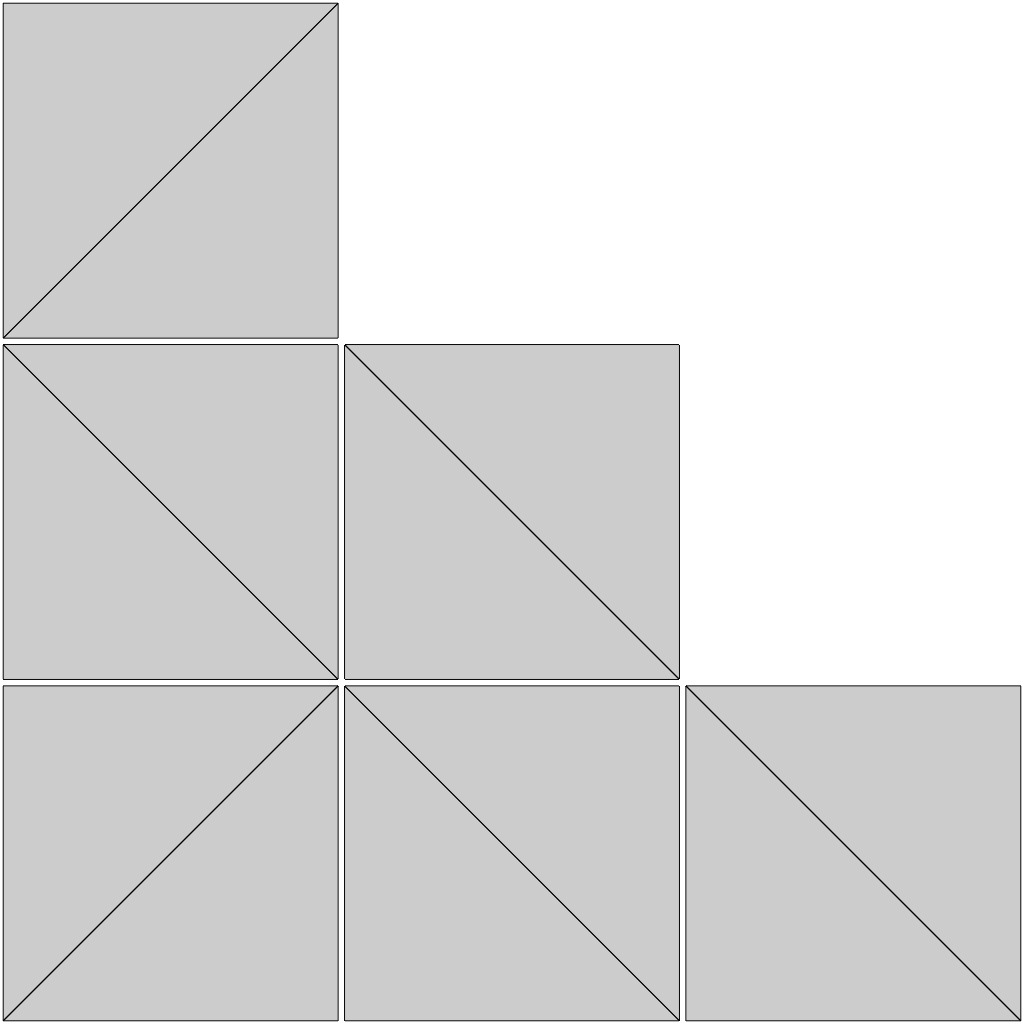} &
        \includegraphics[width=0.2\linewidth]{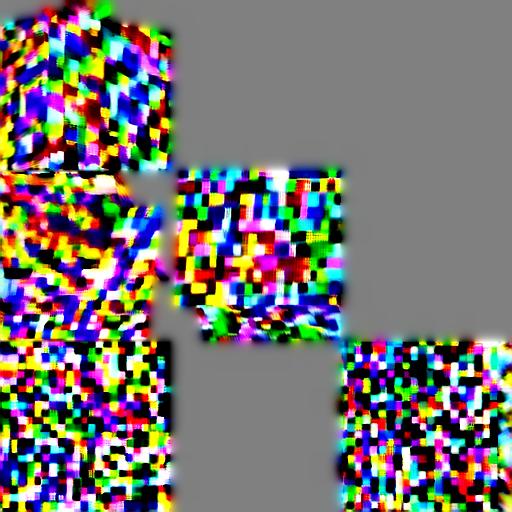} &
        \includegraphics[width=0.2\linewidth]{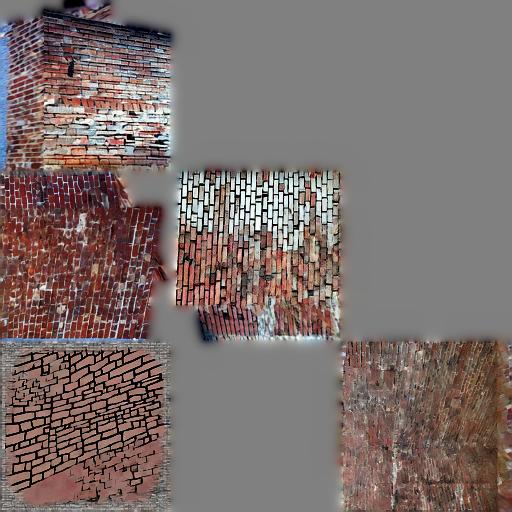} &
        \includegraphics[width=0.2\linewidth]{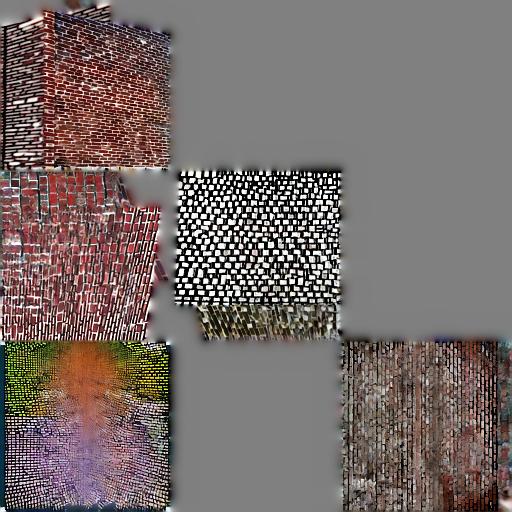} \\
        & \multicolumn{3}{c}{Generated Texture for Prompt ``Bricks''}
    \end{tabular}
    \caption{Latent UV parameterization ablation. Selecting a small texture size leads to poor final results. For this model, a $128\times128$ texture map leads to a degenerate output. Increasing the texture size leads to better output. Note the bottom face is gray as only the upper hemisphere is optimized.}
        \label{fig:ablate_texture_size}
\end{figure}

\begin{figure}[ht]
    \centering
    \begin{tabular}{c c c}
    \includegraphics[width=0.3\linewidth]{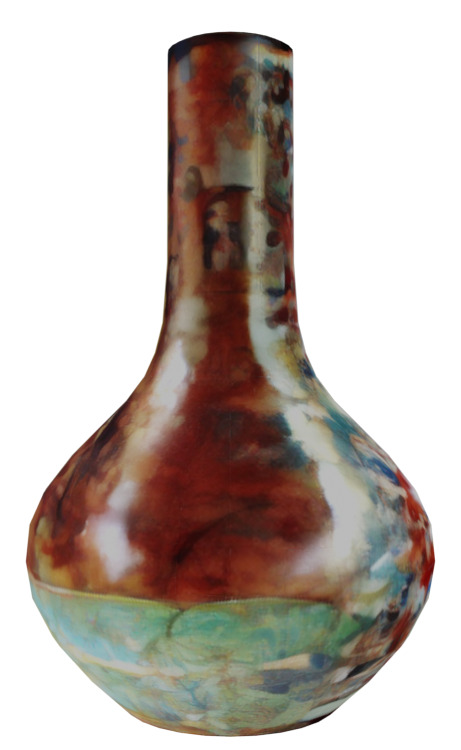} &
    \includegraphics[width=0.3\linewidth]{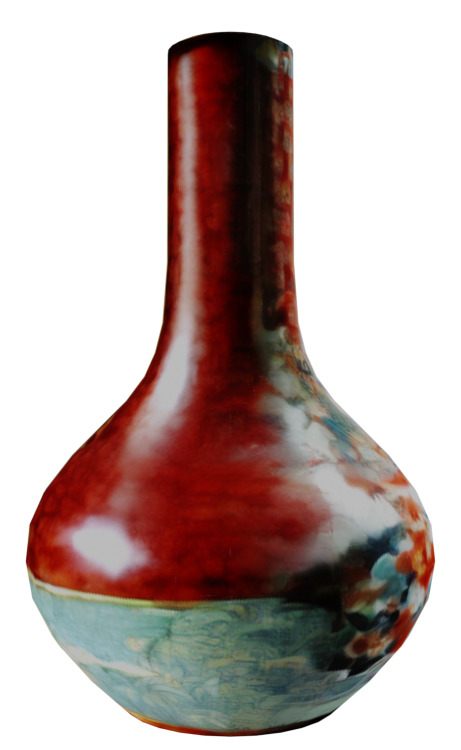} &
    \includegraphics[width=0.3\linewidth]{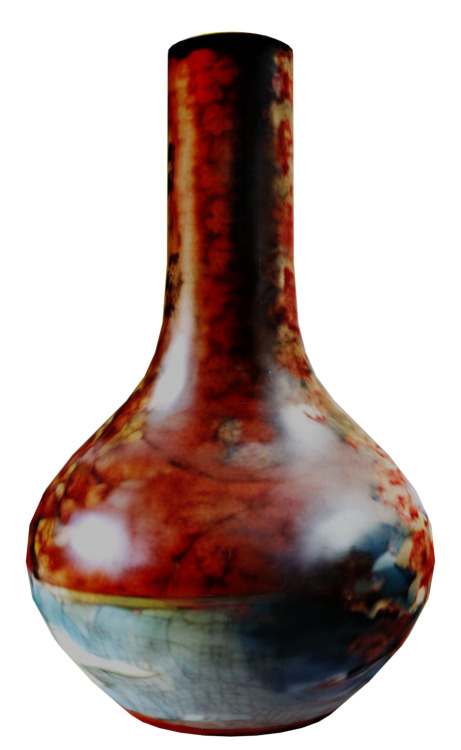} \\
    7.5 & 20 & 32 \\
    \multicolumn{3}{c}{Diffusion Guidance Scale} \\
    \end{tabular}
    \caption{Example of varying guidance scale. Increasing guidance scale leads to oversaturation, but can increase the sharpness of some features in our output. The prompt is ``Chinese Vase'', with 8 views and latent texture size of 128.}
    \label{fig:ablate_guidance_scale}
\end{figure}

\begin{figure}[ht]
    \centering
    \begin{tabular}{c c c}
         8 Views & 16 Views & 32 Views  \\
         \includegraphics[width=0.3\linewidth]{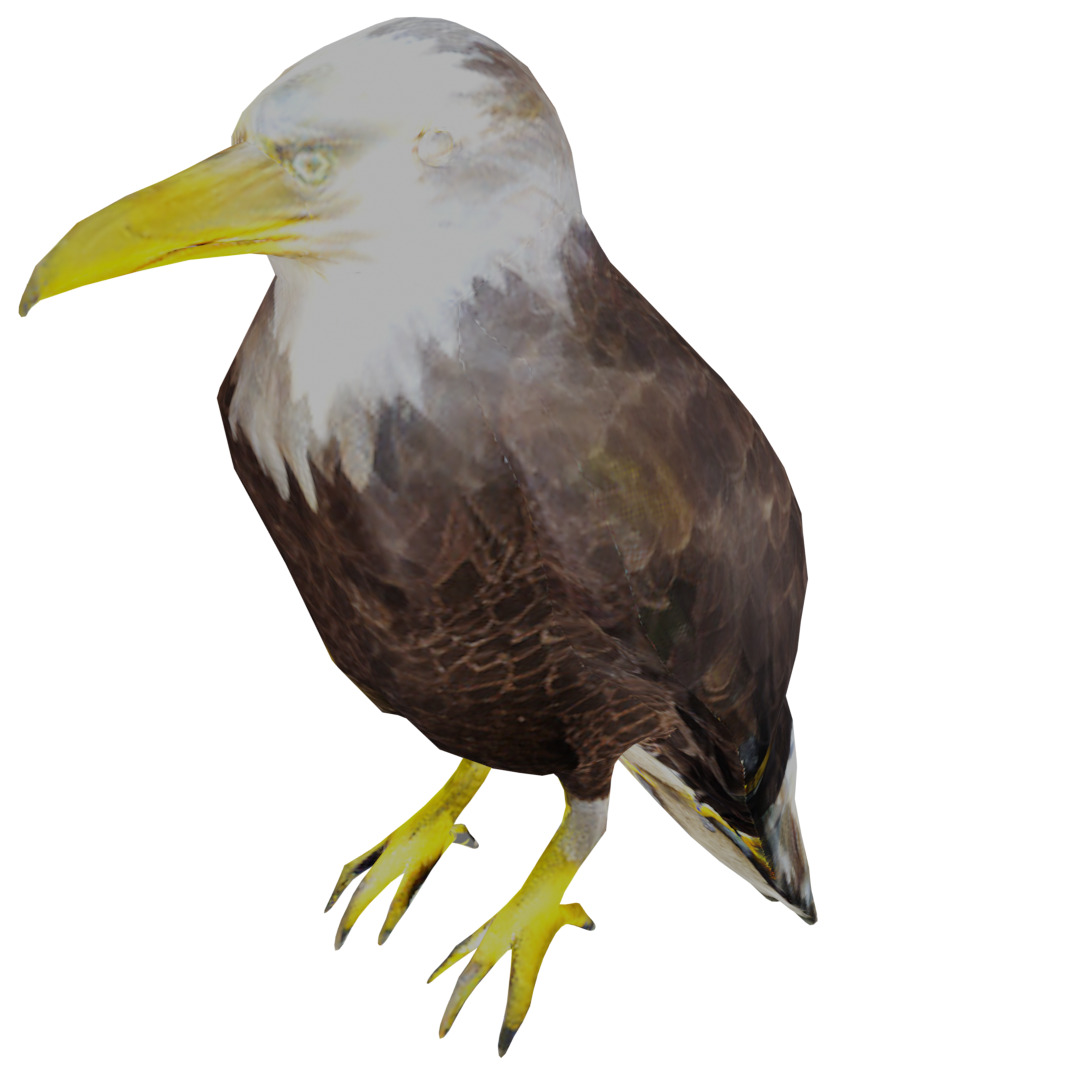} &
         \includegraphics[width=0.3\linewidth]{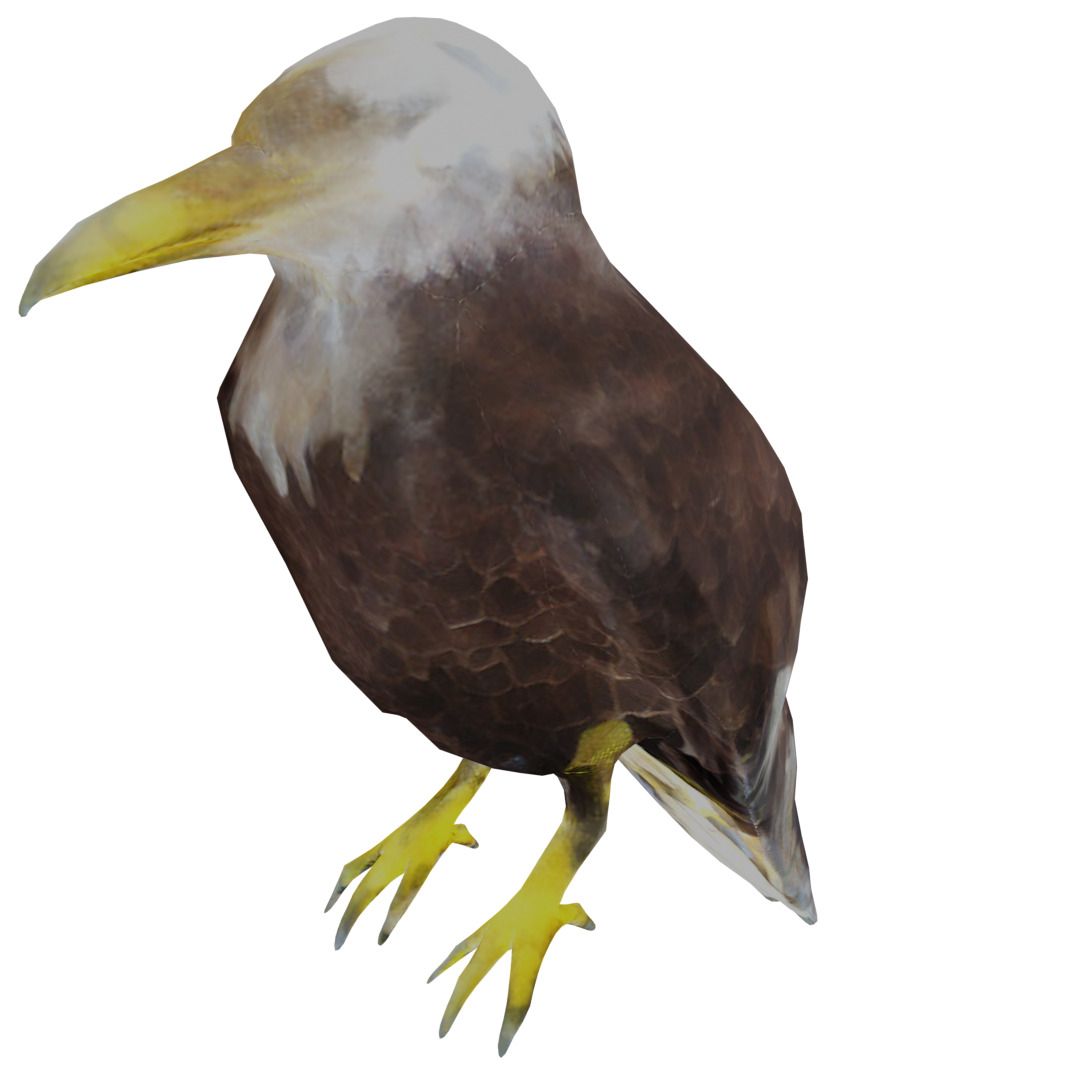} &
         \includegraphics[width=0.3\linewidth]{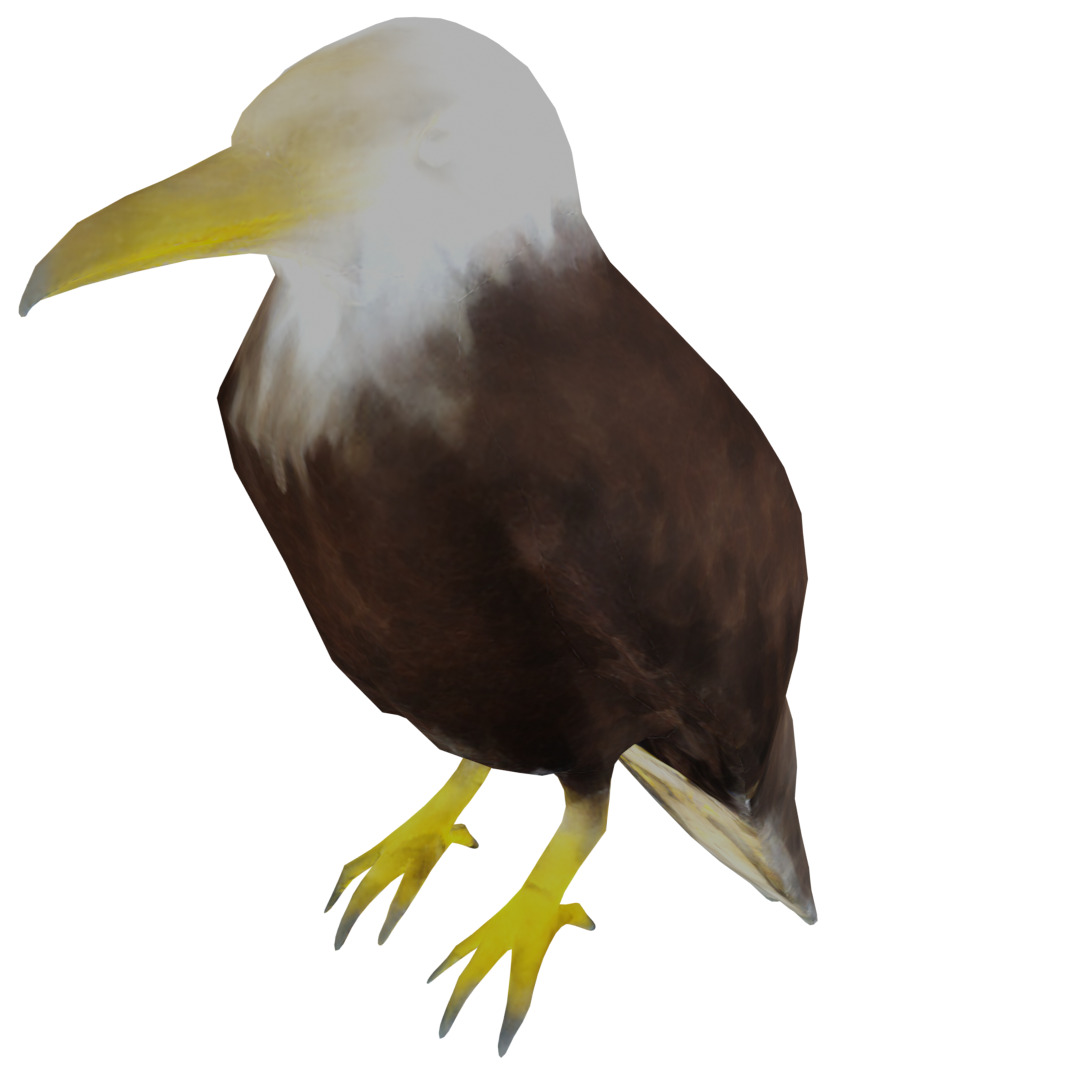} \\
    \end{tabular}
    \caption{Ablation of number of cameras used during diffusion. The mesh is a crow, and the prompt ``Bald Eagle'' was used. We find that as more camera views are added, features become over-smoothed, but it still maintains its overall appearance. Specifically, the eagle's feathers become blurred, and the eye becomes smoothed over.}
    \label{fig:ablate-views}
\end{figure}

\begin{figure}[ht]
    \centering
    \begin{tabular}{c c}
         \includegraphics[width=0.4\linewidth]{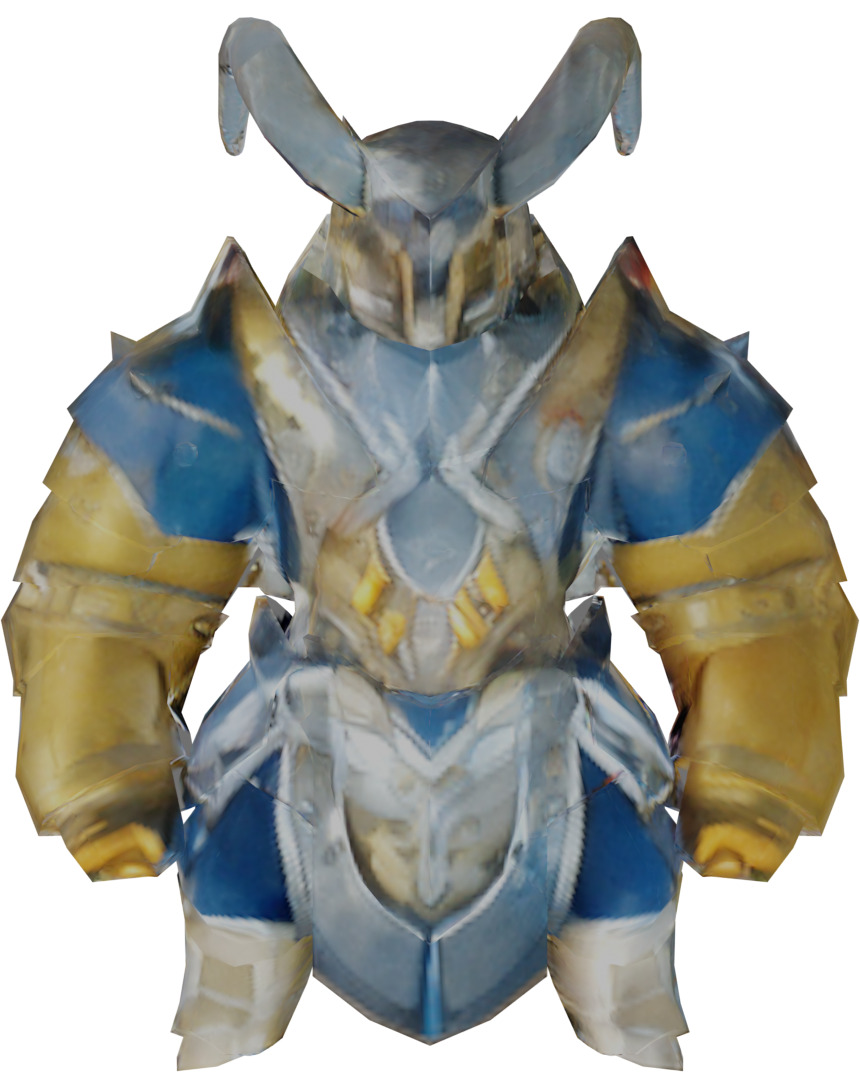} &
         \includegraphics[width=0.4\linewidth]{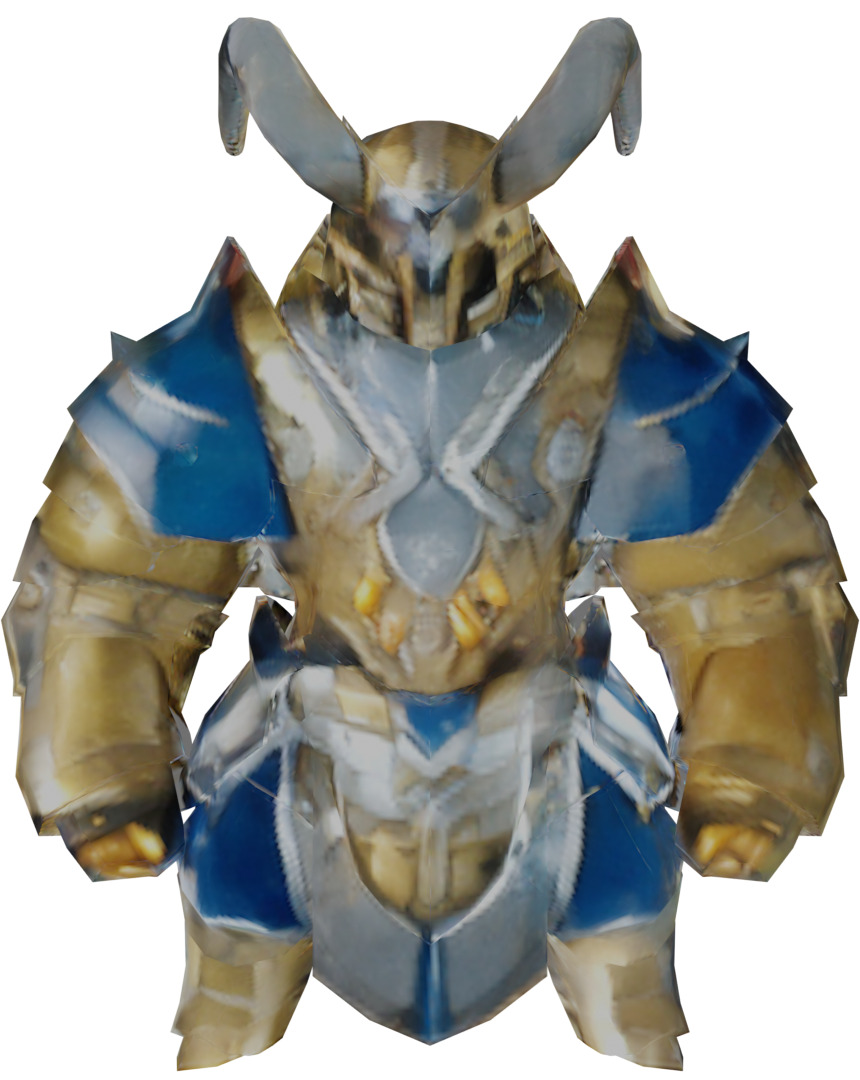} \\
        Hemisphere Sampling & XZ Plane Sampling  \\
    \end{tabular}
    \caption{We optimize the same model with cameras sampled on the hemisphere, and cameras sampled in a circle on the XZ plane. For some models this leads to better results, characterized by sharper textures such as on the face and rest of body of this ``Paladin'' model but more untextured regions, such as the bottom view.}
    \label{fig:ablate_flat_cameras}
\end{figure}

\begin{figure}[ht]
    \centering
    \begin{tabular}{c c}
         \includegraphics[width=0.3\linewidth]{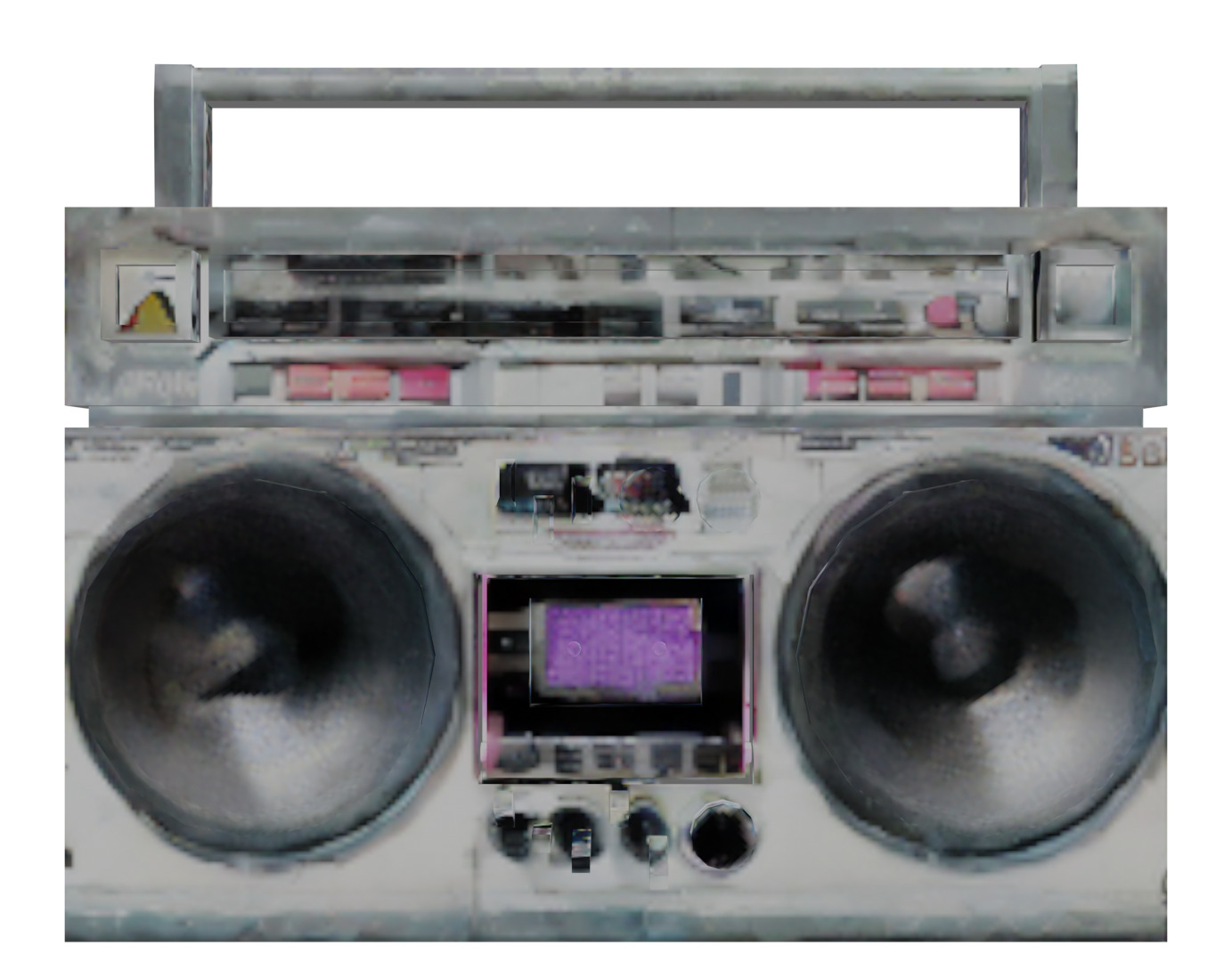}
         \includegraphics[width=0.15\linewidth]{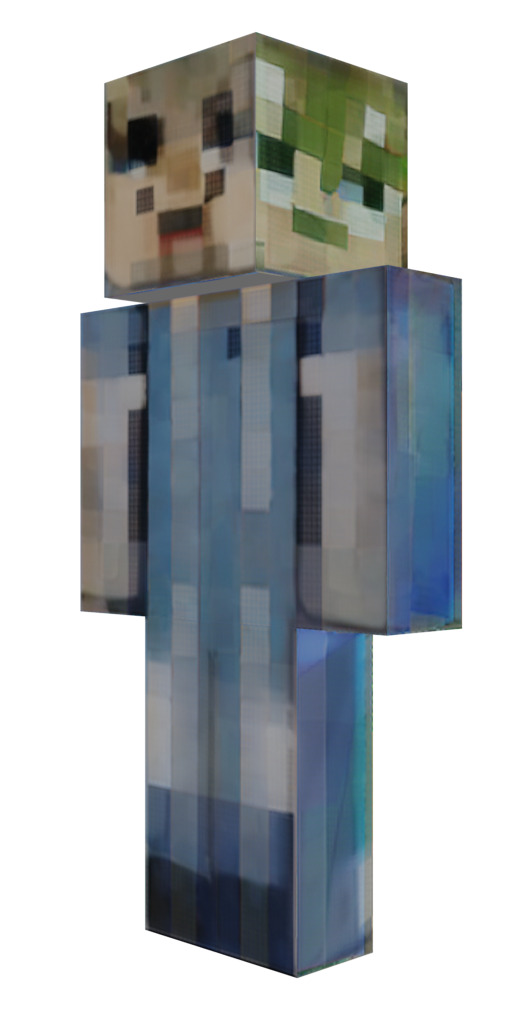}
         &
         \includegraphics[width=0.3\linewidth]{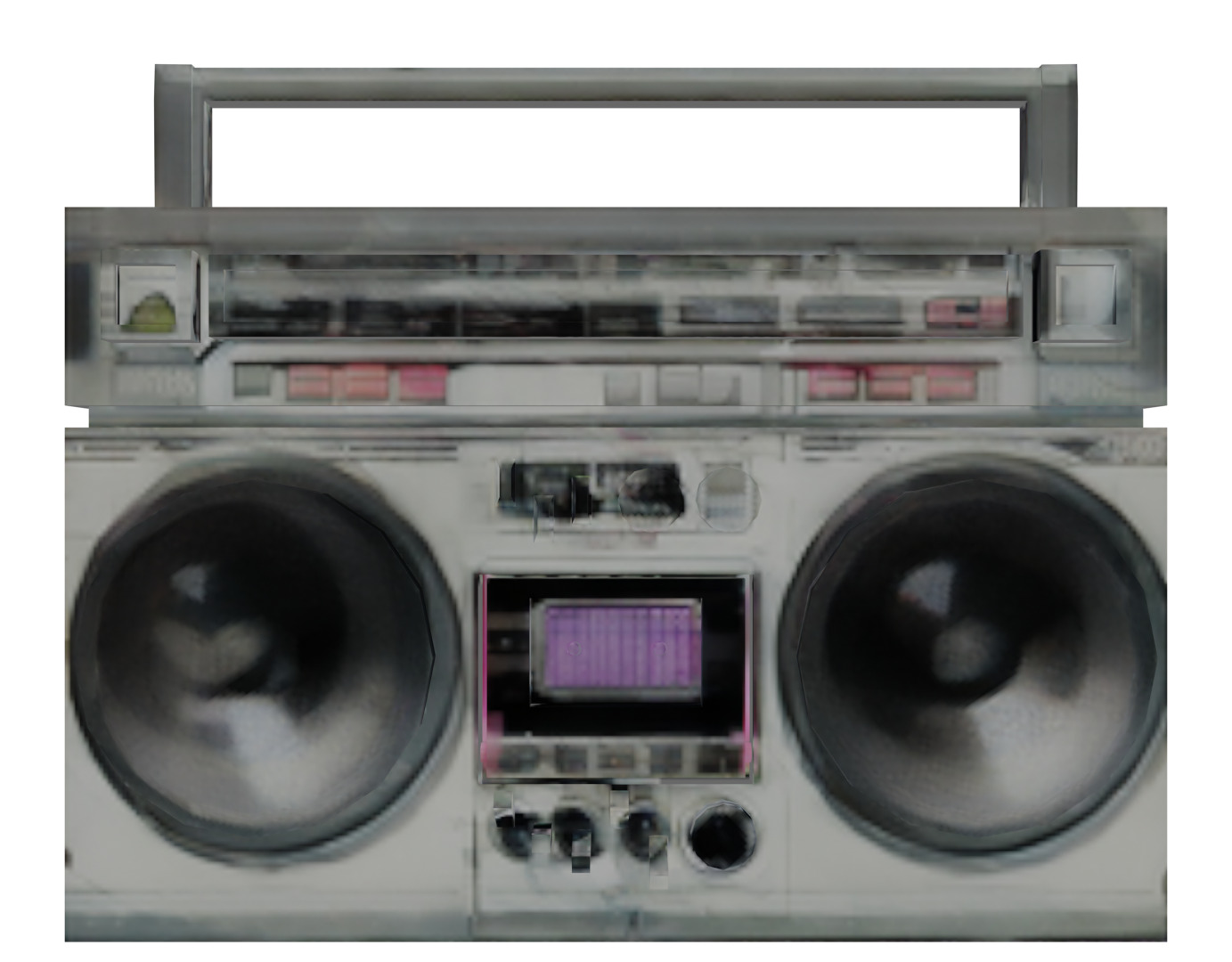}
         \includegraphics[width=0.15\linewidth]{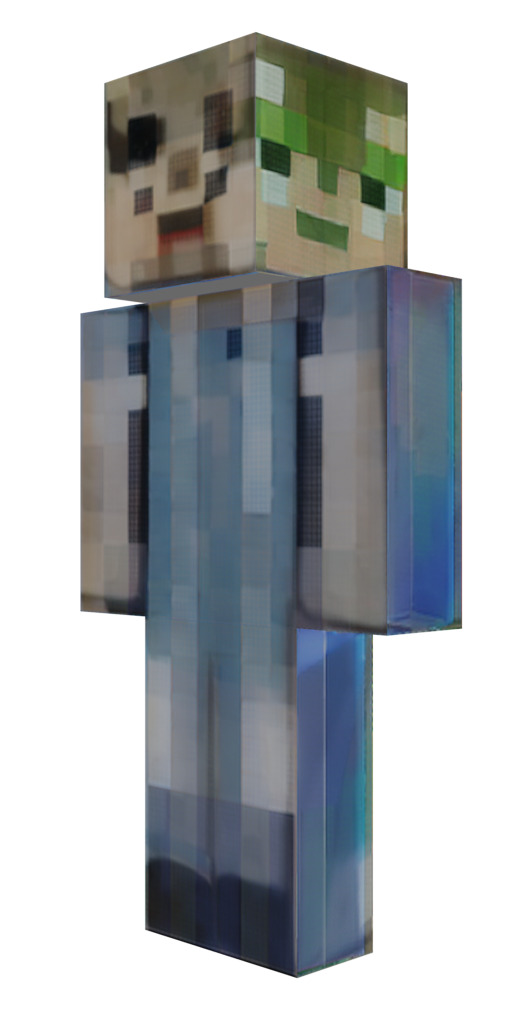} \\
         SH Order 0 & SH Order 1  \\
    \end{tabular}
    \caption{We compare two models with the prompts ``90s boombox'' and ``minecraft steve'', using different orders of Spherical Harmonics. While there is a not a huge difference between the two approaches, SH order 1 can have textures with less noise, such as on both loudspeakers on the boombox and the cassette tape in the center. It can also preserve sharper features, such as the faces on Steve.}
    \label{fig:ablate_sh}
\end{figure}

\begin{figure*}
\centering
\begin{tabular}{c c c c}
    \multicolumn{2}{c}{Ours} & \multicolumn{2}{c}{Online Tool} \\
    \includegraphics[width=0.2\linewidth]{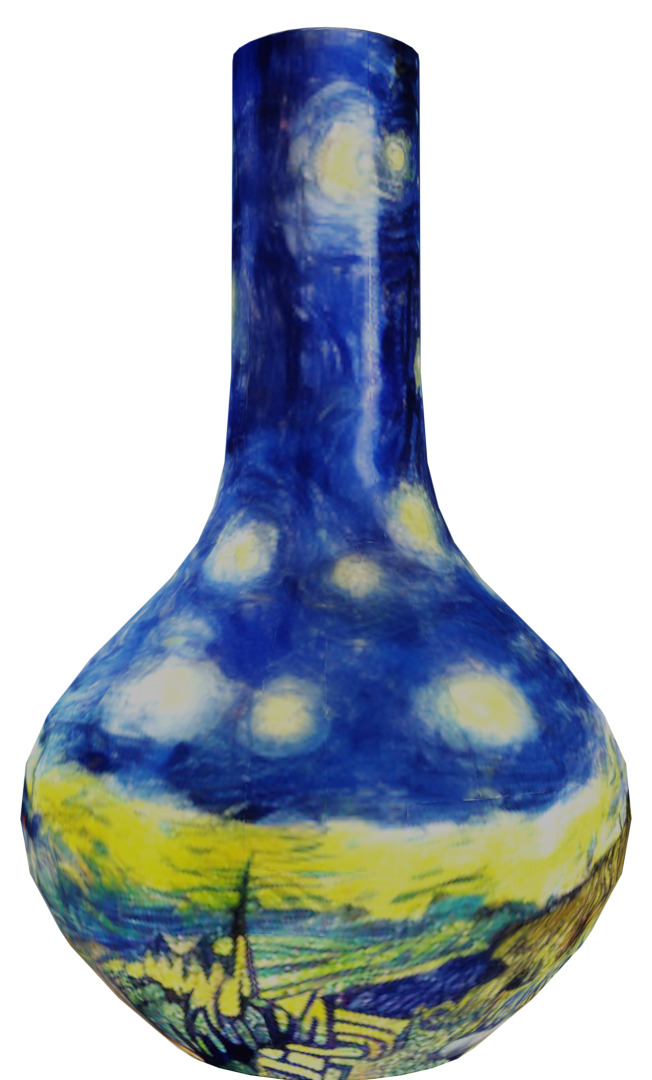} &
    \includegraphics[width=0.2\linewidth]{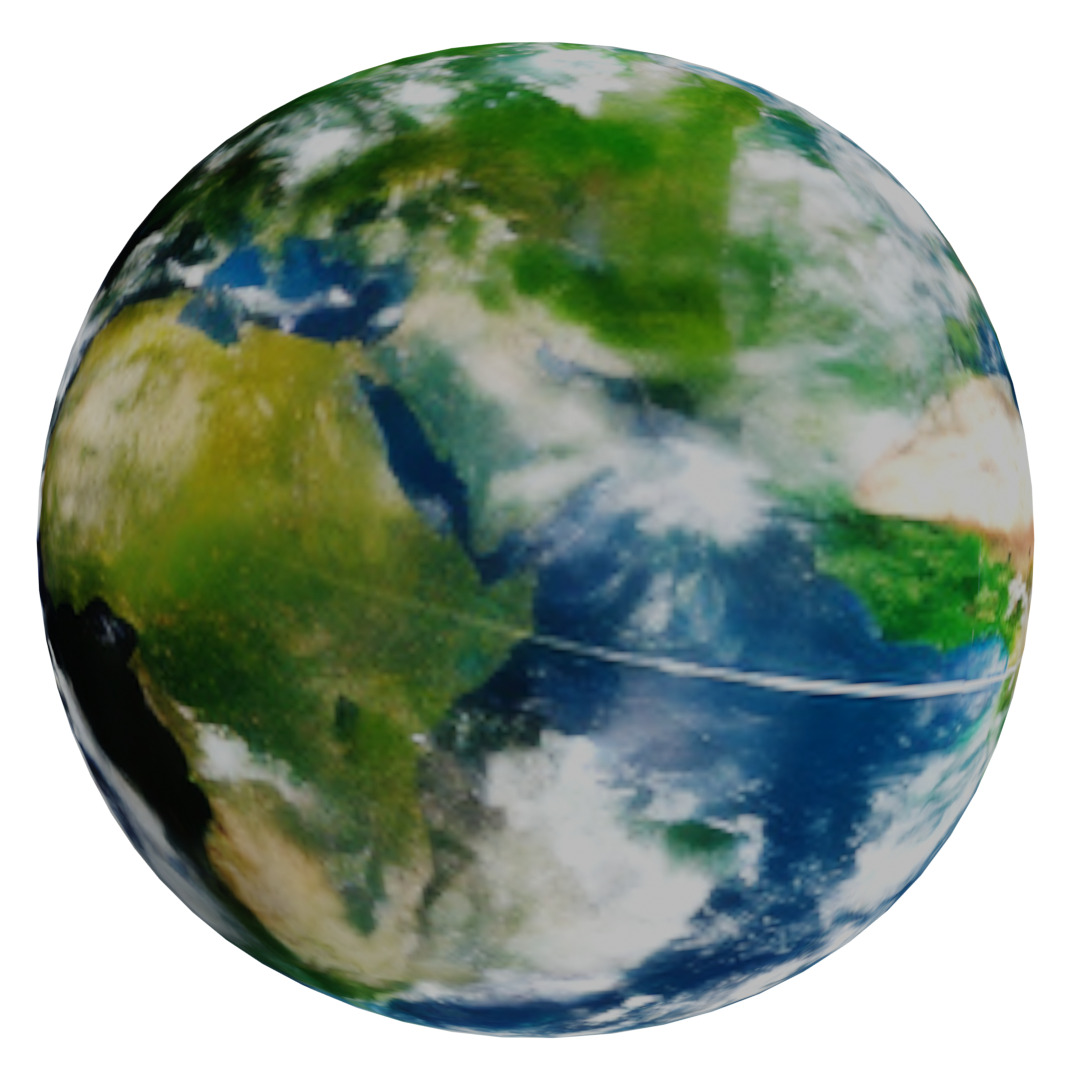} &
    \includegraphics[width=0.2\linewidth]{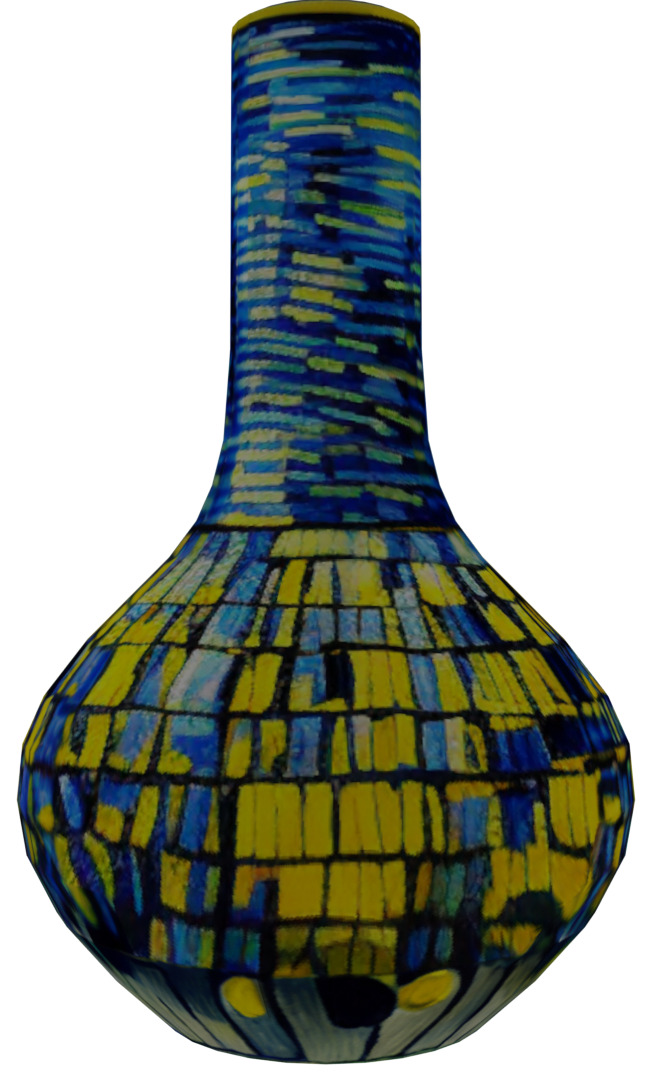} & \includegraphics[width=0.2\linewidth]{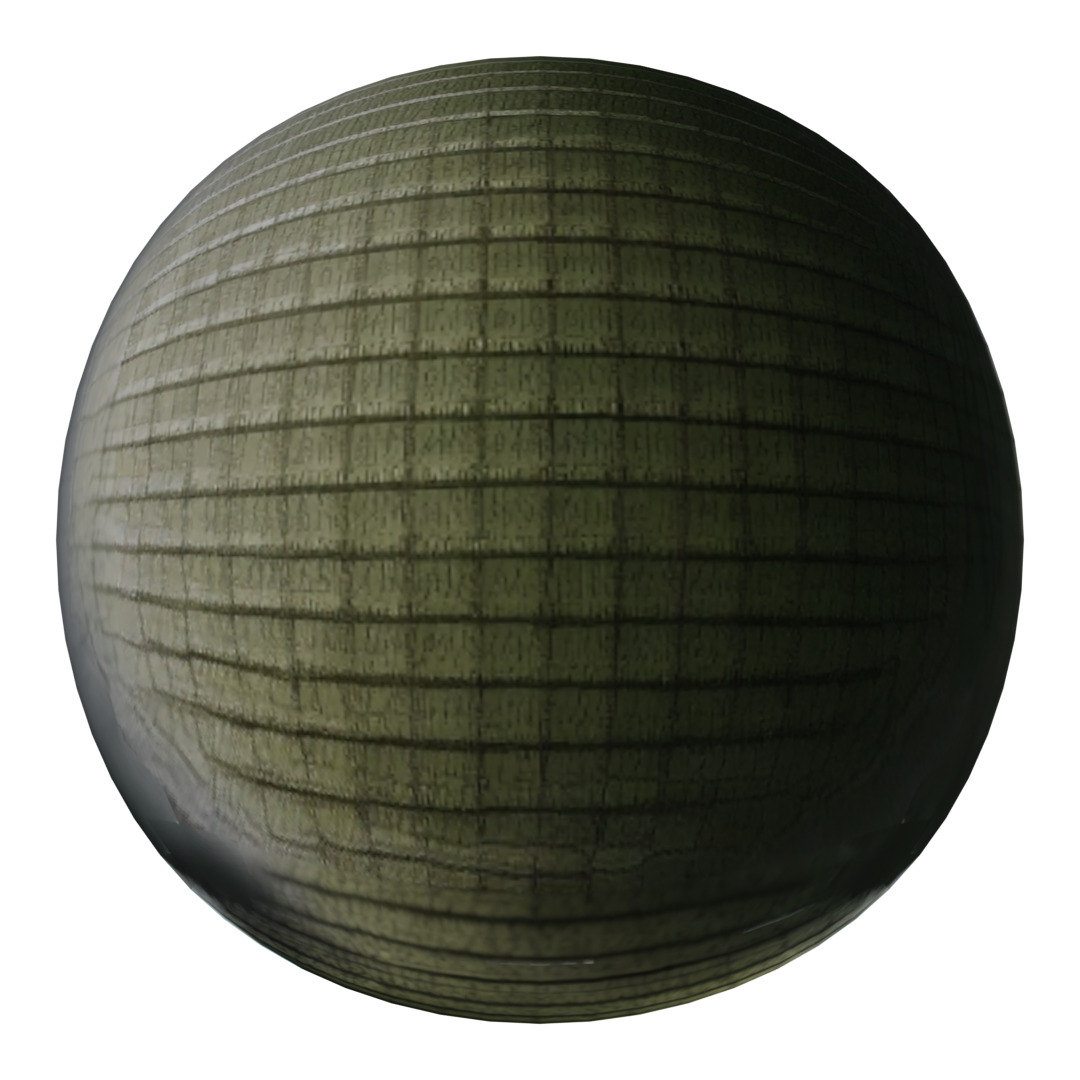} \\
    Starry Night Van Gogh Vase & Earth & Starry Night Van Gogh Vase & Earth
\end{tabular}
\caption{We compare our approach against an online tool as of November 2023. We omit the name of the tool to protect their product, and to protect the authors from any backlash. We find that online tools can produce high quality results, but they do not match what a user might expect. Specifically, in the results there is little blurring, and little texture stretch. At the same time, there is much less diversity in their output than our approach.}
\end{figure*}
\begin{figure*}
    \centering
\begin{tabular}{p{0.55\linewidth} p{0.20\linewidth} p{0.20\linewidth}}
     Prompt & Latent Texture Size & Guidance Scale \\
     \hline
     ``Earth'' & 196 & 7.5 \\
     \multicolumn{3}{c}{\includegraphics[width=\linewidth]{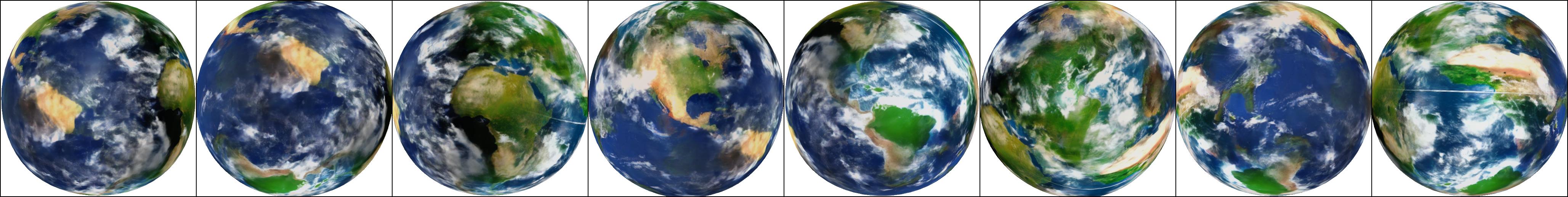}} \\
     ``Starry Night Van Gogh Vase'' & 128 & 7.5 \\
     \multicolumn{3}{c}{\includegraphics[width=\linewidth]{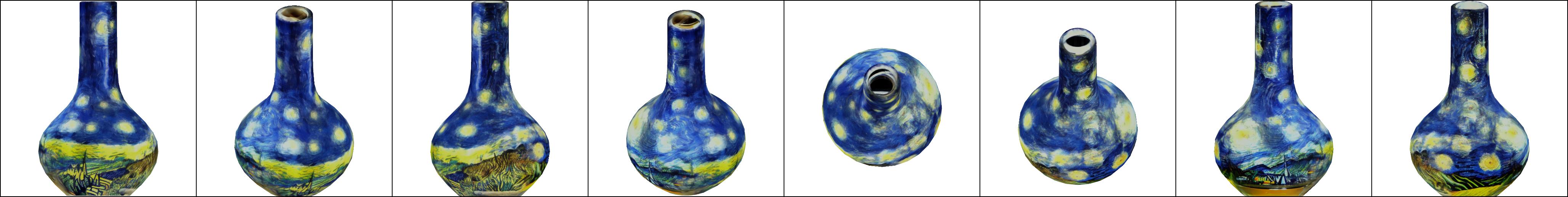}} \\
\end{tabular}
    \caption{Additional views of the above prompts.}
    \label{fig:additional_results}
\end{figure*}

\begin{figure*}[ht]
    \centering
    \begin{tabular}{c c}
        With GAN inversion & Without GAN inversion \\
        \includegraphics[width=0.4\linewidth]{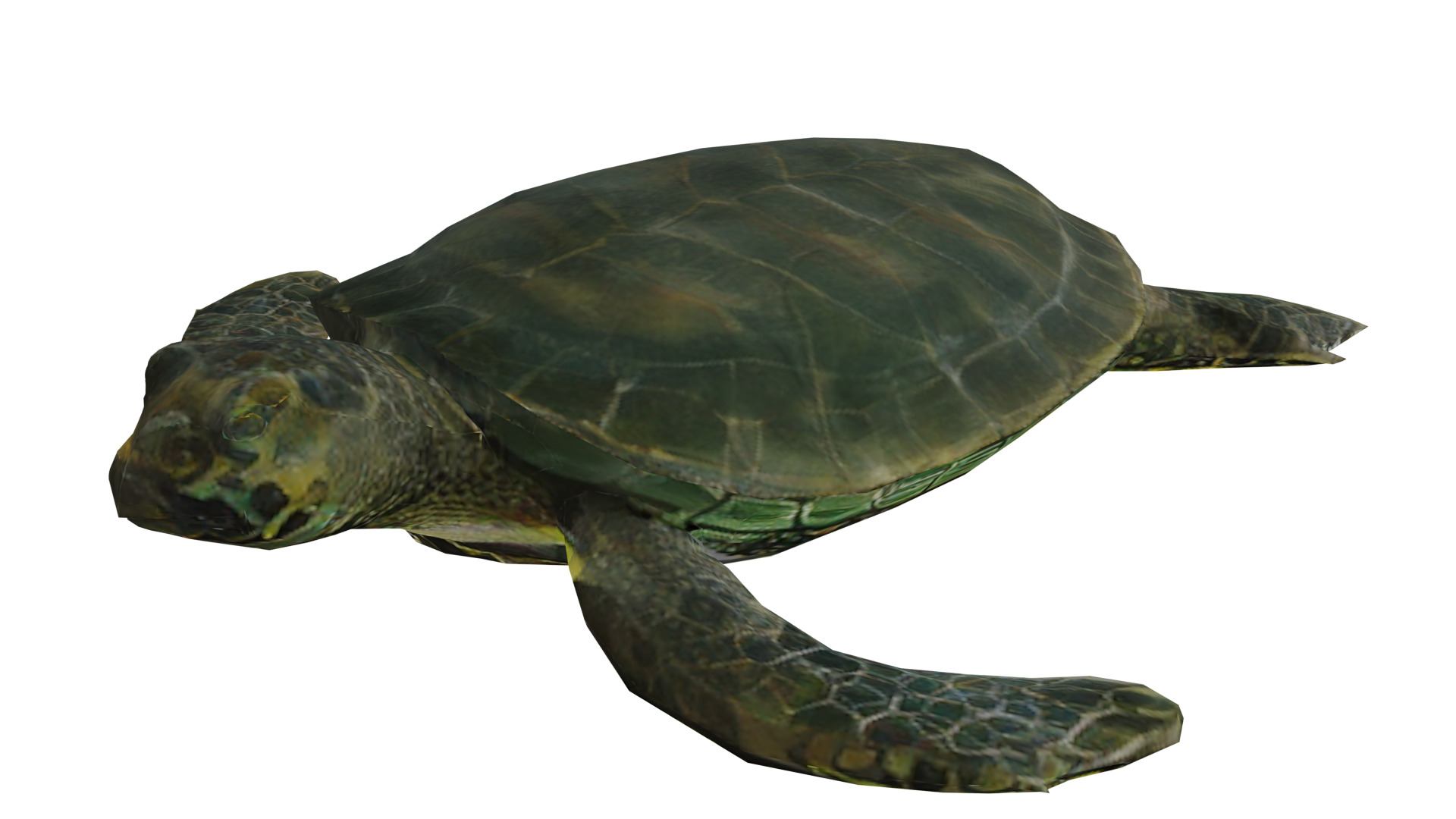} &
        \includegraphics[width=0.4\linewidth]{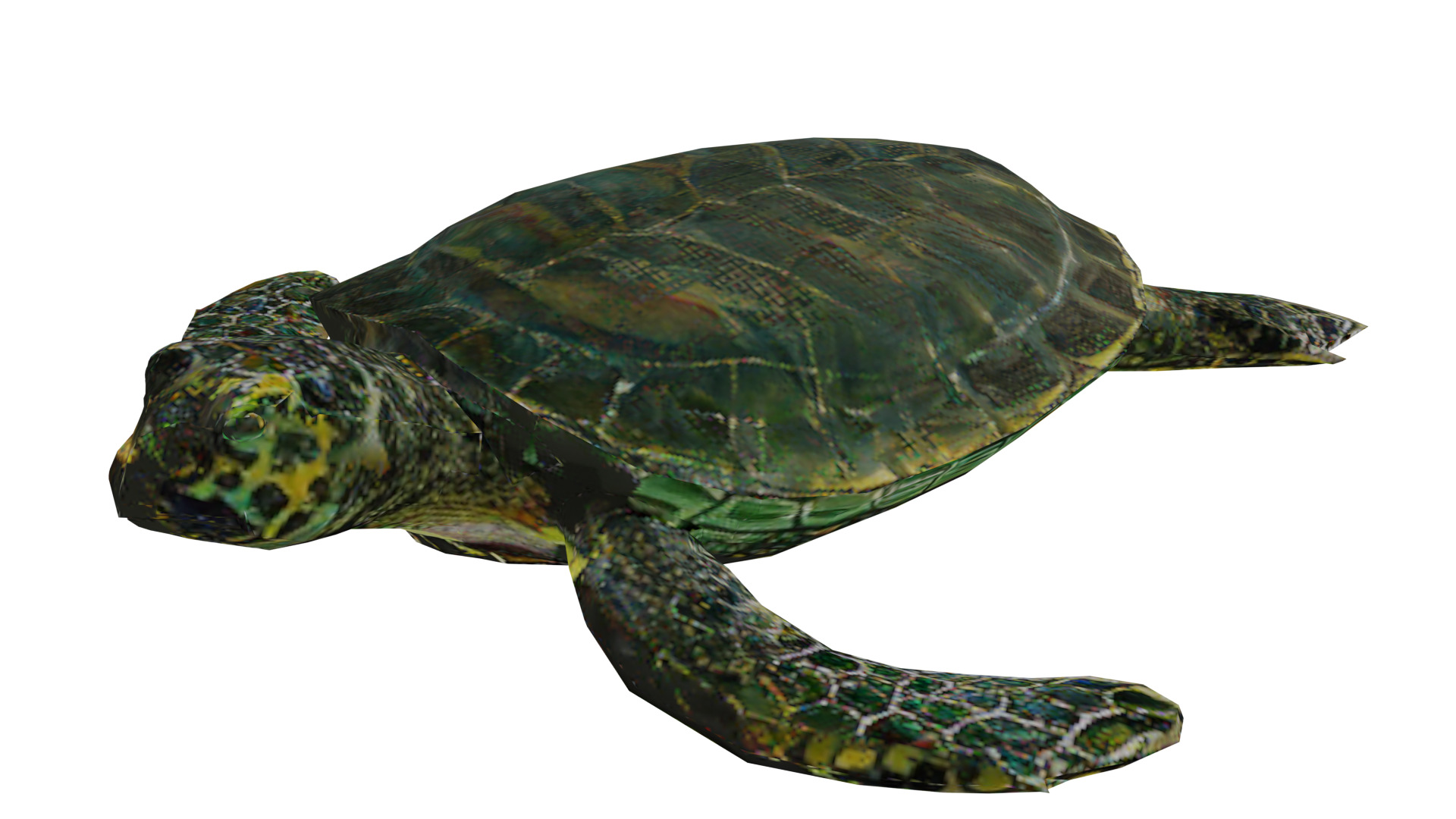} \\ 
    \end{tabular}
    \caption{Our approach with and without GAN inversion (Best Viewed Zoomed In). Without GAN inversion, many regions exhibit significant pixel-wise artifacts. GAN inversion solves this by smoothing over many artifacts, but still maintains the original appearance.}
    \label{fig:gan_inversion_ablation}
\end{figure*}

\section*{Concurrent Work}

During active development of this work, TexFusion~\cite{texfusion} was released, which is similar to our work. We consider it as concurrent to ours, as it was publicly made available a month before our submission. Our work differs from their work on multiple facets, but we cannot directly compare result quality as they did not release code for their work. First, our work does not rely on a 3D prior to fuse different textures, instead it operates within the latent space. Both approaches have a similar goal of merging inconsistent RGB views, but the quality of each is dependent on the quality of the prior.
Second, their work uses $\nabla \text{UV}$ as a per-pixel weight, whereas we rely on the $\langle\text{normal},\text{view}\rangle$. In principle, these ideas are similar, and it is not clear which is better. One note is that if the UV mapping is poor, there may not be a view from which $\nabla \text{UV} = 1$, but there is always a view where the camera is oriented directly at a face. Third, their approach additionally has cascaded multi-resolution texturing, but since they do not ablate this component it is unclear how much it contributes to the final rendering quality. Fourth, it is unclear how much they fine-tune cameras for each mesh. While in practice a user would definitely want such a feature, for comparisons to prior work it would bias their result in their favor. Our approach uses the same canonical set of cameras, but we increase the weight of the forward facing camera for specific meshes such as on people's faces. Finally, we also introduce the ability to vary per view independence through spherical harmonic coefficients, and the parameter $\alpha$. This allows for a smooth interpolation between complete correlation and full disentanglement, whereas TexFusion~\cite{texfusion} uses a single texture map, which is equivalent to full dependence between views. Spherical harmonics also provides a complete analogy between our 2D consistent diffusion, which is not something that TexFusion includes.

\begin{figure*}
\begin{tabular}{p{0.55\linewidth} p{0.20\linewidth} p{0.20\linewidth}}
     Prompt & Latent Texture Size & Guidance Scale \\
     \hline
     ``Hawaiian Shirt'' & 128 & 20 \\
     \multicolumn{3}{c}{\includegraphics[width=\linewidth]{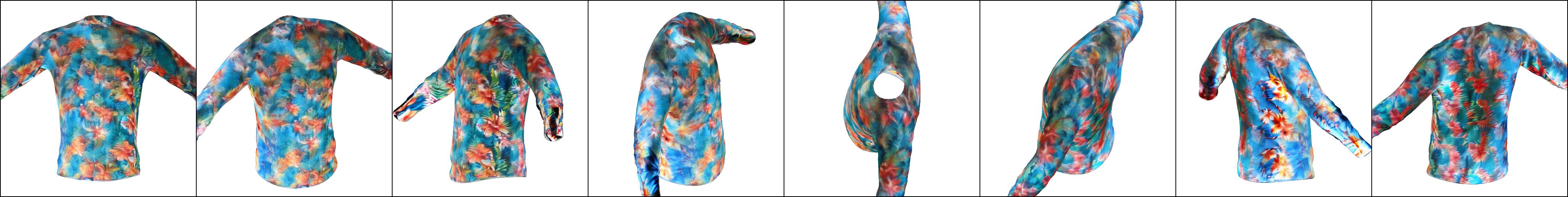}} \\
     ``Molten Magma'' & 196 & 7.5 \\
     \multicolumn{3}{c}{\includegraphics[width=\linewidth]{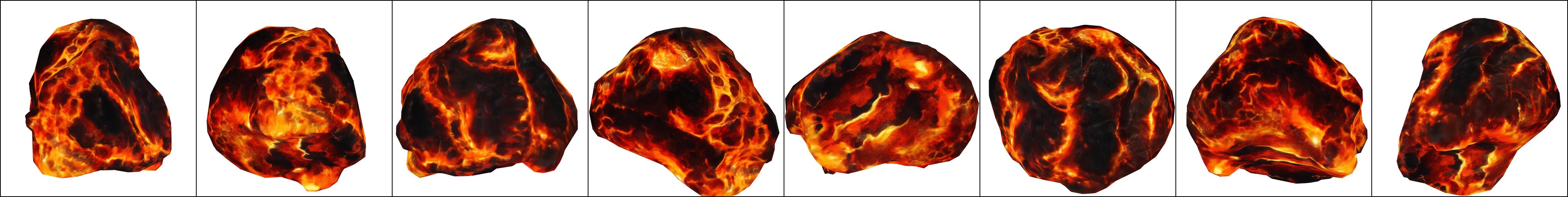}} \\
     `Moon'' & 196 & 7.5 \\
     \multicolumn{3}{c}{\includegraphics[width=\linewidth]{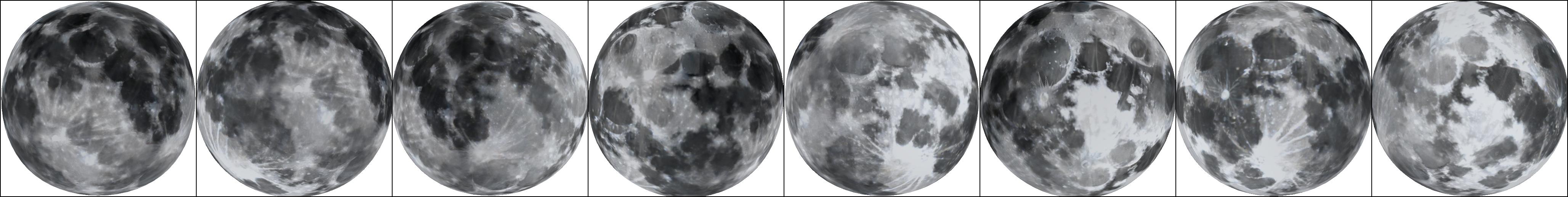}} \\
     ``Piranha Fish'' & 196 & 7.5 \\
     \multicolumn{3}{c}{\includegraphics[width=\linewidth]{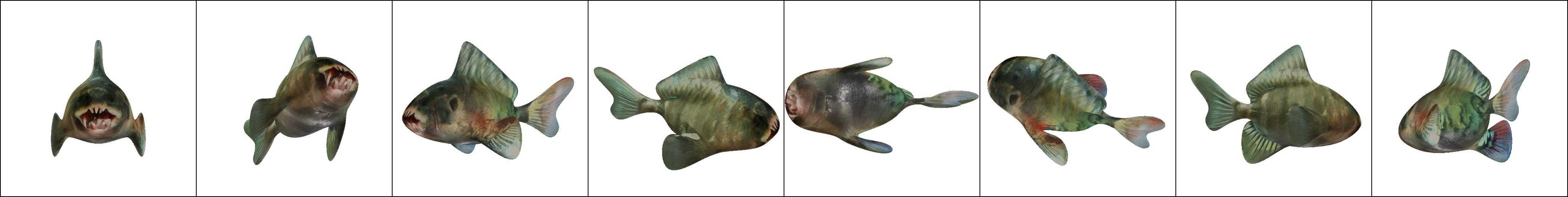}} \\
     ``Coffee Can'' & 196 & 7.5 \\
     \multicolumn{3}{c}{\includegraphics[width=\linewidth]{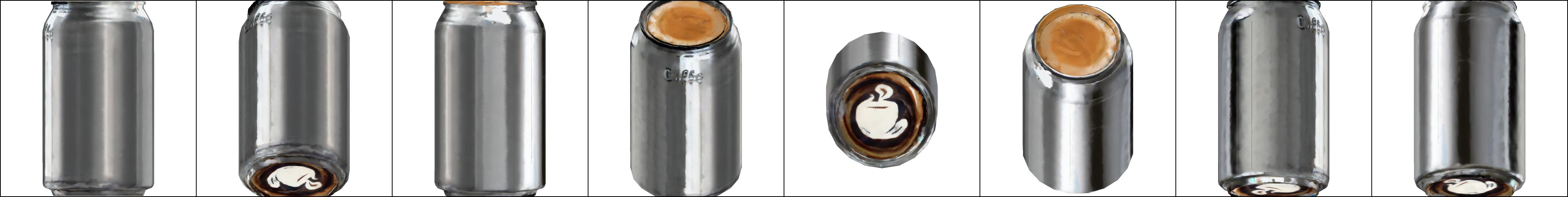}} \\
     ``Toyota Sprinter Trueno AE86'' & 128 & 7.5 \\
     \multicolumn{3}{c}{\includegraphics[width=\linewidth]{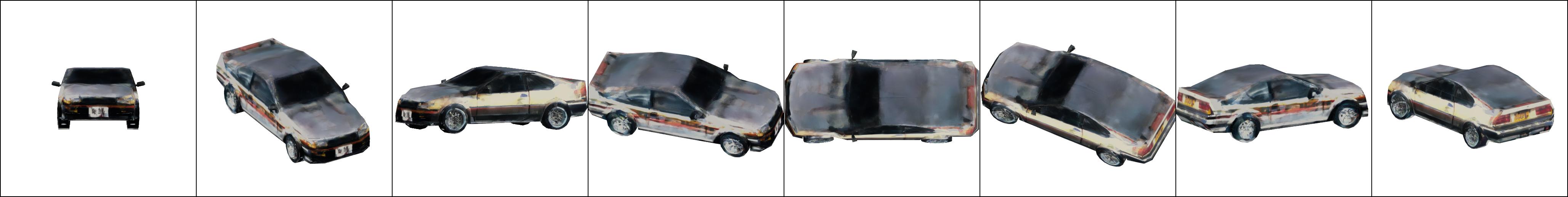}} \\
     ``Necromancer Dullahn'' & 196 & 7.5 \\
     \multicolumn{3}{c}{\includegraphics[width=\linewidth]{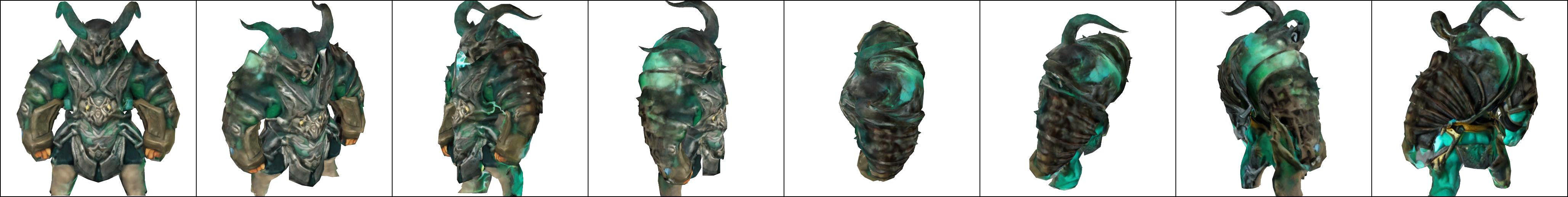}} \\
\end{tabular}
\caption{We show additional results from our multi-diffusion process on a variety of meshes. These results are all produced using the same random seed initialization, and can be deterministically reproduced. All are optimized with 8 views, and we vary texture size between 196 and 128, and guidance scale is varied between 20 and 7.5.}
\end{figure*}

\begin{figure*}
\begin{tabular}{p{0.55\linewidth} p{0.20\linewidth} p{0.20\linewidth}}
     Prompt & Latent Texture Size & Guidance Scale \\
     \hline
     ``90s Boombox'' & 196 & 7.5 \\
     \multicolumn{3}{c}{\includegraphics[width=\linewidth]{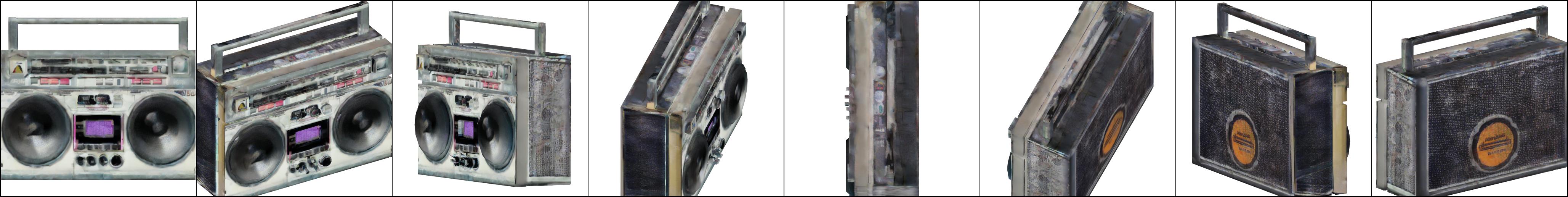}} \\
     ``Blue Kanken Fjallraven Backpack'' & 196 & 20 \\
     \multicolumn{3}{c}{\includegraphics[width=\linewidth]{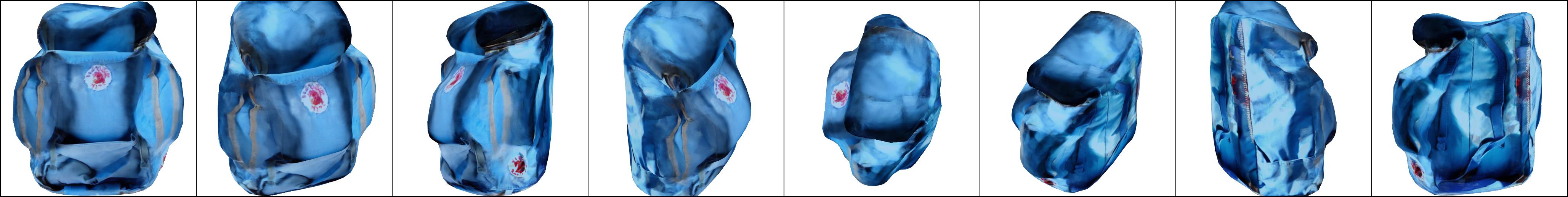}} \\
     ``Pink Axolotl'' & 196 & 7.5 \\
     \multicolumn{3}{c}{\includegraphics[width=\linewidth]{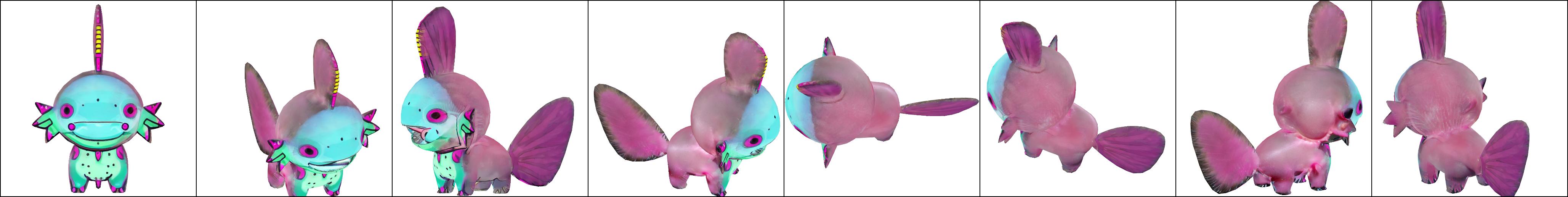}} \\
     ``Neon Green Nike Air Zoom Fencer Volt'' & 196 & 7.5 \\
     \multicolumn{3}{c}{\includegraphics[width=\linewidth]{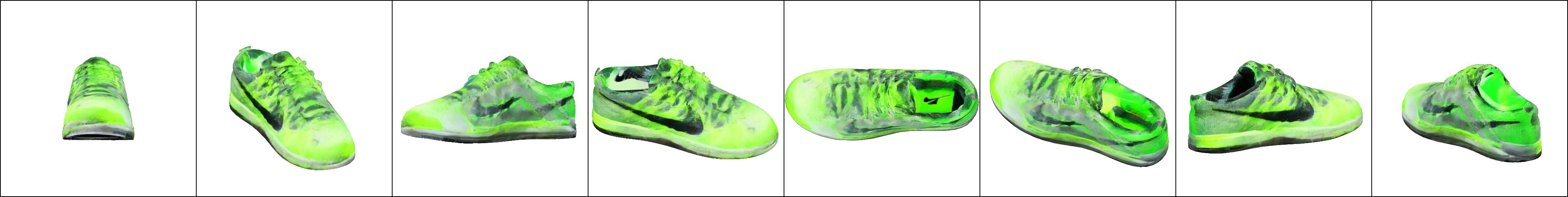}} \\
     ``Famille Rose Teapot'' & 196 & 7.5 \\
     \multicolumn{3}{c}{\includegraphics[width=\linewidth]{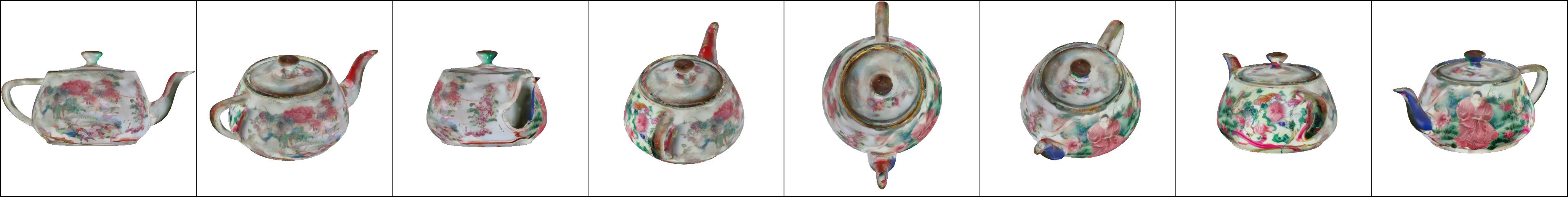}} \\
     ``Chihuly Vase'' & 196 & 7.5 \\
     \multicolumn{3}{c}{\includegraphics[width=\linewidth]{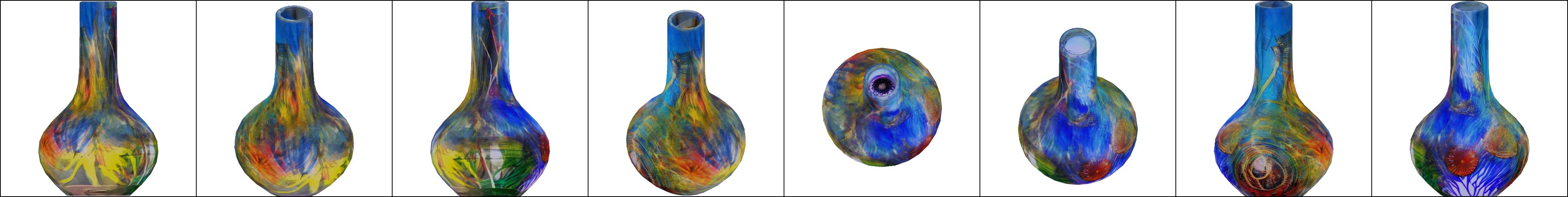}} \\
     ``Apartment Building'' & 196 & 7.5 \\
     \multicolumn{3}{c}{\includegraphics[width=\linewidth]{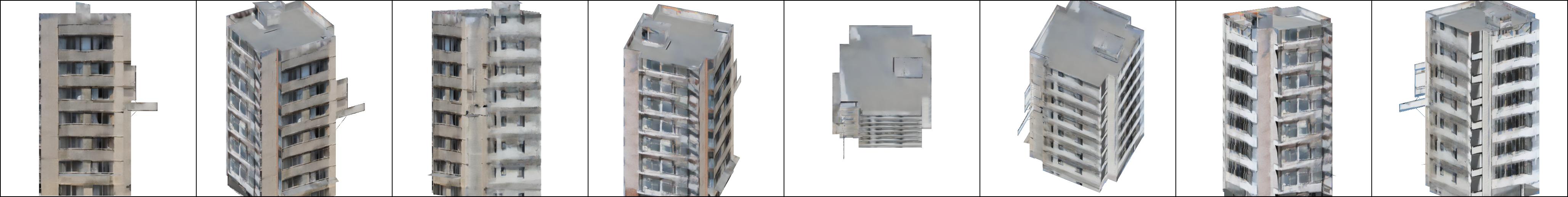}} \\
     ``Parrot'' & 196 & 7.5 \\
     \multicolumn{3}{c}{\includegraphics[width=\linewidth]{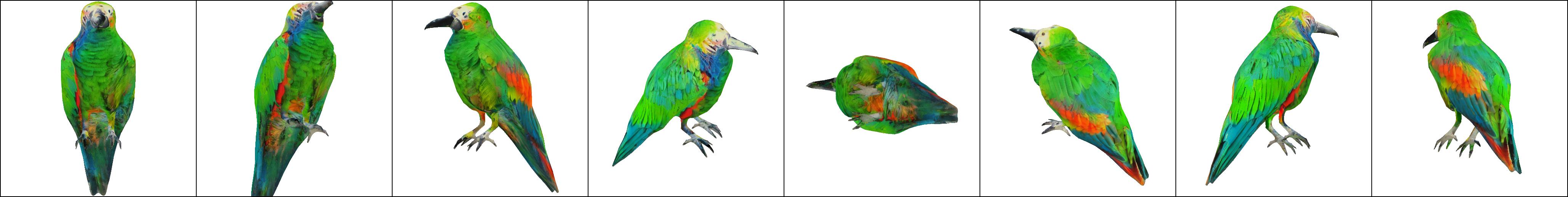}} \\
\end{tabular}
\end{figure*}

\begin{figure*}
\begin{tabular}{p{0.55\linewidth} p{0.20\linewidth} p{0.20\linewidth}}
     Prompt & Latent Texture Size & Guidance Scale \\
     \hline
     ``Musk Ox'' & 196 & 7.5 \\
     \multicolumn{3}{c}{\includegraphics[width=\linewidth]{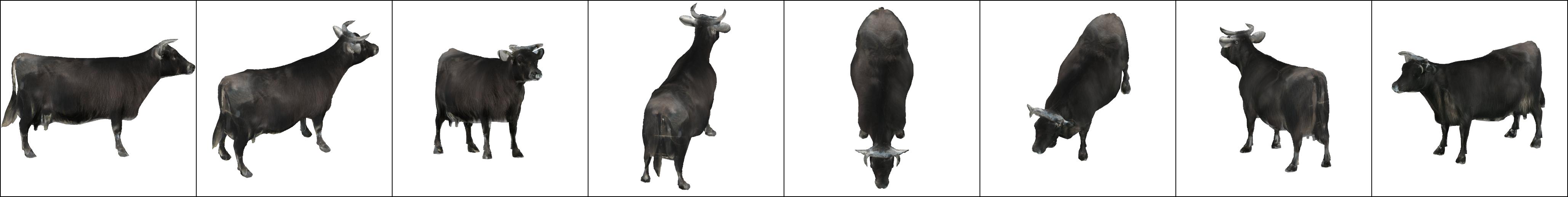}} \\
     ``Compass'' & 196 & 7.5 \\
     \multicolumn{3}{c}{\includegraphics[width=\linewidth]{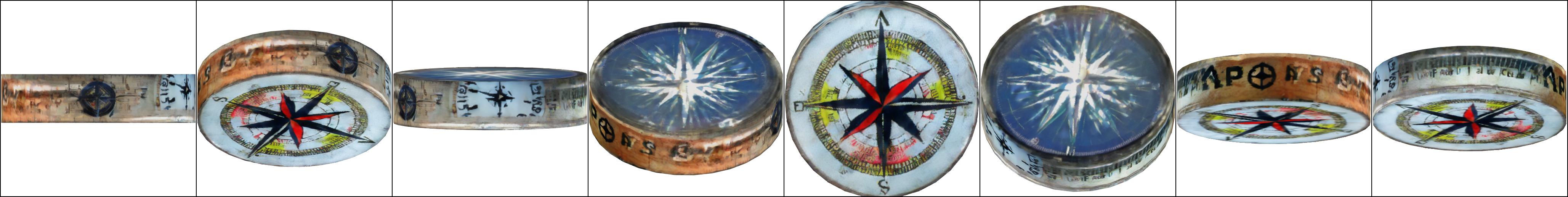}} \\
     ``Gold Trim Yunomi Teacup with a Fish Swimming in Milk Tea'' & 196 & 7.5 \\
     \multicolumn{3}{c}{\includegraphics[width=\linewidth]{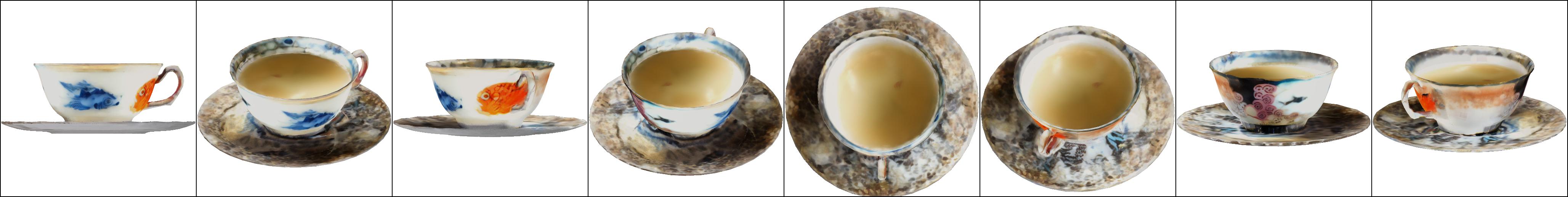}} \\
     ``Bougainvillea Bush'' & 128 & 7.5 \\
     \multicolumn{3}{c}{\includegraphics[width=\linewidth]{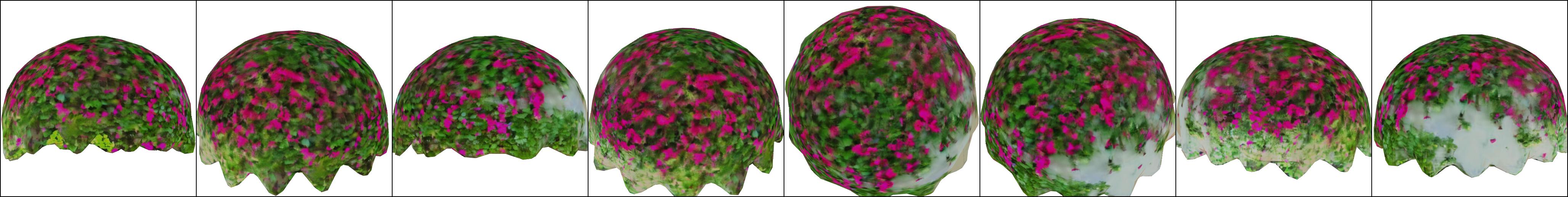}} \\
     ``Obama Painting'' & 128 & 7.5 \\
     \multicolumn{3}{c}{\includegraphics[width=\linewidth]{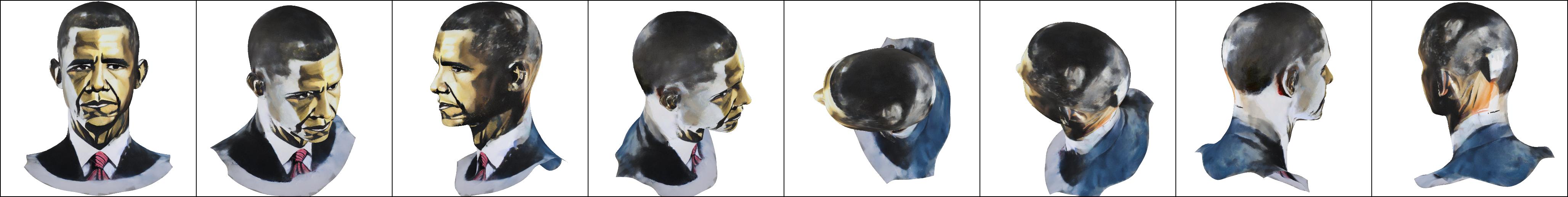}} \\
     ``Wine Barrel'' & 196 & 7.5 \\
     \multicolumn{3}{c}{\includegraphics[width=\linewidth]{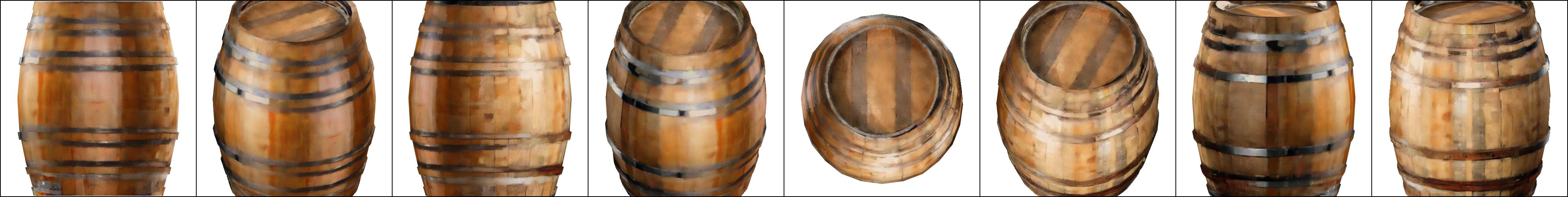}} \\
     ``C3PO'' & 196 & 7.5 \\
     \multicolumn{3}{c}{\includegraphics[width=\linewidth]{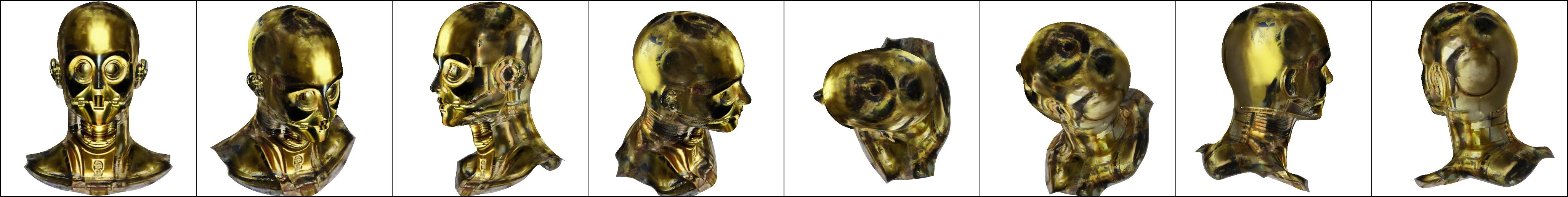}} \\
     ``Glazed Donut'' & 128 & 7.5 \\
     \multicolumn{3}{c}{\includegraphics[width=\linewidth]{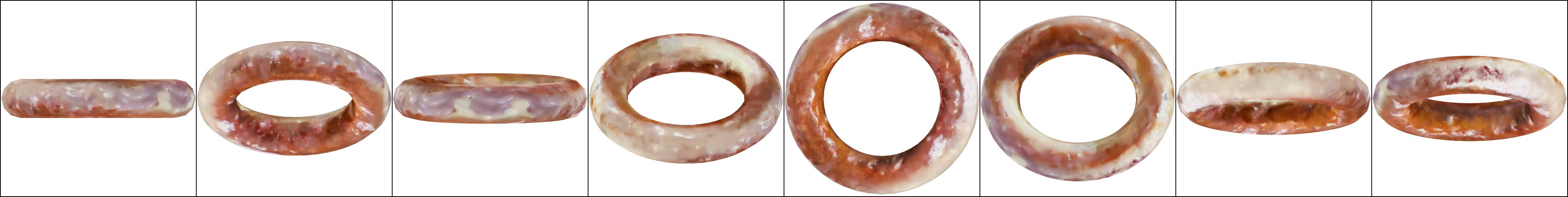}} \\
\end{tabular}
\end{figure*}

\begin{figure*}
\begin{tabular}{p{0.55\linewidth} p{0.20\linewidth} p{0.20\linewidth}}
     Prompt & Latent Texture Size & Guidance Scale \\
     \hline
     ``Violin'' & 128 & 7.5 \\
     \multicolumn{3}{c}{\includegraphics[width=\linewidth]{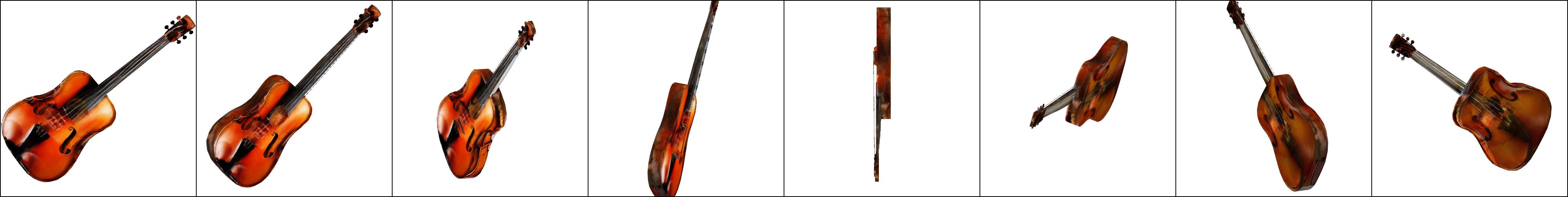}} \\
     ``Victorian Throne with Fleur de Lis'' & 196 & 7.5 \\
     \multicolumn{3}{c}{\includegraphics[width=\linewidth]{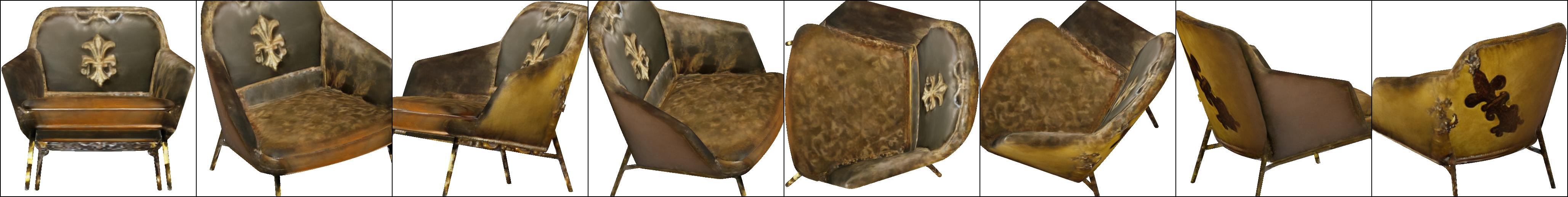}} \\
     ``Wooden Klein Bottle'' & 196 & 7.5 \\
     \multicolumn{3}{c}{\includegraphics[width=\linewidth]{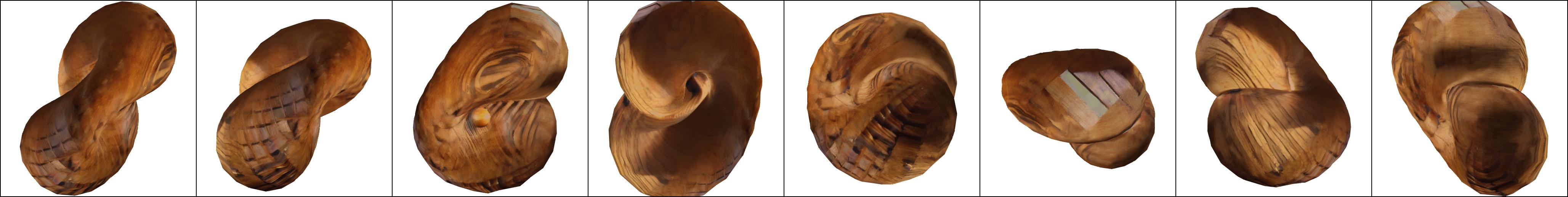}} \\
     ``Albert Einstein'' & 196 & 7.5 \\
     \multicolumn{3}{c}{\includegraphics[width=\linewidth]{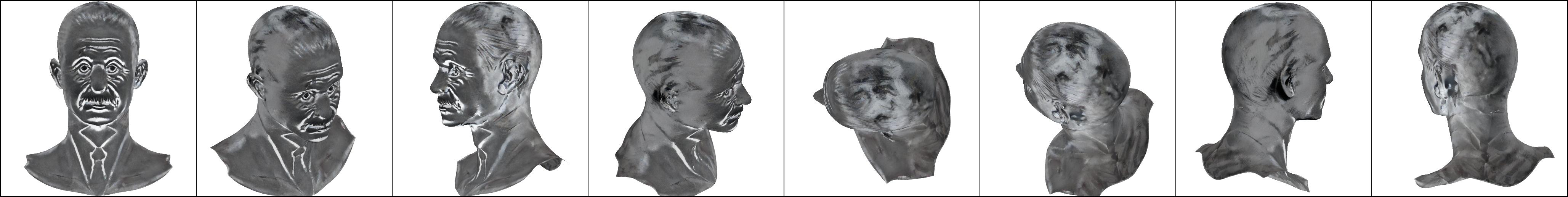}} \\
     ``Apple iMac'' & 128 & 7.5 \\
     \multicolumn{3}{c}{\includegraphics[width=\linewidth]{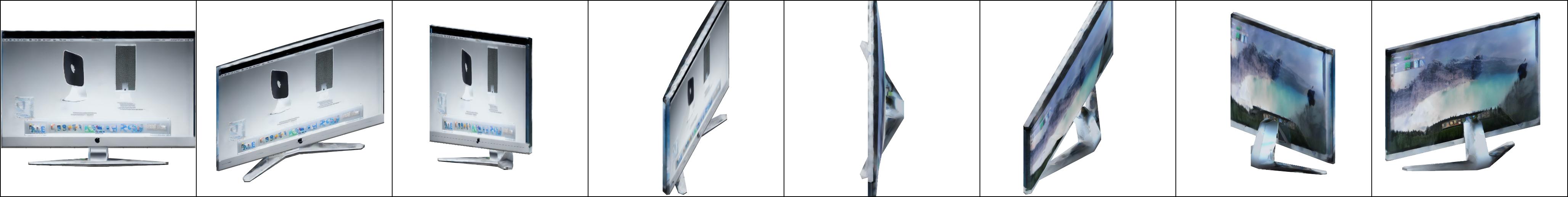}} \\
     ``Purple Geode'' & 196 & 7.5 \\
     \multicolumn{3}{c}{\includegraphics[width=\linewidth]{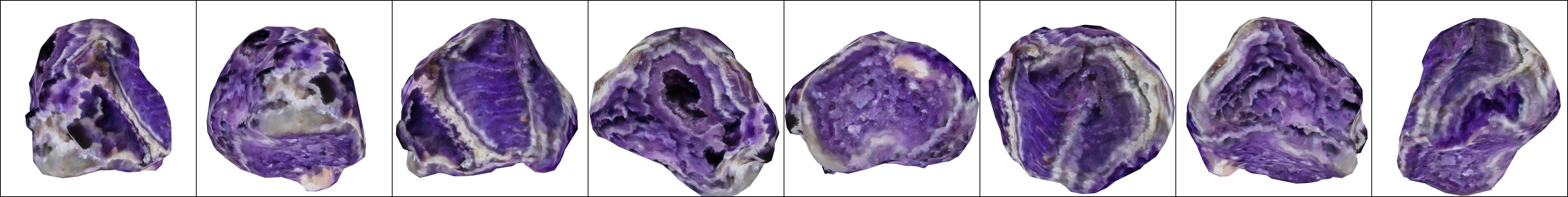}} \\
     ``Turtle'' & 128 & 7.5 \\
     \multicolumn{3}{c}{\includegraphics[width=\linewidth]{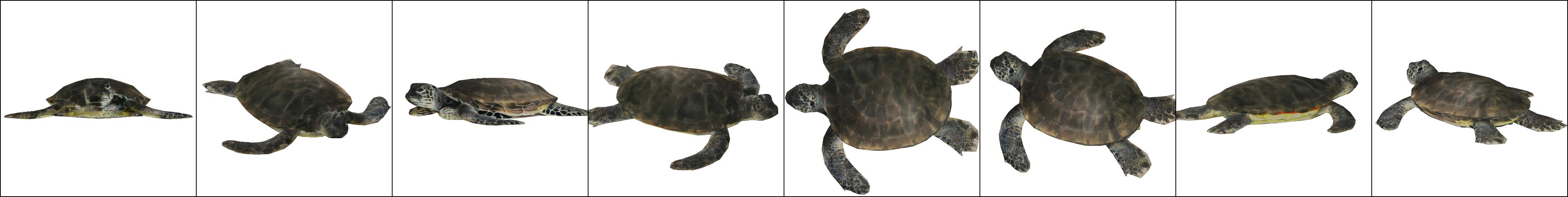}} \\
     ``Pufferfish'' & 196 & 7.5 \\
     \multicolumn{3}{c}{\includegraphics[width=\linewidth]{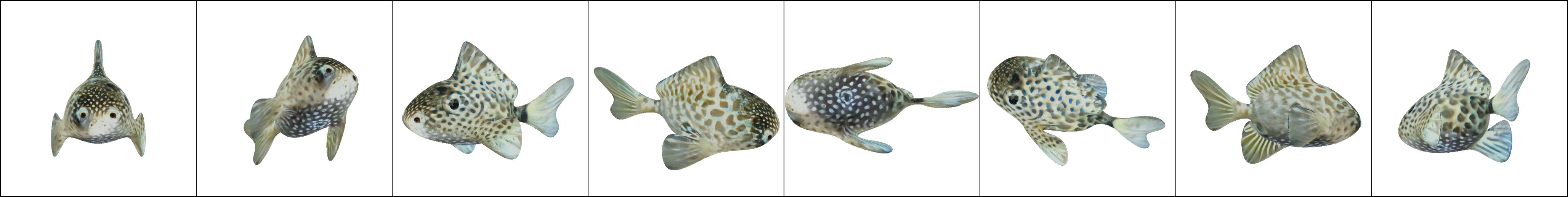}} \\
\end{tabular}
\end{figure*}

\begin{figure*}
\centering
\begin{tabular}{|c|p{6cm}|p{6cm}|}
    \hline
     Mesh Name & Prompts & Source/Artist  \\
    \hline
     Sphere & Earth, Jupiter, Moon, Cabbage & Blender Built-In Shape \\
     \hline
     Human Face & Obama Painting, Albert Einstein, Batman, C3PO & Sketchfab/hannibalhero8 \\
     \hline
    Shirt & Indian Sari, Hawaiian Shirt, Red Gold Changsam & Sketchfab/Kodie Russell \\
    \hline
Vase & Piet Mondrian Vase, Starry Night Van Gogh Vase, Chinese Vase, Chihuly Vase & Sketchfab/Nichgon \\
\hline
Blub & Koi, Pufferfish, Nemo Goldfish, Piranha Fish & Keenan Crane \\
\hline
Apartment & Pagoda, Apartment Building, Big Ben & Sketchfab/Colin.Greenall \\
\hline
Crow & Parrot, Crow, Pigeon & Sketchfab/ClintonAbbott.Art \\
\hline
Car & Toyota Sprinter Trueno AE86, Green Porsche Taycan Turbo S 2020 & TEXTure~\cite{TEXTure}, nascar.obj \\
\hline
Cow & Cow, Sheep, Musk Ox & Common 3D Test Models \newline Viewpoint Animation Engineering \\
\hline
Dog & Shiba Inu, Cat & Sketchfab/Jéssica Magno \\
\hline
Turtle & Turtle & Sketchfab/liamgamedev \\
\hline
Can & Coffee Can, Campbell Soup Can & Sketchfab/Blender3D (artist's name) \\
\hline
Cube & Bricks, Dice & Blender Built-In Shape \\
\hline
Rock & Purple Geode, Mossy Cobblestone, Molten Magma & Artist/Xephira \\
\hline
Steve & Minecraft Steve, Minecraft Creeper & Sketchfab/Vincent Yanex \\
\hline
Torus & Glazed Donut, Floaty & Blender Built-In Shape \\
\hline
Shoe & Red Converse Shoe, Neon Green Nike Air Zoom Fencer Volty & Sketchfab/DailyArt \\
\hline
Chunky Knight & Paladin, Hulk from Star Wars, Necromancer Dullahan & Sketchfab/thanhtp \\
\hline
Napoleon & Napoleon, Clown & TEXTure~\cite{TEXTure}, napoleon.obj \\
\hline
Mudkip & Mudkip, Pink Axolotl & Sketchfab/jacobjksn42 \\
\hline
Teapot & Famille Rose Teapot, Piet Mondrian Teapot & Utah Teapot \\
\hline
Stickman & Megaman, Jet Set Radio Beat & Sketchfab/studentsimf \\
\hline
Chair & Wicker Chair, Steampunk Chair, Victorian Throne with Fleur de Lis & Sketchfab/maxsbond.work \\
\hline
Boombox & 90s Boombox, Ukiyo-e Boombox & Sketchfab/Poly by Google \\
\hline
Guitar & Heavy Metal Guitar, Violin & Sketchfab/Ya \\
\hline
Bunny & Realistic Snow White Rabbit & Stanford Bunny \\
\hline
Klein Bottle & Wooden Klein Bottle & Sketchfab/dpiker \\
\hline
Backpack & Orange Backpack, Blue Kanken Fjallraven Backpack & Sketchfab/Liam3D \\
\hline
Barrel & Wine Barrel, Bejeweled Explosive Barrel & Sketchfab/Joseph Gush \\
\hline
Compass & Compass, Clock  & Sketchfab/Jen S Abbott \\
\hline
Monitor & Apple IMac, Windows Desktop & Sketchfab/Artik \\
\hline
Room & Isometric Gaming Room, Isometric Japanese Tatami Room & Sketchfab/Ava editz \\
\hline
Bush & Rose Bush, Bougainvillea Bush & Sketchfab/Natural\_Disbuster \\
\hline
Teacup & Gold Trim Japanese Yunomi Teacup with a Fish Swimming in Milk Tea, Terracotta Teacup Filled with Poison & Sketchfab/Buntaro \\
\hline
\end{tabular}
\end{figure*}

\end{document}